\documentclass[10pt,twocolumn,letterpaper]{article}

\usepackage[pagenumbers]{cvpr} 

\usepackage{graphicx}
\usepackage{amsmath}
\usepackage{amssymb}
\usepackage{booktabs}

\usepackage[pagebackref,breaklinks,colorlinks]{hyperref}

\usepackage[capitalize]{cleveref}
\crefname{section}{Sec.}{Secs.}
\Crefname{section}{Section}{Sections}
\Crefname{table}{Table}{Tables}
\crefname{table}{Tab.}{Tabs.}
\usepackage{xspace}

\def\cvprPaperID{5604} 
\def\confName{CVPR}
\def\confYear{2023}

\usepackage[x11names, table]{xcolor}

\usepackage{bbm}
\usepackage{acro}
\usepackage{xspace}

\usepackage{pifont}
\newcommand{\cmark}{\ding{51}}%
\newcommand{\xmark}{\ding{55}}%

\usepackage{multirow}

\usepackage{subcaption}

\usepackage{tabularx}
\usepackage{rotating}

\usepackage{siunitx}

\usepackage[accsupp]{axessibility}  

\def\doubleunderline#1{\underline{\underline{#1}}}

\newcommand{\pl}{pseudo-label\xspace}
\newcommand{\pls}{pseudo-labels\xspace}
\newcommand{\snerf}{\hbox{Semantic-}NeRF\xspace}

\begin{document}

\setlength{\abovedisplayskip}{2pt}
\setlength{\belowdisplayskip}{2pt}

\title{Unsupervised Continual Semantic Adaptation through Neural Rendering}

\author{
\parbox{\linewidth}{\centering
Zhizheng Liu\textsuperscript{1}\thanks{Authors share first authorship.}
\hspace{3ex}
Francesco Milano\textsuperscript{1}\footnotemark[1]
\hspace{3ex}
Jonas Frey\textsuperscript{1,2}
\hspace{3ex}
Roland Siegwart\textsuperscript{1}
}
\\
\parbox{\linewidth}{\centering
Hermann Blum\textsuperscript{1}\thanks{Authors share senior authorship.}
\hspace{3ex}
Cesar Cadena\textsuperscript{1}\footnotemark[2]
}
\\[0.5ex]
\parbox{\linewidth}{\centering \textsuperscript{1}ETH Zurich \hspace{3ex} \textsuperscript{2}Max Planck ETH Center for Learning Systems}
}

\twocolumn[{%
\maketitle
\captionsetup{type=figure}
\vspace{-27pt}
\begin{center}
\centering
\subfloat[Pseudo-label formation]{
\centering
\includegraphics[height=0.26\linewidth]{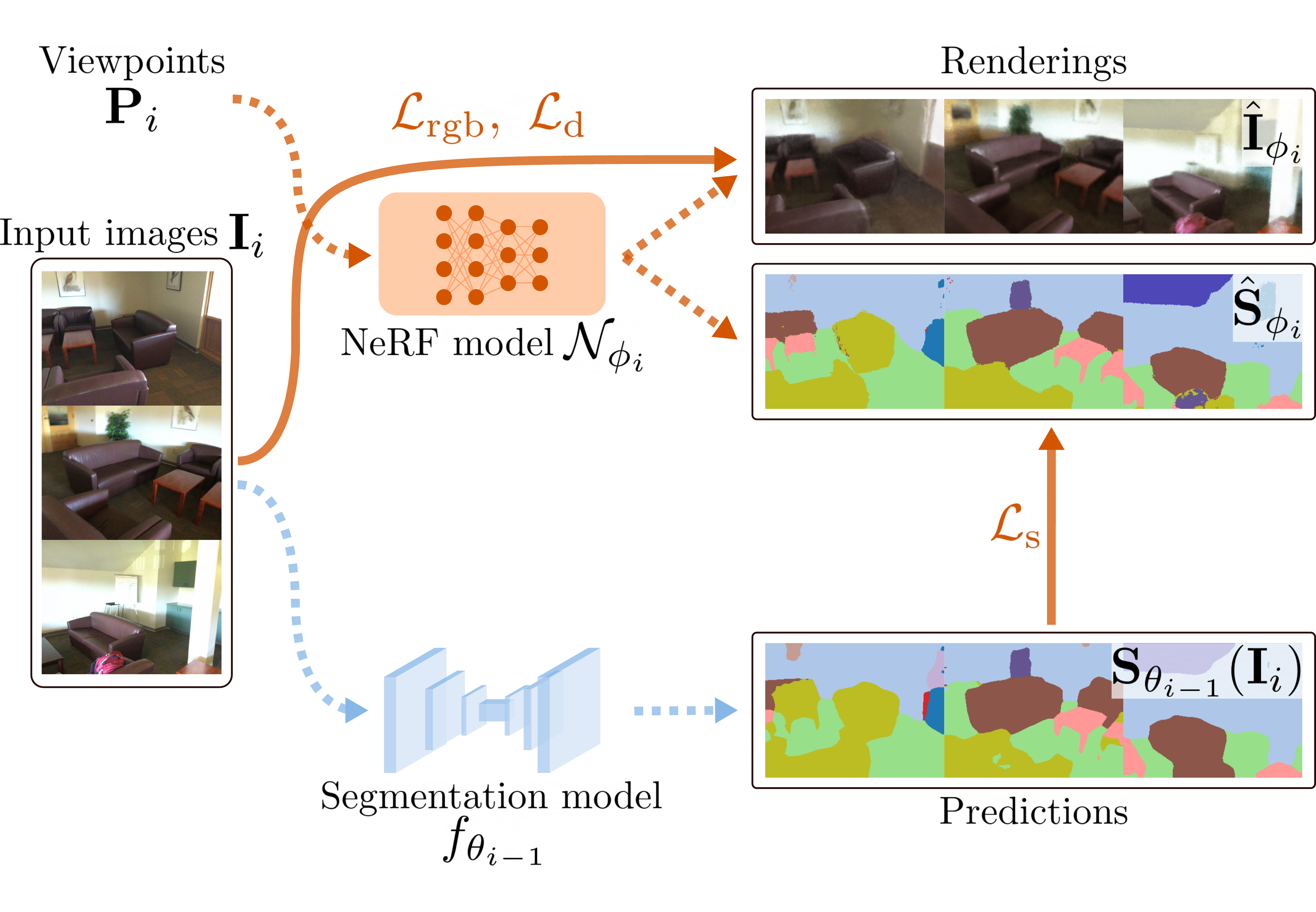}
}
\hfill
\subfloat[Adaptation through joint training and long-term memory]{
\centering
\includegraphics[height=0.26\linewidth]{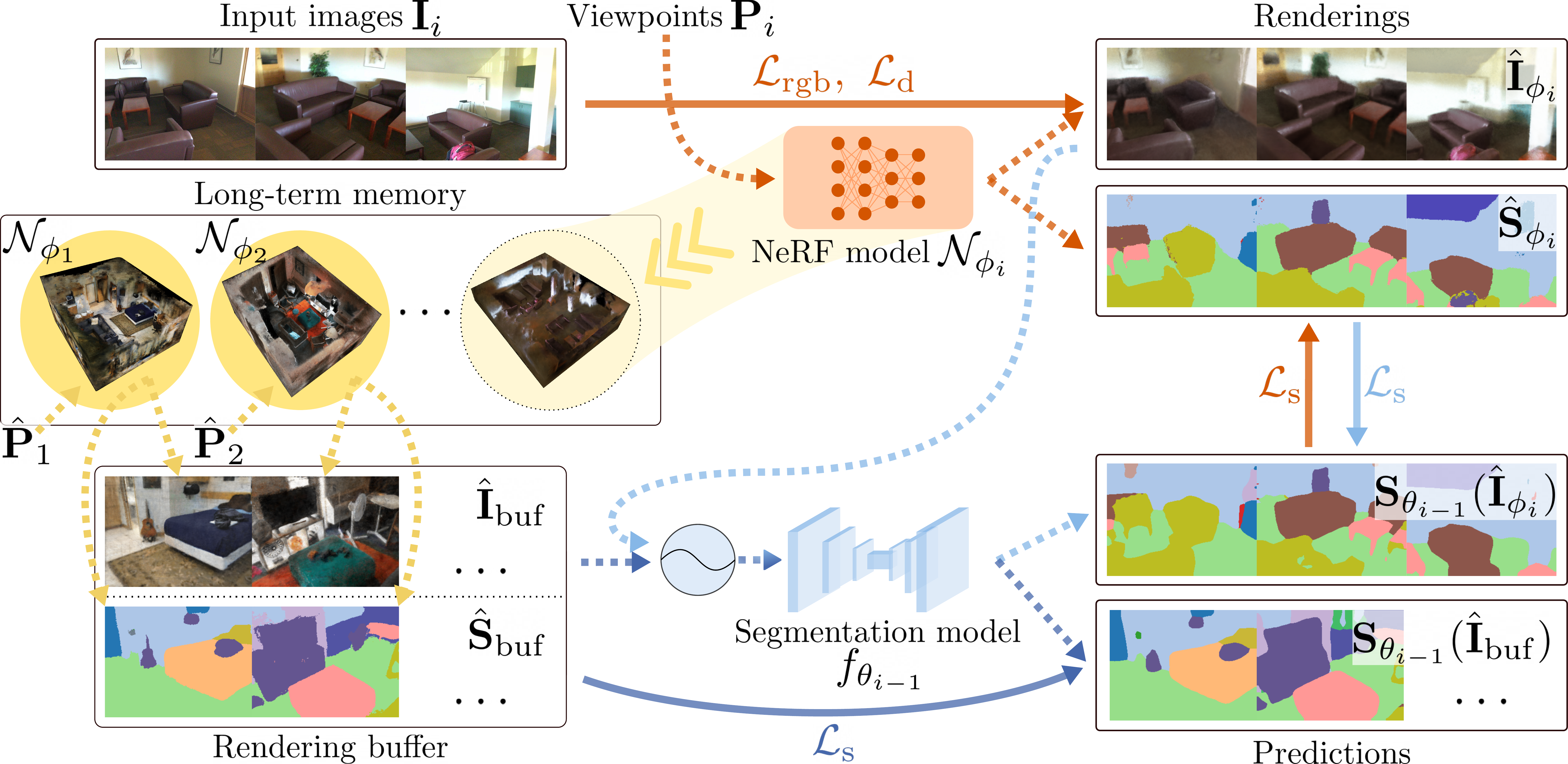}
}
\vspace{-2pt}
 \caption{We propose a method to continually adapt a
 semantic
 segmentation model $f$ in an unsupervised fashion across multiple scenes, 
 using neural rendering. For each scene $\mathcal{S}_i$: a) RGB(-D) images $\mathbf{I}_i$
 from multiple viewpoints $\mathbf{P}_i$ and their corresponding predictions $\mathbf{S}_{\theta_{i-1}}(\mathbf{I}_i)$
by
 the latest model $f_{\theta_{i-1}}$ are used to supervise a (Semantic-)NeRF model $\mathcal{N}_{\phi_i}$;
b) Adaptation on $\mathcal{S}_i$ is performed through a \emph{joint training}, in which the segmentation network is supervised using the 3D-aware, view-consistent pseudo-labels $\hat{\mathbf{S}}_{\phi_i}$ rendered from $\mathcal{N}_{\phi_i}$ and the NeRF model through the smooth predictions of $f_{\theta_{i-1}}$.
For each scene, the NeRF model can be compactly stored in a long-term memory, from which images and pseudo-labels from arbitrary viewpoints $\hat{\mathbf{P}}$ can be rendered into a fixed-size rendering buffer and mixed with the renderings from the current scene to reduce forgetting.
 Bold 
 and dotted lines denote supervision signals and inputs/outputs, respectively.
 \label{fig:method_overview}}
 \vspace{-2pt}
\end{center}}]

\renewcommand*{\thefootnote}{\fnsymbol{footnote}}
\footnotetext[1]{Authors share first authorship.\qquad${}^\dagger$
Authors share senior authorship.}
\renewcommand*{\thefootnote}{\arabic{footnote}}
\begin{abstract}
\vspace{-5pt}
An increasing amount of applications
rely on data-driven models that are deployed for perception tasks across a sequence of
scenes. Due to the mismatch between
training and deployment data,
adapting the model on the new scenes is often crucial to obtain good performance. In this work, we 
study continual multi-scene
adaptation for the task of semantic segmentation,
assuming that no ground-truth labels are available during deployment and that performance on the previous scenes should be maintained.
We propose training a \snerf network for each scene by fusing the predictions of a segmentation model and then using the view-consistent rendered semantic labels as \pls to adapt the model.
Through joint training with the segmentation model, the \snerf model effectively enables 2D-3D knowledge transfer.
Furthermore, due to its compact size, it can be stored in a long-term memory and subsequently used to render
data
from arbitrary viewpoints to reduce forgetting.
We evaluate our approach on ScanNet, where we outperform both a voxel-based baseline and a state-of-the-art unsupervised domain adaptation method.
\end{abstract}
\vspace{-20pt}
\vspace{-10pt}
\section{Introduction}
\label{sec:intro}
Data-driven models trained for perception tasks play an increasing role
in applications
that rely on scene understanding, 
including, \eg, mixed reality
and
robotics.
When deploying
these models
on real-world systems, however, mismatches between the data used for training and those encountered during deployment can lead to poor performance,
prompting the need for an \textit{adaptation} of the
models
to the
new
environment. Oftentimes,
the supervision data required for this adaptation can only be obtained through a laborious labeling process.
Furthermore, 
even when such data
are
available,
a
na\"ive adaptation
to
the new environment results in
decreased performance on the original training data, a phenomenon known as \emph{catastrophic forgetting}\cite{Lesort2020CLForRobotics, Michieli2022DomainAdaptationCLChapter}.

In this work, we focus on the task of adapting a semantic segmentation network 
across
multiple
indoor scenes, under the assumption that no labeled data from the new environment are available. Similar settings are explored in the literature in the areas of \emph{unsupervised domain adaptation (UDA)}~\cite{Michieli2022DomainAdaptationCLChapter, Toldo2020UDAReview} and \emph{continual learning (CL)}~\cite{Lesort2020CLForRobotics}.
However,
works in the UDA literature usually focus on a single source-to-target transfer where the underlying assumption is that the data from both the source and the target domain are available all at once in the respective training stage, and often study the setting in which the knowledge transfer happens between a synthetic and a real environment~\cite{Richter2016GTA, Ros2016SYNTHIA, Cordts2016Cityscapes, Toldo2020UDAReview}.
On the other hand, the CL community, 
which generally explores the adaptation of networks across different \emph{tasks},
has established the \textit{class-incremental} setting as the standard for semantic segmentation, in which new classes are introduced across different scenes from the same domain and ground-truth supervision is provided~\cite{Michieli2022DomainAdaptationCLChapter}.
In contrast, we propose to study network adaptation in a setting that more closely resembles the deployment of semantic networks on real-world systems. In particular, instead of assuming that data from a specific domain
are
available all at once,
we focus on the scenario in which the network is sequentially deployed in multiple \emph{scenes} from a real-world indoor environment (we use the ScanNet dataset~\cite{Dai2017ScanNet}), and therefore has to perform multiple
\emph{stages}
of adaptation from one scene to another.
Our setting further includes the possibility that previously seen scenes may be revisited. Hence, we are interested in achieving high prediction accuracy on each new scene, while at the same time preserving performance on the previous
ones.
Note that
unlike
the
better
explored
class-incremental 
CL,
in
this
setting
we assume a \emph{closed set} of semantic
categories,
but tackle the
covariate shift
across scenes
without the need for ground-truth labels.
We refer to this setting as \emph{continual semantic adaptation}.

In this work, we propose to address this adaptation problem by leveraging advances in neural rendering~\cite{Mildenhall2020NeRF}. Specifically, in a similar spirit to~\cite{Frey2022CLSemanticSegmentation}, when deploying a pre-trained network in a new scene, we aggregate the semantic predictions from the multiple viewpoints traversed by the agent
into a 3D representation, from which we then render \pls that we use to adapt the network on the current scene. However, instead of relying on a voxel-based representation, we propose to aggregate the predictions through a semantics-aware NeRF~\cite{Mildenhall2020NeRF, Zhi2021SemanticNeRF}. 
This formulation has several advantages.
First,
we show that using NeRFs to aggregate the semantic predictions results in higher-quality \pls compared to the voxel-based method of~\cite{Frey2022CLSemanticSegmentation}.
Moreover, we demonstrate that using these \pls to adapt the segmentation network
results in superior performance compared both to~\cite{Frey2022CLSemanticSegmentation} and to the state-of-the-art UDA method CoTTA~\cite{Wang2022CoTTA}.
An even more interesting insight, however, is that due the differentiability of NeRF, we can jointly train the frame-level semantic network and the scene-level NeRF to 
enforce similarity between the predictions of the former and the renderings of the latter.
Remarkably, this joint procedure 
induces better performance
of both labels,
showing the benefit of mutual 2D-3D knowledge transfer. 

A further
benefit of our method is that after adapting to a new scene, the NeRF encoding the appearance, geometry and semantic content for that scene can be compactly
saved
in long-term
storage,
which effectively forms
a ``memory bank" of the previous experiences
and
can be useful in reducing catastrophic forgetting.
Specifically, by
mixing pairs of semantic and color NeRF renderings from a small number of views in the previous scenes and
from views in
the current scene, we show that our method is able to outperform both the baseline of~\cite{Frey2022CLSemanticSegmentation} and CoTTA~\cite{Wang2022CoTTA} on the adaptation
to
the new scene
and
in terms of knowledge retention on the previous scenes. Crucially, the collective size of the NeRF models is lower than that of the explicit replay buffer required by~\cite{Frey2022CLSemanticSegmentation} and of the teacher network used in CoTTA~\cite{Wang2022CoTTA} up to several dozens of scenes. Additionally,
each of the NeRF models stores a potentially infinite number of views that can be used for adaptation, not limited to the training set as in~\cite{Frey2022CLSemanticSegmentation}, and
removes
the need to explicitly keep color images and \pls in memory.

In summary, the main contributions of our work are the following: (i) We propose using NeRFs to adapt a semantic segmentation network to new scenes. 
We find that enforcing 2D-3D knowledge transfer by jointly adapting NeRF and the segmentation network
on a given scene
results in a consistent performance improvement;
(ii) We address the problem of continually adapting the segmentation network across a sequence of scenes by compactly storing the NeRF models in a long-term memory
and mixing rendered images and \pls from
previous scenes with those from the current one.
Our approach allows generating a potentially infinite 
number
of views to use for adaptation at constant memory size for each
scene;
(iii) Through extensive experiments, we show that our method achieves better 
adaptation
and
performance
on the previous scenes compared both to
a recent voxel-based method that explored a similar setting~\cite{Frey2022CLSemanticSegmentation} and to a state-of-the-art UDA method~\cite{Wang2022CoTTA}.

\section{Related work}
\label{sec:related_work}
\paragraph{Unsupervised domain adaptation for semantic segmentation.}
Unsupervised domain adaptation (UDA) studies the problem of transferring knowledge between a source and a target domain under the assumption that no labeled data for the target domain
are
available.
In the following, we provide an overview of the main techniques used in UDA for semantic segmentation and focus on those which are most closely related to our work;
for a more extensive summary
we refer the reader to the recent survey of~\cite{Michieli2022DomainAdaptationCLChapter}.

The majority of the methods rely on auto-encoder CNN architectures, and perform network adaptation either
at the level of the input data~\cite{Li2019BDL, Hoffman2018CyCADA, Chen2019CrDoCo, Zhang2018FCAN, Yang2020FDA}, of the intermediate network representations~\cite{Hoffman2018CyCADA, Chen2017NoMoreDiscrimination, Murez2018ImageToImageTranslationDA, Du2019SSF-DAN, Zhang2018FCAN}, or of the output predictions~\cite{Li2019BDL, Chen2019CrDoCo, Chen2017NoMoreDiscrimination, Sankaranarayanan2018LSD-seg, Du2019SSF-DAN, Saito2018AdversarialDropoutRegularization, Vu2019ADVENT, Zou2018CBST, Zou2019CRST, Michieli2020semanticDA, Spadotto2021semanticDA}.
The main strategies adopted consist in: using adversarial learning techniques to enforce that the
network representations have similar statistical properties across the two domains~\cite{Chen2017NoMoreDiscrimination, Hoffman2018CyCADA, Murez2018ImageToImageTranslationDA, Sankaranarayanan2018LSD-seg, Chen2019CrDoCo, Du2019SSF-DAN, Li2019BDL}, 
performing image-to-image translation to align the data from the two domains~\cite{Li2019BDL, Hoffman2018CyCADA, Chen2019CrDoCo, Zhang2018FCAN, Yang2020FDA}, learning to detect non-discriminative feature representations for the target
domain~\cite{Saito2018AdversarialDropoutRegularization, Lee2019DTA},
and using self-supervised learning based either on minimizing the pixel-level entropy in the target domain~\cite{Vu2019ADVENT} or on self-training techniques~\cite{Zou2018CBST, Zou2019CRST, Li2019BDL, Michieli2020semanticDA, Spadotto2021semanticDA, Choi2019SelfEnsemblingGANBased, Zheng2021MRNetRectifying}. The latter category of methods is the most related to our setting. In particular, a number of works use the network trained on the source domain to generate semantic predictions on the unlabeled target data; the obtained \emph{\pls} are then used as a self-supervisory learning signal to adapt the network to the target domain.
While our work and the self-training UDA methods both use \pls, the latter approaches
neither
exploit the sequential structure of the data
nor
explicitly enforce multi-view consistency in the predictions on the target data.
Furthermore, approaches in UDA mostly focus on single-stage, sim-to-real transfer settings, often for outdoor environments, and generally assume that the data from each domain are available all at once during the respective training stage.
In contrast, we focus on a multi-step adaptation problem, in which data from multiple scenes from an indoor environment are available sequentially. 
\\\indent Within the category of self-training methods, a number of works come closer to our setting by 
presenting
techniques to achieve \textit{continuous}, multi-stage domain adaptation. 
In particular, the recently proposed CoTTA~\cite{Wang2022CoTTA} uses a student-teacher framework, in which the student network is adapted to a target environment through \pls generated by the teacher 
network,
and stochastic restoration of the weights from a pre-trained model is used to preserve source knowledge. 
ACE~\cite{Wu2019ACE} proposes
a style-transfer-based
adaptation
with
replay of
feature
statistics from previous domains, but
assumes
ground-truth source labels
and
focuses on
changes of environmental conditions
within the same scene.
Finally, related to our method is also the
recent
work of Frey et al.~\cite{Frey2022CLSemanticSegmentation}, which addresses a similar problem as ours by
aggregating predictions from different viewpoints in a target domain 
into a 3D
voxel grid
and rendering \pls, but does not 
perform
multi-stage
adaptation.
\\\textbf{Continual learning for semantic segmentation.}
Continual learning for semantic segmentation (CSS) focuses on the problem of updating a segmentation network
in a \textit{class-incremental setting}, in which it is assumed
that the domain is
available in different \emph{tasks} and that new classes are added over time in a sequential fashion~\cite{Michieli2022DomainAdaptationCLChapter}.
The main objective consists in performing adaptation to the new task, mostly using only data from the current stage, while preventing
forgetting of the knowledge from
the previous tasks. The methods proposed in the literature typically adopt a combination of different strategies, including distilling knowledge from a previous model~\cite{Michieli2019ILT, Michieli2021KDIL, Cermelli2020MiB, Douillard2021PLOP}, selectively freezing the network parameters~\cite{Michieli2019ILT, Michieli2021KDIL}, enforcing regularization of the latent representations~\cite{Michieli2021SDR}, and generating or crawling data from the internet to replay~\cite{Maracani2021RECALL, PageFortin2022CSSLeveragingLabelsAndRehearsal}. While similarly to CSS methods we explore a continual setting in which the network is sequentially presented with data from the same domain, we do not tackle the class-incremental problem, and instead focus on a closed-set scenario with shifting distribution of classes and scene appearance.
A further important difference is that
while CSS methods assume each adaptation step to be supervised, in our setting no ground-truth labels from the current adaptation stage are available.
\\\textbf{NeRF-based semantic learning.}
Since the introduction of
NeRF~\cite{Mildenhall2020NeRF}, several works have proposed extensions to the framework to incorporate semantic information into the learned scene representation.
\snerf~\cite{Zhi2021SemanticNeRF} first proposed jointly learning appearance, geometry, and semantics through an additional multi-layer perceptron (MLP)
and by adapting the volume rendering equation to produce
semantic
logits.
Subsequent works have further extended this framework along different directions, including combining NeRF with a feature grid and 3D convolutions to achieve generalization~\cite{Vora2022NeSF}, interactively labeling scenes~\cite{Zhi2021iLabel}, performing panoptic segmentation~\cite{Fu2022PanopticNeRF, Kundu2022PanopticNeuralFields}, and using pre-trained Transformer models to supervise few-shot NeRF training~\cite{Jain2021PuttingNeRFOnADiet}, edit scene properties~\cite{Wang2022CLIP-NeRF}, or distill knowledge
for different image-level tasks~\cite{Kobayashi2022FeatureFieldDistillation, Tschernezki2022NeuralFeatureFusionFields}.
In our work, we rely on
\snerf,
which we use to fuse predictions from a segmentation network and that we jointly train with the latter exploiting differentiability. We include the formed scene representation in a long-term memory and use it to render \pls to adapt the segmentation network.
\vspace{-5pt}

\section{Continual Semantic Adaptation}
\subsection{Problem definition}
In our problem setting, which we refer to as \emph{continual semantic adaptation}, we assume we are provided with a segmentation model $f_{\theta_0},$ with parameters $\theta_0$, that was pre-trained on a dataset 
$\mathcal{P}=\left(\mathbf{I}_\mathrm{pre}, \mathbf{S}^\star_\mathrm{pre}\right)$.
Here
$\mathbf{I}_\mathrm{pre}$
is a set of input color images (potentially with associated depth information) and
$\mathbf{S}^\star_\mathrm{pre}$
are the corresponding pixel-wise ground-truth semantic labels. We aim to adapt $f_{\theta_0}$ across a sequence of $N$ scenes $\mathcal{S}_i,\ i\in\{1, \dots, N\}$ for each of which a set $\mathbf{I}_i$ of color (and depth) images,
are collected from different viewpoints, but no ground-truth semantic labels are available. We assume that the input data 
$\left\{\mathbf{I}_\mathrm{pre}, \mathbf{I}_1, \dots, \mathbf{I}_N\right\}$
originate from similar indoor environments (for instance, we do not consider simultaneously synthetic and real-world data) and that the classes to be predicted by the network belong to a closed set and are all known from the pre-training.
For each scene $\mathcal{S}_i,\ i\in\{1,\dots,N\}$, the objective is to find a set of weights $\theta_i$ of the network, starting from $\theta_{i-1}$, such that the performance of $f_{\theta_i}$ on $\mathcal{S}_i$ is higher than that of $f_{\theta_{i-1}}$. Additionally, it is desirable to preserve the performance of $f_{\theta_i}$ on the previous scenes $\left\{\mathcal{S}_1, \dots, \mathcal{S}_{i-1}\right\}$, in other words mitigate catastrophic forgetting.

The proposed setting aims to replicate the scenario of the deployment of a segmentation network on a real-world perception system (for instance a robot, or an augmented reality platform), where multiple sequential experiences are collected across similar scenes, and only limited data of the previous scenes can be stored on an on-board computing unit. During deployment, environments might be revisited over time,
rendering the preservation of previously learned knowledge
essential for a successful deployment.

\subsection{Methodology}
\label{sec:method}

We present a method to address
continual semantic adaptation
in a self-supervised fashion (Fig.~\ref{fig:method_overview}).
In the following,
$\boldsymbol{I}_i^k$ and $\boldsymbol{P}_i^k$ are the \hbox{$k$-th} RGB(-D) image collected in scene $\mathcal{S}_i$ and its corresponding camera pose, where $k\in\{1,\dots,|\mathbf{I}_i|\}$.
We further denote with 
$\mathbf{S}_{\theta}(\boldsymbol{I}^k_i)$
the
prediction produced by $f_\theta$ for $\boldsymbol{I}^k_i$\footnote{Note that in our experiments $f_\theta$ does not use the depth channel of $\boldsymbol{I}^k_i$.
}.
With a slight abuse of notation, we use $\mathbf{S}_\theta(\mathbf{I}_i)$ in place of $\{\mathbf{S}_\theta(\boldsymbol{I}_i^k),\ \boldsymbol{I}_i^k\in\mathbf{I}_i\}$ and similarly for other quantities that are a function of elements in a set.

For each new scene $\mathcal{S}_i$, we train a
Semantic-NeRF~\cite{Zhi2021SemanticNeRF} model $\mathcal{N}_{\phi_i}$, with learnable parameters $\phi_i$, given for each
viewpoint $\boldsymbol{P}_i^k$
the corresponding
semantic label $\mathbf{S}_{\theta_j}(\boldsymbol{I}_i^k)$ predicted by a previous version $f_{\theta_j}$, $j<i$, of the segmentation model.
From the trained \snerf model $\mathcal{N}_{\phi_i}$ we render
semantic \pls 
$\hat{\mathbf{S}}_{\phi_i}$
and images
$\hat{\mathbf{I}}_{\phi_i}$.
The key observation at the root of our self-supervised adaptation is that semantic labels should be \emph{multi-view consistent}, since they are constrained by the scene geometry that defines them. While the predictions of $f$ often do not reflect this constraint because they are produced for each input frame independently, the NeRF-based \pls are by construction multi-view consistent. Inspired by~\cite{Frey2022CLSemanticSegmentation}, we hypothesize that this consistency constitutes an important prior that can be exploited to guide the adaptation of the network to the scene.
Therefore, we use the renderings from $\mathcal{N}_i$
to adapt
the segmentation network
on scene $\mathcal{S}_i$, by minimizing a cross-entropy loss between the \pls and the network predictions.
Crucially, we can use the NeRF and segmentation network predictions to supervise each other, allowing for joint optimization and adaptation of the two networks, which we find further improves the performance of both models.

To continually adapt the segmentation network $f$ to multiple
scenes in a sequence $\mathcal{S}_1\rightarrow\mathcal{S}_2\rightarrow\cdots\rightarrow\mathcal{S}_N$
and prevent catastrophic forgetting, we leverage the compact
representation of NeRF
by storing the corresponding model weights $\phi_i$ after adaptation in a long-term memory for each scene $\mathcal{S}_i$.
Given that a trained NeRF can be queried from any viewpoint, this formulation allows generating for each scene a theoretically infinite number of views for adaptation, at the fixed storage cost given by the size of $\phi_i$. For each previous scene $\mathcal{S}_j$, images $\hat{\mathbf{I}}_{\phi_j}$ and \pls $\hat{\mathbf{S}}_{\phi_j}$ from both previously seen and novel viewpoints
can be rendered and used in an experience replay strategy to mitigate catastrophic forgetting on the previous scenes. 
An overview of our method is shown in Fig.~\ref{fig:method_overview}.
\\\textbf{NeRF-based \pls.}
We
train
for each scene
a NeRF~\cite{Mildenhall2020NeRF} model, which
implicitly learns the geometry and appearance of the environment from a sparse set of posed images and can be used to render photorealistic novel views. 
More specifically, we extend the NeRF formulation by adding a semantic head as in \snerf~\cite{Zhi2021SemanticNeRF}, and we render semantic labels $\hat{\mathbf{S}}_\phi$
by aggregating through the learned density function the semantic-head predictions for $M$ sample points along each camera ray $\mathbf{r}$, as follows: 
\begin{equation}
    \hat{\mathbf{S}}_\phi(\mathbf{r}) = \sum_{i=1}^{M} T_i\alpha_i \mathbf{s}_i,
\end{equation}
where
$\alpha_i = 1-e^{-\sigma_i\delta_i}, T_i = \prod_{j=1}^{i-1}(1-\alpha_j)$,
with $\delta_i$ being the distance between adjacent sample points along the ray, and $\sigma_i$ and $\mathbf{s}_i$ representing the predicted density and semantic logits at the $i$-th sample point along the ray, respectively. 

We observe that if
\snerf
is directly trained
on the labels predicted by a pre-trained segmentation network on a new scene,
the
lack of view consistency of
these
labels
can severely degrade the quality of the learned geometry, which in turn hurts the performance of the rendered semantic labels.
To alleviate the influence of the inconsistent labels on the geometry, we propose to adopt several modifications.
First, we stop the gradient flow from the semantic head into the density head.
Second, we use depth supervision, as introduced in~\cite{Deng2022DepthSupervisedNeRF}, to regularize the depth $\hat{d}(\mathbf{r})= \sum_{i=1}^{N} T_i\alpha_i \delta_i$ rendered by NeRF
via
$\ell_1$ loss with respect to the ground-truth depth $d(\mathbf{r})$:
\begin{align}
\begin{split}
     \mathcal{L}_{\mathrm{d}}(\mathbf{r}) &= \left\lVert  \hat{d}(\mathbf{r}) -  d(\mathbf{r})\right\rVert_1.
     \label{eq:depth_loss}
\end{split}
\end{align}
Through ablations in the Supplementary, we show that this choice is particularly effective at improving the quality of both the geometry and the rendered labels.
Additionally, we note that since the semantic logits $\mathbf{s}_i$ of each sampled point are unbounded, the logits $\hat{\mathbf{S}}_\phi(\mathbf{r})$ of the ray $\mathbf{r}$ can be dominated by a sampled point with very large semantic logits instead of one that is near the surface of the scene.
This could cause the
semantic labels generated
by the NeRF model to overfit the initial labels of the segmentation model and lose multi-view consistency even when the learned geometry is correct.
To address this issue, we instead first apply softmax to the logits of each sampled point, so these are normalized and contribute
to the final aggregated logits
through the weighting induced by volume rendering,
as follows:
\begin{equation}
    \hat{\mathbf{S}}_\phi^\prime(\mathbf{r}) = \sum_{i=1}^{N} T_i\alpha_i \cdot\mathrm{softmax}(\mathbf{s}_i),\
      \hat{\mathbf{S}}_\phi(\mathbf{r}) = \hat{\mathbf{S}}_\phi^\prime(\mathbf{r}) / \lVert\hat{\mathbf{S}}_\phi^\prime(\mathbf{r}) \rVert_1.
\end{equation}
 The final normalized $\hat{\mathbf{S}}_\phi(\mathbf{r})$ is then a categorical distribution $(\hat{S}(\mathbf{r})_1,\cdots,\hat{S}(\mathbf{r})_C)$ over the $C$ semantic classes predicted by NeRF, and we use a negative log-likelihood loss to supervise the rendered semantic
labels
 with the predictions of the semantic network:
\begin{equation}
      \mathcal{L}_{\mathrm{s}}(\mathbf{r}) = - \sum_{c=1}^C \log (\hat{S}(\mathbf{r})_c)\cdot \mathbbm{1}_{c=c(\mathbf{r})},
      \label{eq:semantic_loss}
\end{equation}
where $c(\mathbf{r})$ is the semantic label predicted by the 
segmentation network
$f_\theta$.
We train the NeRF model
by
randomly
sampling rays from the training views and adding
together the losses in~\eqref{eq:depth_loss} and~\eqref{eq:semantic_loss}, as well as the usual $\ell_2$ loss $\mathcal{L}_\mathrm{rgb}(\mathbf{r})$ on the rendered color~\cite{Mildenhall2020NeRF}, as follows:
\begin{equation}
    \mathcal{L} = \sum_{i=1}^{R}  \mathcal{L}_{\mathrm{rgb}}(\mathbf{r_i}) +   w_{\mathrm{d}}\mathcal{L}_{\mathrm{d}}(\mathbf{r_i}) +   w_{\mathrm{s}}\mathcal{L}_{\mathrm{s}}(\mathbf{r_i}),
    \label{eq:full_loss}
\end{equation}
where $R$ is the number of rays sampled for each batch 
and $w_{\mathrm{d}}$, $w_{\mathrm{s}}$ are the weights for the depth loss and the semantic loss, respectively.
After training the NeRF model, we render from it both
color images $\hat{\mathbf{I}}_\phi$ and
semantic labels $\hat{\mathbf{S}}_\phi$,
as \emph{\pls} for
adapting the segmentation network.

Being able to quickly fuse the semantic predictions and generate \pls might be of particular importance in applications that require fast, possibly online adaptation.
To get closer to this objective, 
we adopt the multi-resolution hash encoding proposed in Instant-NGP~\cite{Mueller2022InstantNGP}, which significantly improves the training and rendering speed compared to the original NeRF formulation. 
In the Supplementary,
we compare the quality of the Instant-NGP-based \pls and those obtained with the original implementation from~\cite{Zhi2021SemanticNeRF}, and show that our method is agnostic to the specific NeRF implementation chosen.
\\\textbf{Adaptation through joint 2D-3D training.}
To adapt the segmentation network $f_{\theta_j}$ on a given scene $\mathcal{S}_i$ (where $i>j$), we use the rendered \pls $\hat{\mathbf{S}}_{\phi_i}$ as supervisory signal by optimizing a cross-entropy loss between the network predictions $\mathbf{S}_{\theta_j}$ and $\hat{\mathbf{S}}_{\phi_i}$, similarly to previous approaches in the literature~\cite{Wu2019ACE, Wang2022CoTTA, Frey2022CLSemanticSegmentation}.
However, we propose two important modifications enabled by our particular setup and by its end-to-end differentiability. First,
rather than adapting via the segmentation predictions for the ground-truth input images $\mathbf{I}_i$, we use $\mathbf{S}_{\theta_j}(\hat{\mathbf{I}}_{\phi_i})$, that is, we feed the \textit{rendered} images as input to $f$. This
removes the need for explicitly storing images for later stages, allows the adaptation to use novel viewpoints for which no observations were made, and as we show in our experiments, results in improved performance over the use of ground-truth images.

Second, we
propose to \textit{jointly train} $\mathcal{N}_{\phi_i}$ and $f_{\theta_j}$ by iteratively generating labels from one and back-propagating the cross-entropy loss gradients through the other in each training step. In practice, to initialize the NeRF \pls we first pre-train $\mathcal{N}_{\phi_i}$ with supervision of the ground-truth input images $\mathbf{I}_i$ and of the associated segmentation predictions $\mathbf{S}_{\theta_j}(\mathbf{I}_i)$, and then jointly train $\mathcal{N}_{\phi_i}$ and $f_{\theta_j}$ as described above. We demonstrate the positive influence of this joint adaptation in the experiments, where we show in particular that this 2D-3D knowledge transfer effectively produces improvements in the visual content of both the network predictions and the \pls.
\\\textbf{Continual NeRF-based replay.}
A simple but effective approach to alleviate catastrophic forgetting as the adaptation proceeds across scenes is to
\emph{replay} previous experiences, \ie,
storing the training data of each newly-encountered scene in a memory buffer, and for each subsequent scene, training the segmentation model using both the data from the new scene and those replayed from the buffer, as done for instance in~\cite{Frey2022CLSemanticSegmentation}.
In practice, the size of the replay buffer is often limited due to memory and storage constraints, thus one can only store a subset of the data for replay, resulting in a loss of potentially useful information.
Unlike previous methods that save explicit data into a buffer, we propose storing the NeRF models in a long-term memory. The advantages of this choice are multifold. 
First,
the memory footprint of multiple NeRF models is significantly smaller than that of explicit images and labels (required by~\cite{Frey2022CLSemanticSegmentation}) or
of
the weights of the
segmentation network, stored by~\cite{Wang2022CoTTA}.
Second, since the NeRF model stores both color and semantic information and attains photorealistic fidelity, it can be used to render a theoretically infinite amount of training views at a fixed storage cost (unlike~\cite{Frey2022CLSemanticSegmentation}, which fits semantics in the map, and could not produce photorealistic renderings even if texture was aggregated in 3D). 
Therefore,
the segmentation network can be provided with images rendered from NeRF as input. 
As we show in the experiments, by
rendering
a small set of
views from the NeRF models stored in the long-term memory, our method is able to effectively mitigate catastrophic forgetting.

\section{Experiments}
\label{sec:experiments}

\subsection{Experimental settings}
\paragraph{Dataset.}
We evaluate our proposed method
on the ScanNet~\cite{Dai2017ScanNet} dataset. 
The
dataset
includes $707$ unique indoor scenes, each containing RGB-D images with associated camera poses and manually-generated semantic annotations. 
In all the experiments we resize the images to a resolution of $320\times240$ pixels.
Similarly to~\cite{Frey2022CLSemanticSegmentation}, we use scenes $11$-$707$ in ScanNet to pre-train the 
semantic segmentation network, taking one image every $100$ frames in each of these scenes, for a total of approximately $\num{25000}$ images.
The pre-training dataset is randomly split into a training set of $20\mathrm{k}$ frames and a validation set of $5\mathrm{k}$ frames. 
We use scene $1$-$10$
to adapt
the pre-trained model (cf. Sec.~\ref{sec:pseudolabel_formation},~\ref{sec:one_step_adaptation},~\ref{sec:multi_step_adaptation}); if the dataset contains more than one video sequence for a given scene, we select only the first one.
We select the first $80\%$ of the frames (we refer to them as \emph{training views}) from
each
sequence to generate predictions with the segmentation network and fuse these into a 3D representation, both by training our \snerf model and with the baseline of~\cite{Frey2022CLSemanticSegmentation}.
The last $20\%$ of the frames (\emph{validation views}) are instead used to test the adaptation performance of the semantic segmentation model on the
scene. 
We stress that
this pre-training-training-testing
setup is close to a
real-world
application scenario
of the segmentation model, in which in an initial stage the network is trained offline on a large dataset, then some 
data collected during deployment may be used to adapt the model in an unsupervised fashion, and finally the model performance is tested
during deployment on a different trajectory.
\\\textbf{Networks.}
We use \hbox{DeepLabv3}~\cite{Chen2017DeepLabv3} with a \hbox{ResNet-101}~\cite{He2016ResNet101} backbone as our semantic segmentation network.
To implement our Semantic-NeRF network, we rely on an open-source \hbox{PyTorch} implementation~\cite{torch-ngp} of Instant-NGP~\cite{Mueller2022InstantNGP}.
Further details about the architectures of both networks can be found in the Supplementary.
For brevity, in the following Sections we refer to \snerf as ``NeRF".
\\\textbf{Baselines.}
As
there are no previous works that
explicitly tackle
the continual semantic adaptation problem, we
compare our proposed method to the two
most-closely
related approaches. 
The first one~\cite{Frey2022CLSemanticSegmentation} uses 
per-frame camera pose and depth information to aggregate predictions from a segmentation network into a voxel map and then renders semantic \pls from the map to adapt the network. 
We
implement the method
using
the
framework of~\cite{Schmid2022PanopticMultiTSDFs}
and
use a voxel resolution of \SI{5}{cm}, as done in~\cite{Frey2022CLSemanticSegmentation}, which yields a 
total map
size comparable to the memory footprint of the NeRF parameters (cf. Supplementary for further details).
The second approach, CoTTA~\cite{Wang2022CoTTA}, focuses on continual test-time domain adaptation
and proposes a
student-teacher framework with label augmentation and stochastic weight restoration to gradually adapt the semantic segmentation model while keeping the knowledge on the source domain.
We use the official 
open-source
implementation,
which we adapt to test its performance on the proposed setting. \\\textbf{Metric.}
For all the experiments, we report mean intersection over union ($\mathrm{mIoU}$, in percentage values) as a metric.

\subsection{Pre-training of the segmentation network}
We pre-train
DeepLab
for $150$ epochs
to minimize
the cross-entropy loss with respect to the ground-truth labels $\mathbf{S}^\star_\mathrm{pre}$.
We
apply common data augmentation techniques, 
including
random flipping/orientation
and color jitter.
After pre-training, we
select
the model with best performance on the validation set
for
adaptation to the new scenes. 

\subsection{Pseudo-label formation}
\label{sec:pseudolabel_formation}
\begin{table}
\centering
\footnotesize
\setlength\tabcolsep{2pt}
\resizebox{0.95\linewidth}{!}{
\begin{tabular}{@{}ccccc@{}}
\toprule
& Pre-train & Mapping~\cite{Frey2022CLSemanticSegmentation} & Ours & Ours Joint Training \\ \midrule
Scene $1$ & 41.1     & 48.9       &   48.8{\scriptsize$\pm$0.7}      & \textbf{54.8}{\scriptsize$\pm$1.8}\\
Scene $2$ & 35.5     & 33.9       &   36.2{\scriptsize$\pm$0.8}       & \textbf{38.3}{\scriptsize$\pm$0.4}\\
Scene $3$ & 23.5     & 25.1       &   \textbf{27.1}{\scriptsize$\pm$0.9}      & 26.4{\scriptsize$\pm$1.8}                       \\
Scene $4$ & 62.8     & \textbf{65.3}       &   62.9{\scriptsize$\pm$0.5}      & 65.0{\scriptsize$\pm$1.1}                       \\
Scene $5$ & 49.8     & 49.3       &   \textbf{55.5}{\scriptsize$\pm$1.3}       & 46.6{\scriptsize$\pm$0.2}                       \\
Scene $6$ & 48.9     & \textbf{51.7}       &   50.4{\scriptsize$\pm$0.4}      & 50.9{\scriptsize$\pm$0.4}                          \\
Scene $7$ & 39.7     & 41.2       &   40.4{\scriptsize$\pm$0.5}       & \textbf{41.7}{\scriptsize$\pm$2.0}                       \\
Scene $8$ & 31.6     & 34.8       &   34.0{\scriptsize$\pm$0.4}       & \textbf{39.0}{\scriptsize$\pm$4.6}                          \\
Scene $9$ & 31.7     & 33.8       &   \textbf{35.6}{\scriptsize$\pm$0.4}      & 31.3{\scriptsize$\pm$0.4}                        \\
Scene $10$ & 52.5     & 55.8       &   \textbf{56.4}{\scriptsize$\pm$0.6}      & 56.2{\scriptsize$\pm$1.0}\\ \midrule
Average & 41.7     & 44.0       &   44.7{\scriptsize$\pm$0.7}     & \textbf{45.0}{\scriptsize$\pm$1.4}\\\bottomrule                       
\end{tabular}}
\caption{Pseudo-label
performance averaged over the training views and
$10$ different seeds for
Ours
\pls. ``Pre-train" denotes the performance of the segmentation model $f_{\theta_0}$.}
\label{tab:pl}
\vspace{-4ex}
\end{table}
We train the
NeRF
network by minimizing~\eqref{eq:full_loss} for $60$ epochs using the training views.
While with our method
we can render
\pls from
any viewpoint,
to allow a controlled comparison against~\cite{Frey2022CLSemanticSegmentation} in Sec.~\ref{sec:one_step_adaptation} and~\ref{sec:multi_step_adaptation},
we generate the \pls from our NeRF model using
the same training viewpoints.
While the \pls of~\cite{Frey2022CLSemanticSegmentation} are deterministic, to account for the stochasticity of 
NeRF,
we run our method with $10$ different random seeds and report the mean and variance over these.
As shown in Tab.~\ref{tab:pl}, the \pls produced by our method outperform on average those of~\cite{Frey2022CLSemanticSegmentation}.
A further improvement can be obtained
by
jointly training NeRF and
the
DeepLab
model,
which we discuss in the next Section.

\subsection{One-step adaptation}
\label{sec:one_step_adaptation}
\begin{figure*}[t]
\centering
\def\colwidth{0.108\textwidth}
\def\extraspacewidth{0.005\textwidth}
\newcolumntype{M}[1]{>{\centering\arraybackslash}m{#1}}
\addtolength{\tabcolsep}{-4pt}
\vspace{-5pt}
\begin{tabular}{M{\colwidth} M{\colwidth} M{\colwidth} M{\extraspacewidth} M{\colwidth}  M{\colwidth} M{\colwidth} M{\extraspacewidth} M{\colwidth} M{\colwidth}}
 \multicolumn{3}{c}{NeRF \pls} & \hfill & \multicolumn{3}{c}{Segmentation network
 predictions} & \hfill & \multicolumn{2}{c}{Ground-truth}  \\
 \cmidrule(r){1-3} \cmidrule(r){5-7} \cmidrule(r){9-10} 
 \small Epoch $0$ & \small Epoch $10$ & \small Epoch $50$ & \hfill & \small Epoch $0$ & \small Epoch $10$ & \small Epoch $50$ & \hfill & \small Images & \small Labels \tabularnewline
\includegraphics[width=\linewidth]{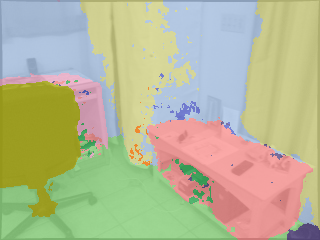} & 
\includegraphics[width=\linewidth]{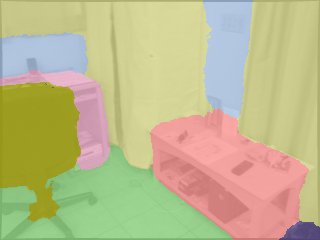} &
\includegraphics[width=\linewidth]{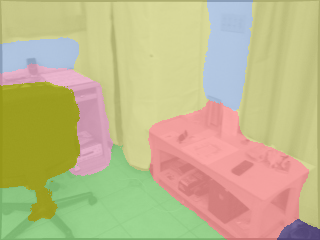} &
\hfill &
\includegraphics[width=\linewidth]{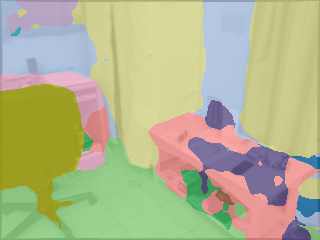} &
\includegraphics[width=\linewidth]{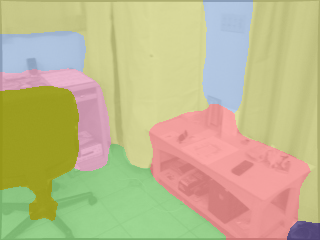} &
\includegraphics[width=\linewidth]{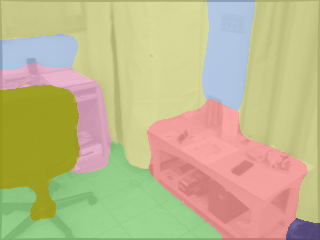} & \hfill & \includegraphics[width=\linewidth]{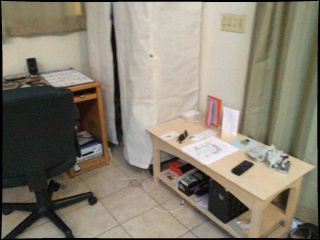} & \includegraphics[width=\linewidth]{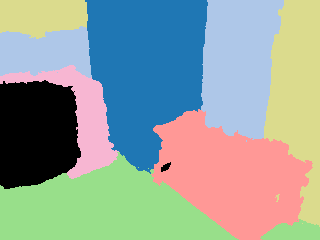} \tabularnewline
\includegraphics[width=\linewidth]{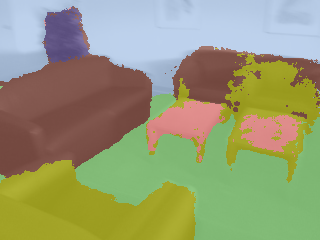} &
\includegraphics[width=\linewidth]{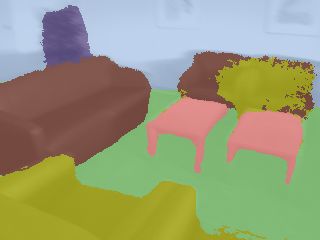} &
\includegraphics[width=\linewidth]{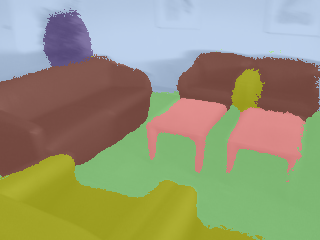} &
\hfill &
\includegraphics[width=\linewidth]{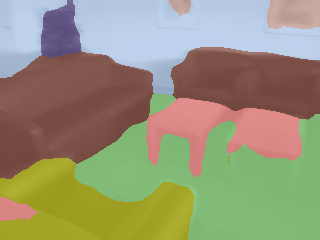} &
\includegraphics[width=\linewidth]{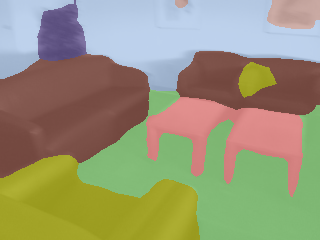} &
\includegraphics[width=\linewidth]{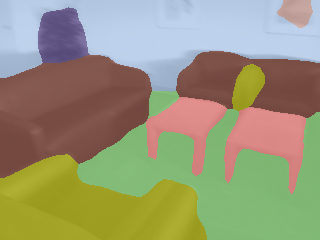} &
\hfill & \includegraphics[width=\linewidth]{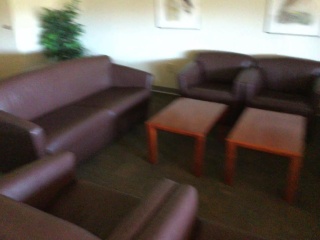} & \includegraphics[width=\linewidth]{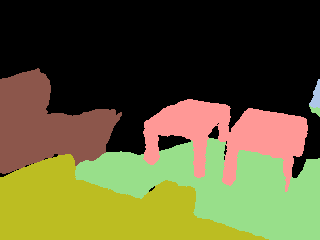} \tabularnewline
\includegraphics[width=\linewidth]{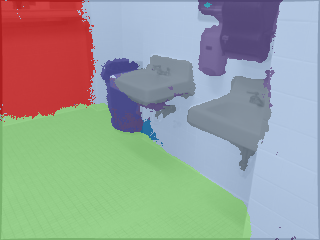} &
\includegraphics[width=\linewidth]{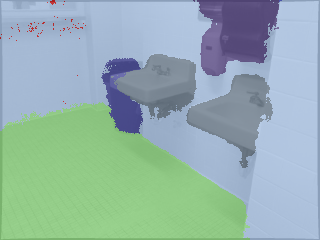} &
\includegraphics[width=\linewidth]{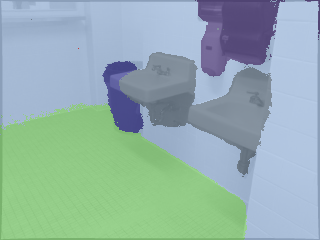} &
\hfill &
\includegraphics[width=\linewidth]{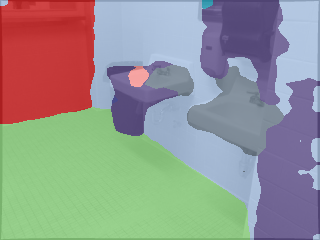} &
\includegraphics[width=\linewidth]{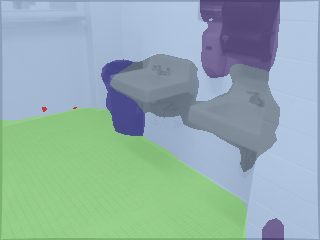} &
\includegraphics[width=\linewidth]{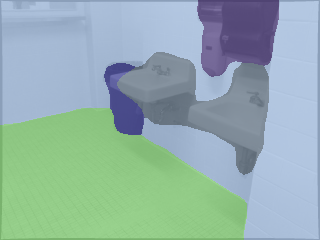} &
\hfill & \includegraphics[width=\linewidth]{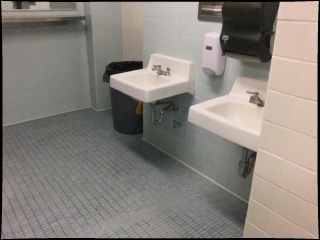} & \includegraphics[width=\linewidth]{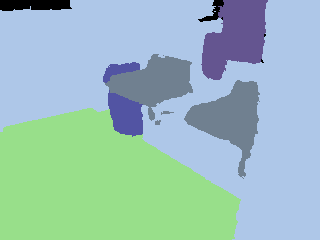} \tabularnewline
\end{tabular}

\addtolength{\tabcolsep}{4pt}
\vspace{-1ex}
\caption{Effect of joint training over the \pls and the
predictions of the segmentation network (DeepLab). Color-coded labels are overlaid on the corresponding color images. Black pixels in the ground-truth labels denote missing annotation. First scene:
The noisy predictions of DeepLab are corrected and the segmentation results conform much better to the geometry of the scene. Second scene:
The geometric details can be better recovered even for the legs of the table. Third scene: By enforcing multi-view consistency,
the initial wrong predictions on the wall are corrected through the predictions from other views. Note that
the obtained labels adhere accurately to the scene geometry, often even better than in the ground-truth annotations.
\label{fig:joint_training_effect}}
\vspace{-15pt}
\end{figure*}

As a first adaptation experiment, we evaluate the performance of the different methods when letting the segmentation network $f_{\theta_0}$ adapt in a single stage to each of the scenes $1$-$10$. This setup is similar to that of one-stage
UDA, and we thus
compare to
the state-of-the-art method CoTTA~\cite{Wang2022CoTTA}. 

We evaluate our method in two different settings.
In the first one, which we refer to as \emph{fine-tuning}, we simply use the \pls rendered as in Sec.~\ref{sec:pseudolabel_formation} to adapt the segmentation network through cross-entropy loss on its predictions. In the second one, we \emph{jointly train} 
NeRF and DeepLab
via
iterative mutual supervision. For a fair comparison, in both settings we
optimize
the pre-trained NeRF
for
the same number of
additional epochs, while maintaining
supervision through color and depth images. In fine-tuning, we
perform
NeRF pre-training
for $60$ epochs, according to
Sec.~\ref{sec:pseudolabel_formation}. In joint training, we instead first pre-train NeRF for $10$ epochs, and then train
NeRF concurrently with DeepLab for $50$ epochs. 
We run each method $10$ times and report mean and standard deviation across the runs.
Given that the baselines do not support generating images from novel viewpoints, both in fine-tuning and in joint training we use images from the training viewpoints as input to DeepLab.
Additionally, since our method allows \emph{rendering} images, we evaluate the difference between feeding ground-truth images vs. NeRF renderings from the same viewpoints to DeepLab.

Table~\ref{tab:1_step} presents the adaptation performance of the different methods on the validation views, which provides a measure of the knowledge transfer induced
within the scene by the self-training. 
Since our method is unsupervised, in the Supplementary we additionally report the improvement in performance on the training views, which is indicative of the effectiveness of the self-supervised adaptation and is of practical utility for real-world deployment scenarios where a scene might be revisited from similar viewpoints.
\\\indent
As shown in Tab.~\ref{tab:1_step}, fine-tuning with our \pls results in improved performance compared to the pre-trained model, and outperforms both baselines for most of the scenes. Interestingly, using rendered images ($\mathrm{NI}+\mathrm{NL}$)
consistently produces better results than fine-tuning with the ground-truth images ($\mathrm{GI}+\mathrm{NL}$). We hypothesize that
this is due to the
small image artifacts introduced by the NeRF rendering 
acting
as an augmentation mechanism.
We further observe that the 
$\mathrm{mIoU}$ can vary largely across the
scenes. This
can be explained with
the variability in the room types and lighting conditions,
which is also reflected in the scenes with more
extreme
illumination (and hence more challenging for NeRF to reconstruct the geometry) having a larger variance with our approach.
However, the main observation is
that jointly training NeRF and DeepLab (using rendered images as input) results in remarkably better adaptation on almost all the scenes.
This improvement can be attributed to the positive knowledge transfer induced between the frame-level predictions of DeepLab and the 3D-aware NeRF \pls.
As shown in Fig.~\ref{fig:joint_training_effect}, this strategy allows effectively resolving local artifacts in the NeRF \pls through the smoothing effect of the DeepLab
labels, while at the same time addressing inconsistencies in the per-frame outputs of the segmentation network
due to
its lack of view consistency.

\vspace{-5pt}
\subsection{Multi-step adaptation}
\label{sec:multi_step_adaptation}
\begin{table*}[ht]
\vspace{-5pt}
\centering
\resizebox{0.91\linewidth}{!}{
\begin{tabular}{@{}ccccccc@{}}
\toprule
        & Pre-train & CoTTA~\cite{Wang2022CoTTA}                     & Fine-tuning ($\mathrm{GI}+\mathrm{ML}$) & Ours Fine-tuning ($\mathrm{GI}+\mathrm{NL}$) & Ours Fine-tuning ($\mathrm{NI}+\mathrm{NL}$) & Ours Joint Training            \\ \midrule
Scene $1$ & 43.9     & 44.0{\scriptsize$\pm$0.0} & 46.3{\scriptsize$\pm$0.3}              &     46.2{\scriptsize$\pm$1.0}                                     &   47.1{\scriptsize$\pm$1.0}                                    & \textbf{50.0}{\scriptsize$\pm$1.3} \\ 
Scene $2$ & 41.3     & 41.2{\scriptsize$\pm$0.0} & 39.4{\scriptsize$\pm$0.3}              &       39.5{\scriptsize$\pm$1.0}                                   &   44.2{\scriptsize$\pm$1.0}                                     & \textbf{47.1}{\scriptsize$\pm$1.2} \\ 
Scene $3$ & \textbf{23.0}     & 22.8{\scriptsize$\pm$0.0} & 21.6{\scriptsize$\pm$0.1}              &   21.9{\scriptsize$\pm$0.7}                                       &      21.5{\scriptsize$\pm$1.0}                                                     & 19.9{\scriptsize$\pm$2.3} \\ 
Scene $4$ & 50.2     & 50.3{\scriptsize$\pm$0.0} & 52.4{\scriptsize$\pm$0.2}              &         51.5{\scriptsize$\pm$0.5}                                 &         52.8{\scriptsize$\pm$0.8}                                     & \textbf{53.7}{\scriptsize$\pm$2.4} \\ 
Scene $5$ & 40.1     & 40.1{\scriptsize$\pm$0.0} & 49.4{\scriptsize$\pm$0.5}              &     50.6{\scriptsize$\pm$2.4}                                     &     \textbf{52.8}{\scriptsize$\pm$2.9}                                        & 42.7{\scriptsize$\pm$1.0} \\ 
Scene $6$ & 37.6     & 37.6{\scriptsize$\pm$0.0} & 33.7{\scriptsize$\pm$0.3}              &    36.2{\scriptsize$\pm$1.6}                                      &         37.1{\scriptsize$\pm$2.4}                                    & \textbf{40.8}{\scriptsize$\pm$1.2} \\ 
Scene $7$ & 55.8     & 55.9{\scriptsize$\pm$0.0} & 50.7{\scriptsize$\pm$0.5}              &     50.7{\scriptsize$\pm$1.8}                                     &         52.1{\scriptsize$\pm$1.3}                                & \textbf{56.5}{\scriptsize$\pm$4.8} \\ 
Scene $8$ & \textbf{27.9}     & \textbf{27.9}{\scriptsize$\pm$0.0} & 24.7{\scriptsize$\pm$0.2}              &   23.8{\scriptsize$\pm$0.4}                                       &            25.3{\scriptsize$\pm$0.8}                                          & 25.7{\scriptsize$\pm$2.9} \\ 
Scene $9$ & 54.9     & 54.9{\scriptsize$\pm$0.0} & 62.2{\scriptsize$\pm$1.3}              &   57.6{\scriptsize$\pm$5.3}                                       &        52.1{\scriptsize$\pm$2.7}                                              & \textbf{63.7}{\scriptsize$\pm$3.3} \\ 
Scene $10$ & 73.5     & 73.5{\scriptsize$\pm$0.0} & \textbf{73.8}{\scriptsize$\pm$0.2}              &        \textbf{73.8}{\scriptsize$\pm$0.2}                                  &           73.5{\scriptsize$\pm$0.4}                                           & 73.7{\scriptsize$\pm$0.5} \\ 
\midrule
Average & 44.8     & 44.8{\scriptsize$\pm$0.0} & 45.4{\scriptsize$\pm$0.4}              &     45.2{\scriptsize$\pm$1.5}                                     &       45.9{\scriptsize$\pm$1.4}                                               & \textbf{47.4}{\scriptsize$\pm$2.1} \\ \bottomrule
\end{tabular}}
\caption{Performance of the segmentation network on the validation set of each scene after one-step adaptation. $\mathrm{GI}$ and $\mathrm{NI}$ denote respectively ground-truth color images and NeRF-rendered color images. $\mathrm{ML}$ and $\mathrm{NL}$ indicate adaptation using \pls formed respectively with the method of~\cite{Frey2022CLSemanticSegmentation} and with our
approach. In joint training, we use NeRF-based renderings and \pls.}
\label{tab:1_step}
\end{table*}

\begin{table*}[ht]
\centering
\resizebox{0.91\linewidth}{!}{
\begin{tabular}{llccccccccccc}
\toprule
& & Step $1$ & Step $2$ & Step $3$ & Step $4$ & Step $5$ & Step $6$ & Step $7$ & Step $8$ & Step $9$ & Step $10$ & Average \\ \midrule
$\mathrm{Pre}{\text-}\mathrm{train}$ & & 43.9 & 41.3 & 23.0 & 50.2 & 40.1 & 37.6 & 55.8 & \textbf{27.9} & 54.9 & 73.5 & 44.8 \\
\arrayrulecolor{black!20}\specialrule{0.2pt}{0.2pt}{0.2pt}
\multirow{5}{*}{$\mathrm{Adapt}$} & CoTTA~\cite{Wang2022CoTTA} & 44.0{\scriptsize$\pm$0.0} & 40.9{\scriptsize$\pm$0.0} & 22.7{\scriptsize$\pm$0.0} & 50.2{\scriptsize$\pm$0.1} & 40.0{\scriptsize$\pm$0.0} & 37.5{\scriptsize$\pm$0.0} & 56.0{\scriptsize$\pm$0.1} & 26.9{\scriptsize$\pm$0.0} & 54.5{\scriptsize$\pm$0.0} & \textbf{73.8}{\scriptsize$\pm$0.0} & 44.7{\scriptsize$\pm$0.0} \\
& Mapping~\cite{Frey2022CLSemanticSegmentation} & 46.8{\scriptsize$\pm$0.4} & 42.1{\scriptsize$\pm$2.0} & 23.6{\scriptsize$\pm$0.7} & \textbf{50.6}{\scriptsize$\pm$2.6} & \textbf{44.0}{\scriptsize$\pm$0.1} & 35.8{\scriptsize$\pm$0.5} & 56.7{\scriptsize$\pm$1.3} & 26.5{\scriptsize$\pm$1.8} & 68.3{\scriptsize$\pm$1.4} & 72.7{\scriptsize$\pm$1.0} & 46.7{\scriptsize$\pm$1.2}\\
&
Ours ($\mathbf{I}_\textrm{pre}$ replay only)
& 53.3{\scriptsize$\pm$0.7} & \textbf{48.0}{\scriptsize$\pm$2.4} & 20.5{\scriptsize$\pm$0.1} & 49.0{\scriptsize$\pm$1.5} & 43.4{\scriptsize$\pm$0.0} & 39.0{\scriptsize$\pm$1.4} & \textbf{62.1}{\scriptsize$\pm$6.2} & 26.7{\scriptsize$\pm$3.0} & 65.7{\scriptsize$\pm$5.6} & 73.0{\scriptsize$\pm$0.5} & \textbf{48.1}{\scriptsize$\pm$2.1} \\
& Ours & 53.7{\scriptsize$\pm$1.3} &  46.3{\scriptsize$\pm$0.7} & \textbf{24.3}{\scriptsize$\pm$2.0} & 49.1{\scriptsize$\pm$0.9} & 43.7{\scriptsize$\pm$0.3} & \textbf{40.4}{\scriptsize$\pm$1.5} & 55.8{\scriptsize$\pm$0.8} & 26.2{\scriptsize$\pm$0.9} & \textbf{68.9}{\scriptsize$\pm$3.2} & 72.5{\scriptsize$\pm$1.6} & \textbf{48.1}{\scriptsize$\pm$1.3} \\
& {Ours (novel viewpoints)}
& \textbf{53.8}{\scriptsize$\pm$0.4} &  46.7{\scriptsize$\pm$2.1} & 23.2{\scriptsize$\pm$3.3} & 49.0{\scriptsize$\pm$1.0} & 42.9{\scriptsize$\pm$0.4} & 40.1{\scriptsize$\pm$0.7} & 58.0{\scriptsize$\pm$8.5} & 23.2{\scriptsize$\pm$2.0} & 66.7{\scriptsize$\pm$7.1} & 71.5{\scriptsize$\pm$2.2} &47.5{\scriptsize$\pm$2.8}

\\

\arrayrulecolor{black}
\midrule
\multirow{5}{*}{$\mathrm{Previous}$} & CoTTA~\cite{Wang2022CoTTA} & $-$ & 44.0{\scriptsize$\pm$0.0} & 42.2{\scriptsize$\pm$0.0} & 35.6{\scriptsize$\pm$0.0} & 39.3{\scriptsize$\pm$0.0} & 39.4{\scriptsize$\pm$0.0} & 39.1{\scriptsize$\pm$0.0} & 41.5{\scriptsize$\pm$0.0} & 39.7{\scriptsize$\pm$0.0} & 41.3{\scriptsize$\pm$0.0} & 40.2{\scriptsize$\pm$0.0} \\
& Mapping~\cite{Frey2022CLSemanticSegmentation} & $-$ & 46.5{\scriptsize$\pm$0.1} & 42.8{\scriptsize$\pm$1.0} & 37.3{\scriptsize$\pm$0.9} & 40.4{\scriptsize$\pm$0.6} & 40.9{\scriptsize$\pm$0.7} & 39.9{\scriptsize$\pm$1.1} & 42.2{\scriptsize$\pm$0.5} & 40.0{\scriptsize$\pm$0.4} & 42.8{\scriptsize$\pm$0.7} & 41.4{\scriptsize$\pm$0.7} \\
&
Ours ($\mathbf{I}_\textrm{pre}$ replay only)
& $-$ & 52.3{\scriptsize$\pm$0.3} & 47.5{\scriptsize$\pm$1.1} & 38.6{\scriptsize$\pm$1.1} & 40.8{\scriptsize$\pm$0.7} & 42.4{\scriptsize$\pm$0.3} & 41.5{\scriptsize$\pm$0.7} & 44.3{\scriptsize$\pm$1.4} & 41.4{\scriptsize$\pm$0.6} & 43.9{\scriptsize$\pm$0.9} & 43.6{\scriptsize$\pm$0.8}\\
& Ours & $-$& 53.2{\scriptsize$\pm$0.9} & 48.2{\scriptsize$\pm$0.8} & 41.5{\scriptsize$\pm$0.8} & 42.8{\scriptsize$\pm$0.8} & 43.2{\scriptsize$\pm$0.8} & 42.2{\scriptsize$\pm$0.8} & 44.1{\scriptsize$\pm$0.2} & \textbf{41.7}{\scriptsize$\pm$0.2} & \textbf{44.3}{\scriptsize$\pm$0.3} & 44.6{\scriptsize$\pm$0.6} \\
& {Ours (novel viewpoints)} & $-$& \textbf{54.8}{\scriptsize$\pm$0.9} & \textbf{50.4}{\scriptsize$\pm$2.1} & \textbf{41.8}{\scriptsize$\pm$0.9} & \textbf{43.8}{\scriptsize$\pm$0.8} & \textbf{43.4}{\scriptsize$\pm$0.9} & \textbf{42.7}{\scriptsize$\pm$1.0} & \textbf{44.8}{\scriptsize$\pm$0.9} & 41.6{\scriptsize$\pm$0.7} & \textbf{44.3}{\scriptsize$\pm$0.2} & \textbf{45.3}{\scriptsize$\pm$0.9}\\
\bottomrule
\end{tabular}}
\caption{Multi-step performance evaluated on the validation set of each scene. At Step $i$, $\mathrm{Pre}{\text-}\mathrm{train}$ and $\mathrm{Adapt}$ denote respectively the performance of the pre-trained network $f_{\theta_0}$ and of the adapted network $f_{\theta_i}$ on the current scene $\mathcal{S}_i$, while $\mathrm{Previous}$ represents the average performance of $f_{\theta_i}$ on scenes $\mathcal{S}_1$ to $\mathcal{S}_{i-1}$. All Ours are with \emph{joint training}.
Our baseline with novel viewpoints used for replay ($\mathrm{Ours\ (novel\ viewpoints)}$) is able to consistently retain knowledge better than the other methods.
}
\label{tab:multi_step_novel_viewpoints}
\vspace{-15pt}
\end{table*}

\vspace{-5pt}
To evaluate our method in the full scenario of continual semantic adaptation, we perform multi-step adaptation across scenes $1$-$10$, where in the $i$-th
step
the segmentation network $f_{\theta_{i-1}}$ gets adapted on scene $\mathcal{S}_i$, resulting in $f_{\theta_i}$, and the NeRF model $\mathcal{N}_i$ is added to the long-term memory
at the end of the stage. 
For steps $i\in\{2,\dots, 10\}$, to counteract forgetting on the previous scenes we render images and \pls
for each of the $\mathcal{N}_j$ models ($1\le j\le i-1$) in the long-term memory.
In practice,
we construct
a memory buffer of fixed size 
$100$,
to which
at stage $i$ each of the previous models
$\mathcal{N}_j$
contribute equally with 
images $\hat{\mathbf{I}}_\mathrm{buf}$ and
\pls $\hat{\mathbf{S}}_\mathrm{buf}$
rendered
from $\lfloor100/(i-1)\rfloor$
randomly
chosen
training views.
Following~\cite{Frey2022CLSemanticSegmentation},
we additionally randomly select $10\%$ of the pre-training data and combine them to the data from the previous scenes,
which acts as
prior knowledge and prevents the model from overfitting to the new scenes and losing its generalization performance.
This has a similar effect to the regularization scheme used by CoTTA~\cite{Wang2022CoTTA} to preserve previous knowledge, namely storing the network parameters for the initial pre-trained model and the teacher network.
Note that
both the size of our memory buffer ($\SI{14}{MB}$) and that of the replayed pre-training data ($\SI{65}{MB}$) are
much smaller than
the size
of two sets of DeepLab weights ($2\times\SI{225}{MB}$), so our method actually requires less storage space than CoTTA~\cite{Wang2022CoTTA}. A detailed analysis of the memory footprint of the different approaches is presented in the Supplementary;
we show in particular that since our method is agnostic to the specific NeRF implementation, 
with the slower but lighter implementation of \snerf~\cite{Zhi2021SemanticNeRF} the storage comparison is in our favor up to 
$90$ scenes. We deem this to be a realistic margin for real-world deployment scenarios (\eg,
it is hardly the case that an agent sequentially visits more than a few scenes during the same mission).
For the baseline of~\cite{Frey2022CLSemanticSegmentation} we use the same setup as our method, but 
with
mapping-based \pls and ground-truth images in the memory buffer, due to its inability to generate
images.
For a fair comparison, we use \emph{training} views for replay also for our method.
The latter, however,
also allows generating data from \emph{novel} viewpoints for replay; 
very
interestingly, we find this to yield better knowledge retention (cf. Tab.~\ref{tab:multi_step_novel_viewpoints} and for a detailed discussion Sec.~\ref{sec:appendix_replay_novel_viewpoints} of the Supplementary).
\\\indent
The multi-step adaptation results are shown in Tab.~\ref{tab:multi_step_novel_viewpoints}, where for each method the mean and standard deviation across $3$ runs are reported. 
To better show the effect of NeRF-based replay, we also run our adaptation method with only replay from the pre-training dataset, without replaying from the old NeRF models ($\mathrm{Ours\ (}\mathbf{I}_\textrm{pre}\mathrm{\ replay\ only)}$).
Our method achieves the best average adaptation performance ($\mathrm{Adapt}$) across the new scenes in the multi-step setting,
improving
by
$\sim\num{3}\%\ \mathrm{mIoU}$
over the pre-trained model. Note that this improvement
is
consistent with the
one observed in
one-step adaptation (Tab.~\ref{tab:1_step}), which 
validates that our method can successfully adapt across multiple scenes,
without the performance dropping after
a specific number of
steps.
At the same time, while
NeRF-based replay of the old scenes
on average does not induce
a positive forward transfer in the adaptation to the new scenes ($\mathrm{Adapt}$),
its usage can
significantly
alleviate 
forgetting
compared to the case with no replay.
As a result, when using NeRF-based replay,
our method
is able to maintain
in almost all the adaptation steps the best average performance over the previous scenes ($\mathrm{Previous}$).
Further in-detail results for each scene and after each adaptation step are reported in the Supplementary.

\vspace{-2ex}

\section{Conclusion}
\label{sec:conclusions}
\vspace{-1ex}
In this work, we present a novel approach for unsupervised continual adaptation of a semantic segmentation network to multiple novel scenes using neural rendering. 
We exploit the fact that the
new
scenes are observed from multiple viewpoints
and jointly train
in each scene
a \snerf model and
the
segmentation network.
We show
that the induced 2D-3D knowledge transfer
results in improved unsupervised adaptation performance compared to state-of-the-art methods.
We further propose a
NeRF-based replay strategy which allows efficiently mitigating catastrophic forgetting
and enables
rendering
a potentially infinite number of
images
for adaptation
at constant storage cost. We believe this opens up interesting avenues for replay-based adaptation, particularly for
use
on real-world perception systems, which can compactly store collected experiences on board and generate past data as needed.
We discuss the limitations of our method in the Supplementary.

{\small
\\\textbf{Acknowledgements:} This work has
received funding from
the European Union's Horizon 2020 research and innovation program under grant agreement No. 101017008 (Harmony),
the Max Planck ETH Center for Learning Systems, and the HILTI group. We thank Marco Hutter for his guidance and support.
}

{\small
\bibliographystyle{ieee_fullname}
\bibliography{egbib}

\begin{thebibliography}{10}\itemsep=-1pt

\bibitem{Cermelli2020MiB}
Fabio Cermelli, Massimiliano Mancini, Samuel~Rota Bul{\'o}, Elisa Ricci, and
  Barbara Caputo.
\newblock
  \href{https://openaccess.thecvf.com/content_CVPR_2020/html/Cermelli_Modeling_the_Background_for_Incremental_Learning_in_Semantic_Segmentation_CVPR_2020_paper.html}{Modeling
  the Background for Incremental Learning in Semantic Segmentation}.
\newblock In {\em CVPR}, 2020.

\bibitem{Chen2017DeepLabv3}
Liang-Chieh Chen, George Papandreou, Florian Schroff, and Hartwig Adam.
\newblock \href{https://arxiv.org/abs/1706.05587}{Rethinking Atrous Convolution
  for Semantic Image Segmentation}.
\newblock {\em CoRR:1706.05587}, 2017.

\bibitem{Chen2019CrDoCo}
Yun-Chun Chen, Yen-Yu Lin, Ming-Hsuan Yang, and Jia-Bin Huang.
\newblock
  \href{https://openaccess.thecvf.com/content_CVPR_2019/html/Chen_CrDoCo_Pixel-Level_Domain_Transfer_With_Cross-Domain_Consistency_CVPR_2019_paper.html}{CrDoCo:
  Pixel-Level Domain Transfer With Cross-Domain Consistency}.
\newblock In {\em CVPR}, 2019.

\bibitem{Chen2017NoMoreDiscrimination}
Yi-Hsin Chen, Wei-Yu Chen, Yu-Ting Chen, Bo-Cheng Tsai, Yu-Chiang~Frank Wang,
  and Min Sun.
\newblock
  \href{https://openaccess.thecvf.com/content_iccv_2017/html/Chen_No_More_Discrimination_ICCV_2017_paper.html}{No
  More Discrimination: Cross City Adaptation of Road Scene Segmenters}.
\newblock In {\em ICCV}, 2017.

\bibitem{Choi2019SelfEnsemblingGANBased}
Jaehoon Choi, Taekyung Kim, and Changick Kim.
\newblock
  \href{https://openaccess.thecvf.com/content_ICCV_2019/html/Choi_Self-Ensembling_With_GAN-Based_Data_Augmentation_for_Domain_Adaptation_in_Semantic_ICCV_2019_paper.html}{Self-Ensembling
  With GAN-Based Data Augmentation for Domain Adaptation in Semantic
  Segmentation}.
\newblock In {\em ICCV}, 2019.

\bibitem{Cordts2016Cityscapes}
Marius Cordts, Mohamed Omran, Sebastian Ramos, Timo Rehfeld, Markus Enzweiler,
  Rodrigo Benenson, Uwe Franke, Stefan Roth, and Bernt Schiele.
\newblock
  \href{https://openaccess.thecvf.com/content_cvpr_2016/html/Cordts_The_Cityscapes_Dataset_CVPR_2016_paper.html}{The
  Cityscapes Dataset for Semantic Urban Scene Understanding}.
\newblock In {\em CVPR}, 2016.

\bibitem{Dai2017ScanNet}
Angela Dai, Angel~X. Chang, Manolis Savva, Maciej Halber, Thomas Funkhouser,
  and Matthias Nie{\ss}ner.
\newblock
  \href{https://openaccess.thecvf.com/content_cvpr_2017/html/Dai_ScanNet_Richly-Annotated_3D_CVPR_2017_paper.html}{ScanNet:
  Richly-Annotated 3D Reconstructions of Indoor Scenes}.
\newblock In {\em CVPR}, 2017.

\bibitem{Deng2022DepthSupervisedNeRF}
Kangle Deng, Andrew Liu, Jun-Yan Zhu, and Deva Ramanan.
\newblock
  \href{https://openaccess.thecvf.com/content/CVPR2022/html/Deng_Depth-Supervised_NeRF_Fewer_Views_and_Faster_Training_for_Free_CVPR_2022_paper.html}{Depth-supervised
  NeRF: Fewer Views and Faster Training for Free}.
\newblock In {\em CVPR}, 2022.

\bibitem{Diaz2018DontForget}
Natalia D{\'\i}az-Rodr{\'\i}guez, Vincenzo Lomonaco, David Filliat, and Davide
  Maltoni.
\newblock \href{https://arxiv.org/abs/1810.13166}{Don't forget, there is more
  than forgetting: new metrics for Continual Learning}.
\newblock In {\em NeurIPS Workshop}, 2018.

\bibitem{Douillard2021PLOP}
Arthur Douillard, Yifu Chen, Arnaud Dapogny, and Matthieu Cord.
\newblock
  \href{https://openaccess.thecvf.com/content/CVPR2021/html/Douillard_PLOP_Learning_Without_Forgetting_for_Continual_Semantic_Segmentation_CVPR_2021_paper.html}{PLOP:
  Learning without Forgetting for Continual Semantic Segmentation}.
\newblock In {\em CVPR}, 2021.

\bibitem{Du2019SSF-DAN}
Liang Du, Jingang Tan, Hongye Yang, Jianfeng Feng, Xiangyang Xue, Qibao Zheng,
  Xiaoqing Ye, and Xiaolin Zhang.
\newblock
  \href{https://openaccess.thecvf.com/content_ICCV_2019/html/Du_SSF-DAN_Separated_Semantic_Feature_Based_Domain_Adaptation_Network_for_Semantic_ICCV_2019_paper.html}{SSF-DAN:
  Separated Semantic Feature Based Domain Adaptation Network for Semantic
  Segmentation}.
\newblock In {\em ICCV}, 2019.

\bibitem{Frey2022CLSemanticSegmentation}
Jonas Frey, Hermann Blum, Francesco Milano, Roland Siegwart, and Cesar Cadena.
\newblock \href{https://doi.org/10.1109/LRA.2022.3203812}{Continual Adaptation
  of Semantic Segmentation using Complementary 2D-3D Data Representations}.
\newblock {\em IEEE Robot. Autom. Lett.}, 2022.

\bibitem{Fu2022PanopticNeRF}
Xiao Fu, Shangzhan Zhang, Tianrun Chen, Yichong Lu, Lanyun Zhu, Xiaowei Zhou,
  Andreas Geiger, and Yiyi Liao.
\newblock \href{https://arxiv.org/abs/2203.15224}{Panoptic NeRF: 3D-to-2D Label
  Transfer for Panoptic Urban Scene Segmentation}.
\newblock In {\em 3DV}, 2022.

\bibitem{He2016ResNet101}
Kaiming He, Xiangyu Zhang, Shaoqing Ren, and Jian Sun.
\newblock
  \href{https://openaccess.thecvf.com/content_cvpr_2016/html/He_Deep_Residual_Learning_CVPR_2016_paper.html}{Deep
  Residual Learning for Image Recognition}.
\newblock In {\em CVPR}, 2016.

\bibitem{Hoffman2018CyCADA}
Judy Hoffman, Eric Tzeng, Taesung Park, Jun-Yan Zhu, Phillip Isola, Kate
  Saenko, Alexei Efros, and Trevor Darrell.
\newblock \href{https://proceedings.mlr.press/v80/hoffman18a}{CyCADA:
  Cycle-Consistent Adversarial Domain Adaptation}.
\newblock In {\em ICML}, 2018.

\bibitem{Jain2021PuttingNeRFOnADiet}
Ajay Jain, Matthew Tancik, and Pieter Abbeel.
\newblock
  \href{https://openaccess.thecvf.com/content/ICCV2021/html/Jain_Putting_NeRF_on_a_Diet_Semantically_Consistent_Few-Shot_View_Synthesis_ICCV_2021_paper.html}{Putting
  NeRF on a Diet: Semantically Consistent Few-Shot View Synthesis}.
\newblock In {\em ICCV}, 2021.

\bibitem{Kingma2015Adam}
Diederik~P Kingma and Jimmy Ba.
\newblock \href{https://arxiv.org/abs/1412.6980}{Adam: A Method for Stochastic
  Optimization}.
\newblock In {\em ICLR}, 2015.

\bibitem{Kobayashi2022FeatureFieldDistillation}
Sosuke Kobayashi, Eiichi Matsumoto, and Vincent Sitzmann.
\newblock \href{https://arxiv.org/abs/2205.15585}{Decomposing NeRF for Editing
  via Feature Field Distillation}.
\newblock In {\em NeurIPS}, 2022.

\bibitem{Kundu2022PanopticNeuralFields}
Abhijit Kundu, Kyle Genova, Xiaoqi Yin, Alireza Fathi, Caroline Pantofaru,
  Leonidas Guibas, Andrea Tagliasacchi, Frank Dellaert, and Thomas Funkhouser.
\newblock
  \href{https://openaccess.thecvf.com/content/CVPR2022/html/Kundu_Panoptic_Neural_Fields_A_Semantic_Object-Aware_Neural_Scene_Representation_CVPR_2022_paper.html}{Panoptic
  Neural Fields: A Semantic Object-Aware Neural Scene Representation}.
\newblock In {\em CVPR}, 2022.

\bibitem{Lee2019DTA}
Seungmin Lee, Dongwan Kim, Namil Kim, and Seong-Gyun Jeong.
\newblock
  \href{https://openaccess.thecvf.com/content_ICCV_2019/html/Lee_Drop_to_Adapt_Learning_Discriminative_Features_for_Unsupervised_Domain_Adaptation_ICCV_2019_paper.html}{Drop
  to Adapt: Learning Discriminative Features for Unsupervised Domain
  Adaptation}.
\newblock In {\em ICCV}, 2019.

\bibitem{Lesort2020CLForRobotics}
Timoth{\'e}e Lesort, Vincenzo Lomonaco, Andrei Stoian, Davide Maltoni, David
  Filliat, and Natalia Díaz-Rodríguez.
\newblock
  \href{https://www.sciencedirect.com/science/article/abs/pii/S1566253519307377}{Continual
  Learning for Robotics: Definition, Framework, Learning Strategies,
  Opportunities and Challenges}.
\newblock {\em Information Fusion}, 58, 2020.

\bibitem{Li2016AdaBN}
Yanghao Li, Naiyan Wang, Jianping Shi, Jiaying Liu, and Xiaodi Hou.
\newblock \href{https://arxiv.org/abs/1603.04779}{Revisiting Batch
  Normalization For Practical Domain Adaptation}.
\newblock {\em CoRR:1603.04779}, 2016.

\bibitem{Li2019BDL}
Yunsheng Li, Lu Yuan, and Nuno Vasconcelos.
\newblock
  \href{https://openaccess.thecvf.com/content_CVPR_2019/html/Li_Bidirectional_Learning_for_Domain_Adaptation_of_Semantic_Segmentation_CVPR_2019_paper.html}{Bidirectional
  Learning for Domain Adaptation of Semantic Segmentation}.
\newblock In {\em CVPR}, 2019.

\bibitem{Lin2014COCO}
Tsung-Yi Lin, Michael Maire, Serge Belongie, James Hays, Pietro Perona, Deva
  Ramanan, Piotr Doll{\'a}r, and C.~Lawrence Zitnick.
\newblock \href{https://doi.org/10.1007/978-3-319-10602-1_48}{Microsoft COCO:
  Common Objects in Context}.
\newblock In {\em ECCV}, 2014.

\bibitem{Lopez2017GEM}
David Lopez-Paz and Marc'Aurelio Ranzato.
\newblock
  \href{https://proceedings.neurips.cc/paper/2017/hash/f87522788a2be2d171666752f97ddebb-Abstract.html}{Gradient
  Episodic Memory for Continual Learning}.
\newblock In {\em NeurIPS}, 2017.

\bibitem{Maracani2021RECALL}
Andrea Maracani, Umberto Michieli, Marco Toldo, and Pietro Zanuttigh.
\newblock
  \href{https://openaccess.thecvf.com/content/ICCV2021/html/Maracani_RECALL_Replay-Based_Continual_Learning_in_Semantic_Segmentation_ICCV_2021_paper.html}{RECALL:
  Replay-based Continual Learning in Semantic Segmentation}.
\newblock In {\em ICCV}, 2021.

\bibitem{Michieli2020semanticDA}
Umberto Michieli, Matteo Biasetton, Gianluca Agresti, and Pietro Zanuttigh.
\newblock \href{https://doi.org/10.1109/TIV.2020.2980671}{Adversarial Learning
  and Self-Teaching Techniques for Domain Adaptation in Semantic Segmentation}.
\newblock {\em IEEE TIV}, 5, 2020.

\bibitem{Michieli2022DomainAdaptationCLChapter}
Umberto Michieli, Marco Toldo, and Pietro Zanuttigh.
\newblock
  \href{https://www.sciencedirect.com/science/article/pii/B9780128221099000175}{Domain
  adaptation and continual learning in semantic segmentation}.
\newblock In {\em Advanced Methods and Deep Learning in Computer Vision},
  chapter~8, pages 275--303. Elsevier, 2022.

\bibitem{Michieli2019ILT}
Umberto Michieli and Pietro Zanuttigh.
\newblock
  \href{https://openaccess.thecvf.com/content_ICCVW_2019/html/TASK-CV/Michieli_Incremental_Learning_Techniques_for_Semantic_Segmentation_ICCVW_2019_paper.html}{Incremental
  Learning Techniques for Semantic Segmentation}.
\newblock In {\em ICCVW}, 2019.

\bibitem{Michieli2021SDR}
Umberto Michieli and Pietro Zanuttigh.
\newblock
  \href{https://openaccess.thecvf.com/content/CVPR2021/html/Michieli_Continual_Semantic_Segmentation_via_Repulsion-Attraction_of_Sparse_and_Disentangled_Latent_CVPR_2021_paper.html}{Continual
  Semantic Segmentation via Repulsion-Attraction of Sparse and Disentangled
  Latent Representations}.
\newblock In {\em CVPR}, 2021.

\bibitem{Michieli2021KDIL}
Umberto Michieli and Pietro Zanuttigh.
\newblock
  \href{https://www.sciencedirect.com/science/article/pii/S1077314221000114}{Knowledge
  Distillation for Incremental Learning in Semantic Segmentation}.
\newblock {\em J. Comput. Vis. Image Understanding}, 205, 2021.

\bibitem{Mildenhall2020NeRF}
Ben Mildenhall, Pratul~P. Srinivasan, Matthew Tancik, Jonathan~T. Barron, Ravi
  Ramamoorthi, and Ren Ng.
\newblock \href{https://doi.org/10.1007/978-3-030-58452-8_24}{NeRF:
  Representing Scenes as Neural Radiance Fields for View Synthesis}.
\newblock In {\em ECCV}, 2020.

\bibitem{Mueller2022InstantNGP}
Thomas M\"uller, Alex Evans, Christoph Schied, and Alexander Keller.
\newblock \href{https://doi.org/10.1145/3528223.3530127}{Instant Neural
  Graphics Primitives with a Multiresolution Hash Encoding}.
\newblock {\em ACM TOG}, 41(4):102:1--102:15, 2022.

\bibitem{Murez2018ImageToImageTranslationDA}
Zak Murez, Soheil Kolouri, David Kriegman, Ravi Ramamoorthi, and Kyungnam Kim.
\newblock
  \href{https://openaccess.thecvf.com/content_cvpr_2018/html/Murez_Image_to_Image_CVPR_2018_paper.html}{Image
  to Image Translation for Domain Adaptation}.
\newblock In {\em CVPR}, 2018.

\bibitem{PageFortin2022CSSLeveragingLabelsAndRehearsal}
Mathieu Pag{\'e}~Fortin and Brahim Chaib-draa.
\newblock \href{https://www.ijcai.org/proceedings/2022/177}{Continual Semantic
  Segmentation Leveraging Image-level Labels and Rehearsal}.
\newblock In {\em IJCAI}, 2022.

\bibitem{Park2021Nerfies}
Keunhong Park, Utkarsh Sinha, Jonathan~T. Barron, Sofien Bouaziz, Dan~B
  Goldman, Steven~M. Seitz, and Ricardo Martin-Brualla.
\newblock
  \href{https://openaccess.thecvf.com/content/ICCV2021/html/Park_Nerfies_Deformable_Neural_Radiance_Fields_ICCV_2021_paper.html}{Nerfies:
  Deformable Neural Radiance Fields }.
\newblock In {\em ICCV}, 2021.

\bibitem{Pumarola2021D-NeRF}
Albert Pumarola, Enric Corona, Gerard Pons-Moll, and Francesc Moreno-Noguer.
\newblock
  \href{https://openaccess.thecvf.com/content/CVPR2021/html/Pumarola_D-NeRF_Neural_Radiance_Fields_for_Dynamic_Scenes_CVPR_2021_paper.html}{D-NeRF:
  Neural Radiance Fields for Dynamic Scenes}.
\newblock In {\em CVPR}, 2021.

\bibitem{Richter2016GTA}
Stephan~R. Richter, Vibhav Vineet, Stefan Roth, and Vladlen Koltun.
\newblock
  \href{https://link.springer.com/chapter/10.1007/978-3-319-46475-6_7}{Playing
  for Data: Ground Truth from Computer Games}.
\newblock In {\em ECCV}, 2016.

\bibitem{Ros2016SYNTHIA}
German Ros, Laura Sellart, Joanna Materzynska, David Vazquez, and Antonio~M.
  Lopez.
\newblock
  \href{https://openaccess.thecvf.com/content_cvpr_2016/html/Ros_The_SYNTHIA_Dataset_CVPR_2016_paper.html}{The
  SYNTHIA Dataset: A Large Collection of Synthetic Images for Semantic
  Segmentation of Urban Scenes}.
\newblock In {\em CVPR}, 2016.

\bibitem{Saito2018AdversarialDropoutRegularization}
Kuniaki Saito, Yoshitaka Ushiku, Tatsuya Harada, and Kate Saenko.
\newblock \href{https://openreview.net/forum?id=HJIoJWZCZ}{Adversarial Dropout
  Regularization}.
\newblock In {\em ICLR}, 2018.

\bibitem{Sankaranarayanan2018LSD-seg}
Swami Sankaranarayanan, Yogesh Balaji, Arpit Jain, Ser~Nam Lim, and Rama
  Chellappa.
\newblock
  \href{https://openaccess.thecvf.com/content_cvpr_2018/html/Sankaranarayanan_Learning_From_Synthetic_CVPR_2018_paper.html}{Learning
  From Synthetic Data: Addressing Domain Shift for Semantic Segmentation}.
\newblock In {\em CVPR}, 2018.

\bibitem{Schmid2022PanopticMultiTSDFs}
Lukas Schmid, Jeffrey Delmerico, Johannes Sch{\"o}nberger, Juan Nieto, Marc
  Pollefeys, Roland Siegwart, and Cesar Cadena.
\newblock \href{https://doi.org/10.1109/ICRA46639.2022.9811877}{Panoptic
  Multi-TSDFs: a Flexible Representation for Online Multi-resolution Volumetric
  Mapping and Long-term Dynamic Scene Consistency}.
\newblock In {\em ICRA}, 2022.

\bibitem{Shoemake1985Slerp}
Ken Shoemake.
\newblock \href{https://doi.org/10.1145/325334.325242}{Animating Rotation with
  Quaternion Curves}.
\newblock In {\em SIGGRAPH}, 1985.

\bibitem{Spadotto2021semanticDA}
Teo Spadotto, Marco Toldo, Umberto Michieli, and Pietro Zanuttigh.
\newblock
  \href{https://doi.ieeecomputersociety.org/10.1109/ICPR48806.2021.9412894}{Unsupervised
  Domain Adaptation with Multiple Domain Discriminators and Adaptive
  Self-Training}.
\newblock In {\em ICPR}, 2021.

\bibitem{torch-ngp}
Jiaxiang Tang.
\newblock {Torch-ngp: a PyTorch implementation of instant-ngp}.
\newblock \url{https://github.com/ashawkey/torch-ngp}, 2022.

\bibitem{Toldo2020UDAReview}
Marco Toldo, Andrea Maracani, Umberto Michieli, and Pietro Zanuttigh.
\newblock \href{https://doi.org/10.3390/technologies8020035}{Unsupervised
  Domain Adaptation in Semantic Segmentation: a Review}.
\newblock {\em Technologies}, 8, 2020.

\bibitem{Tschernezki2022NeuralFeatureFusionFields}
Vadim Tschernezki, Iro Laina, Diane Larlus, and Andrea Vedaldi.
\newblock \href{https://arxiv.org/abs/2209.03494}{Neural Feature Fusion Fields:
  3D Distillation of Self-Supervised 2D Image Representations}.
\newblock In {\em 3DV}, 2022.

\bibitem{Vora2022NeSF}
Suhani Vora, Noha Radwan, Klaus Greff, Henning Meyer, Kyle Genova, Mehdi S.~M.
  Sajjadi, Etienne Pot, Andrea Tagliasacchi, and Daniel Duckworth.
\newblock \href{https://openreview.net/forum?id=ggPhsYCsm9}{NeSF: Neural
  Semantic Fields for Generalizable Semantic Segmentation of 3D Scenes}.
\newblock {\em TMLR}, 2022.

\bibitem{Vu2019ADVENT}
Tuan-Hung Vu, Himalaya Jain, Maxime Bucher, Matthieu Cord, and Patrick Perez.
\newblock
  \href{https://openaccess.thecvf.com/content_CVPR_2019/html/Vu_ADVENT_Adversarial_Entropy_Minimization_for_Domain_Adaptation_in_Semantic_Segmentation_CVPR_2019_paper.html}{ADVENT:
  Adversarial Entropy Minimization for Domain Adaptation in Semantic
  Segmentation}.
\newblock In {\em CVPR}, 2019.

\bibitem{Wang2022CLIP-NeRF}
Can Wang, Menglei Chai, Mingming He, Dongdong Chen, and Jing Liao.
\newblock
  \href{https://openaccess.thecvf.com/content/CVPR2022/html/Wang_CLIP-NeRF_Text-and-Image_Driven_Manipulation_of_Neural_Radiance_Fields_CVPR_2022_paper.html}{CLIP-NeRF:
  Text-and-Image Driven Manipulation of Neural Radiance Fields}.
\newblock In {\em CVPR}, 2022.

\bibitem{Wang2022CoTTA}
Qin Wang, Olga Fink, Luc Van~Gool, and Dengxin Dai.
\newblock
  \href{https://openaccess.thecvf.com/content/CVPR2022/html/Wang_Continual_Test-Time_Domain_Adaptation_CVPR_2022_paper.html}{Continual
  Test-Time Domain Adaptation}.
\newblock In {\em CVPR}, 2022.

\bibitem{Wu2019ACE}
Zuxuan Wu, Xin Wang, Joseph~E. Gonzalez, Tom Goldstein, and Larry~S. Davis.
\newblock
  \href{https://openaccess.thecvf.com/content_ICCV_2019/html/Wu_ACE_Adapting_to_Changing_Environments_for_Semantic_Segmentation_ICCV_2019_paper.html}{ACE:
  Adapting to Changing Environments for Semantic Segmentation}.
\newblock In {\em ICCV}, 2019.

\bibitem{Yang2020FDA}
Yanchao Yang and Stefano Soatto.
\newblock
  \href{https://openaccess.thecvf.com/content_CVPR_2020/html/Yang_FDA_Fourier_Domain_Adaptation_for_Semantic_Segmentation_CVPR_2020_paper.html}{FDA:
  Fourier Domain Adaptation for Semantic Segmentation}.
\newblock In {\em CVPR}, 2020.

\bibitem{Zhang2018FCAN}
Yiheng Zhang, Zhaofan Qiu, Ting Yao, Dong Liu, and Tao Mei.
\newblock
  \href{https://openaccess.thecvf.com/content_cvpr_2018/html/Zhang_Fully_Convolutional_Adaptation_CVPR_2018_paper.html}{Fully
  Convolutional Adaptation Networks for Semantic Segmentation}.
\newblock In {\em CVPR}, 2018.

\bibitem{Zheng2021MRNetRectifying}
Zhedong Zheng and Yi Yang.
\newblock
  \href{https://link.springer.com/article/10.1007/s11263-020-01395-y}{Rectifying
  Pseudo Label Learning via Uncertainty Estimation for Domain Adaptive Semantic
  Segmentation}.
\newblock {\em IJCV}, 2021.

\bibitem{Zhi2021SemanticNeRF}
Shuaifeng Zhi, Tristan Laidlow, Stefan Leutenegger, and Andrew~J. Davison.
\newblock
  \href{https://openaccess.thecvf.com/content/ICCV2021/html/Zhi_In-Place_Scene_Labelling_and_Understanding_With_Implicit_Scene_Representation_ICCV_2021_paper.html}{In-Place
  Scene Labelling and Understanding with Implicit Scene Representation}.
\newblock In {\em ICCV}, 2021.

\bibitem{Zhi2021iLabel}
Shuaifeng Zhi, Edgar Sucar, Andre Mouton, Iain Haughton, Tristan Laidlow, and
  Andrew~J. Davison.
\newblock \href{https://arxiv.org/abs/2111.14637}{iLabel: Interactive Neural
  Scene Labelling}.
\newblock {\em CoRR:2111.14637}, 2021.

\bibitem{Zou2018CBST}
Yang Zou, Zhiding Yu, B.V.K.~Vijaya Kumar, and Jinsong Wang.
\newblock
  \href{https://openaccess.thecvf.com/content_ECCV_2018/html/Yang_Zou_Unsupervised_Domain_Adaptation_ECCV_2018_paper.html}{Unsupervised
  Domain Adaptation for Semantic Segmentation via Class-Balanced
  Self-Training}.
\newblock In {\em ECCV}, 2018.

\bibitem{Zou2019CRST}
Yang Zou, Zhiding Yu, Xiaofeng Liu, B.V.K.~Vijaya Kumar, and Jinsong Wang.
\newblock
  \href{https://openaccess.thecvf.com/content_ICCV_2019/html/Zou_Confidence_Regularized_Self-Training_ICCV_2019_paper.html}{Confidence
  Regularized Self-Training}.
\newblock In {\em ICCV}, 2019.

\end{thebibliography}
}

\cleardoublepage
\appendix
\section*{Supplementary Material}
The Supplementary Material is organized as follows.
 In Sec.~\ref{sec:appendix_implementation_details}, we
provide
 additional implementation details.
 In Sec.~\ref{sec:appendix_nerf_pseudolabels}, we present ablations on the NeRF-based \pls, showing the effect on their quality of different parameters and components of our method.
In Sec.~\ref{sec:appendix_one_step_adaptation}, we report additional evaluations for the one-step adaptation experiments.
In Sec.~\ref{sec:appendix_multi_step_adaptation} we include in-detail results for the multi-step adaptation experiments and ablate on the replay-based strategy proposed by our method. 
In Sec.~\ref{appendix:memory_footprint} we analyze in detail the memory footprint required by our method and by the different baselines that we compare against in the main paper.
In Sec.~\ref{sec:appendix_further_visualizations}, we provide further visualizations, including examples of the \pls and network predictions produced by our method and the baselines.
In Sec.~\ref{sec:appendix_limitations}, we discuss limitations of our method and potential ways to address them.
We will further
release
the code to reproduce our results.

Similarly to the main paper, in all the experiments we report mean intersection over union ($\mathrm{mIoU}$, in percentage values) as a metric.
\section{Additional implementation details}
\label{sec:appendix_implementation_details}
\textbf{NeRF.}
Following Instant-NGP~\cite{Mueller2022InstantNGP, torch-ngp}, to facilitate training of the hash encoding, we re-scale and re-center the poses used to train NeRF so that they fit in a fixed-size cube.
For each ray that is cast from the training viewpoints,
to render the
aggregated
colors and semantics labels
we first sample $256$ points at a fixed interval and
then
randomly select
$256$ additional points
according to the density values of the initial points.

The base NeRF network uses a multi-resolution hash encoding with a $16$-level hash table of size $2^{19}$ and a feature dimension of $2$.
Similarly to \snerf~\cite{Zhi2021SemanticNeRF}, we implement the additional semantic head as a
$2$-layer MLP.
In all the experiments, we train all the components of the Semantic-NeRF network concurrently, setting the 
hyperparameters
in
Eq.~\hbox{(5)}
from the main paper
to $w_\mathrm{d}=0.1$ and $w_\mathrm{s}=0.04$ as suggested in~\cite{Zhi2021SemanticNeRF},
sampling $4096$ rays for each viewpoint,
and using the Adam~\cite{Kingma2015Adam} optimizer with a fixed learning rate of $1\mathrm{e}{-2}$.

In all the experiments in which the semantic segmentation model is trained using NeRF-rendered images, 
we use Adaptive Batch Normalization (AdaBN)~\cite{Li2016AdaBN}
when performing inference on the ground-truth images, to improve the generalization ability of the model between NeRF-rendered images and ground-truth images.
\\\textbf{Dataset.}
For convenience of notation, we re-map the scene indices in the dataset from $\mathrm{0000}-\mathrm{0706}$ to $1-707$ (so that we refer to scene $\mathrm{0000}$ as scene $1$, to scene $\mathrm{0001}$ as scene $2$, etc.).
For sample efficiency, we
downsample each sequence
from the original \SI{30}{fps}
to \SI{3}{fps}, resulting in a total of $100$ to $500$ frames for each video sequence.
\\\textbf{Pre-training.}
To pre-train DeepLab on scenes $11-707$ from ScanNet, we initialize the model parameters with the
weights pre-trained on the COCO semantic segmentation dataset~\cite{Lin2014COCO}.
We then run the pre-training on ScanNet using
the Adam~\cite{Kingma2015Adam} optimizer with
batch size of $4$,
and let
the learning rate
decay
linearly from $1\mathrm{e}{-4}$ to $1\mathrm{e}{-6}$ over $150$ epochs.
\\\textbf{One-step adaptation.}
In all the one-step experiments with our method and with the baseline of~\cite{Frey2022CLSemanticSegmentation}, the semantic segmentation model is trained for $50$ epochs with a fixed learning rate of $1\mathrm{e}{-5}$ and batch size of $4$. 
Since CoTTA is an \emph{online} adaptation method, in accordance with the settings introduced in the original paper, we adapt the segmentation network for a single epoch and with batch size $1$, setting the learning rate to $2.5\mathrm{e}{-6}$. 
To prevent overfitting the semantic segmentation model to the training views of the new scene,
we
apply the same data augmentation procedure as in pre-training
in each training step for our method and for~\cite{Frey2022CLSemanticSegmentation}. Since CoTTA already implements a label augmentation mechanism for ensembling, we
apply to the method only the augmentations used by its authors.
\\\textbf{Multi-step adaptation.}
In the multi-step adaptation experiments, we use a batch size of $4$ during training, where $2$ samples come from
the
subset
of the pre-training dataset used for replay (cf. main paper), and the other $2$
data points
are
uniformly sampled
from
the training frames of the new scene and the
replay
buffer of the previous scenes.
\\\textbf{Hardware.}
We train all our
models using an AMD Ryzen 9 5900X with $\SI{32}{GB}$ RAM, and an NVIDIA RTX3090 GPU with $\SI{24}{GB}$ VRAM.
\begin{table*}[!ht]
\centering
\resizebox{\linewidth}{!}{
\begin{tabular}{ccccccccccccccc}
\toprule
\multicolumn{2}{c}{Components} & \multicolumn{11}{c}{Scene}                                                                                                                                                                               \\ \midrule
$\mathcal{L}_{\mathrm{d}}$  & $\mathcal{L}_{\mathrm{s}}$   & Scene 1 & Scene 2 & Scene 3 & Scene 4 & Scene 5   & Scene 6 & Scene 7 & Scene 8 & Scene 9 & Scene 10 & Average \\ \midrule
\xmark       & \snerf~\cite{Zhi2021SemanticNeRF}   &    44.3{\scriptsize$\pm$1.5}         &  34.2{\scriptsize$\pm$0.1}       &   22.4{\scriptsize$\pm$0.9}          &  
\textbf{63.5}{\scriptsize$\pm$1.2}           &   52.3{\scriptsize$\pm$1.2}            & 47.3{\scriptsize$\pm$0.5}                                & 38.9{\scriptsize$\pm$0.6}                                & 33.8{\scriptsize$\pm$0.4}                                &  32.4{\scriptsize$\pm$0.5}                               & 53.3{\scriptsize$\pm$0.6}  &
 42.2{\scriptsize$\pm$0.7}   \\  
\xmark       & Ours          &   46.4{\scriptsize$\pm$1.1}        &   33.0{\scriptsize$\pm$0.2}      &    24.2{\scriptsize$\pm$0.3}       &    62.6{\scriptsize$\pm$0.7}      &    53.4{\scriptsize$\pm$0.7}        &    46.8{\scriptsize$\pm$1.1}                          &  39.3{\scriptsize$\pm$0.8}                            &     \textbf{34.5}{\scriptsize$\pm$0.6}                         &      33.8{\scriptsize$\pm$0.6}                        &    55.8{\scriptsize$\pm$0.2} &43.0{\scriptsize$\pm$0.6}                           \\ \ 
\cmark       & \snerf~\cite{Zhi2021SemanticNeRF}   &   44.0{\scriptsize$\pm$0.6}         &    34.8{\scriptsize$\pm$0.5}        &    22.8{\scriptsize$\pm$0.9}        &   63.1{\scriptsize$\pm$0.7}         &  55.8{\scriptsize$\pm$2.0}  &  49.1{\scriptsize$\pm$1.2}                              &  39.0{\scriptsize$\pm$0.8}                              &  33.9{\scriptsize$\pm$0.5}                              & 33.0{\scriptsize$\pm$1.5}                               & 55.1{\scriptsize$\pm$0.6} &
 43.1{\scriptsize$\pm$0.9}  \\  
\cmark       & Ours            &   \textbf{48.4}{\scriptsize$\pm$0.9}       &   \textbf{36.0}{\scriptsize$\pm$0.3}         &  \textbf{26.1}{\scriptsize$\pm$0.4}          &   61.6{\scriptsize$\pm$0.5}         &   \textbf{57.0}{\scriptsize$\pm$1.8}          &     \textbf{50.3}{\scriptsize$\pm$0.6}                          &     \textbf{39.8}{\scriptsize$\pm$0.2}                         &  33.5{\scriptsize$\pm$0.6}                              &  \textbf{35.4}{\scriptsize$\pm$0.7}                              &       \textbf{57.4}{\scriptsize$\pm$0.1}         &
 \textbf{44.6}{\scriptsize$\pm$0.7}  \\ \bottomrule
\end{tabular}}
\caption{Effect of
the $\ell_1$ depth loss $\mathcal{L}_\mathrm{d}$ and of different types of semantic losses (either the original one proposed in~\cite{Zhi2021SemanticNeRF} or ours) on the pseudo-label quality. The performance is evaluated on the training views of each scene and averaged over 3 runs.
}
\label{tab:ab_nerf_loss}
\vspace{-3ex}
\end{table*}

\section{NeRF-based \pls}
\label{sec:appendix_nerf_pseudolabels}
In the following Section, we present 
ablations on the NeRF-based \pls, showing how the chosen NeRF implementation and the losses used in our method influence their segmentation accuracy.

\subsection{Comparison of NeRF frameworks}
\label{appendix:comparison_semanticnerf}
We compare the segmentation quality of the \pls obtained with
our
Instant-NGP~\cite{Mueller2022InstantNGP, torch-ngp}-based implementation to that achieved with the original \snerf~\cite{Zhi2021SemanticNeRF} implementation, which we adapt to include the newly-introduced semantic loss (cf. Sec.~\hbox{3.2} in the main paper and Sec.~\ref{appendix:comparison_semantic_losses}).
To this purpose, we train a semantics-aware NeRF model for scene $1$ with both the methods, running the experiments 
$3$ times for each method.
In each run, we train the original implementation of \snerf~\cite{Zhi2021SemanticNeRF} for $\num{200}\mathrm{k}$ steps and
the one based on Instant-NGP~\cite{torch-ngp} for $10$ epochs (for a total of $10 \times 447 = \num{4470}$ steps), which allows
achieving a similar
color
reconstruction quality (measured as PSNR) for the two methods.
\begin{table}[!ht]

    \centering
    \resizebox{\linewidth}{!}{
    \begin{tabular}{@{}lcc@{}}
\toprule
                                  & \snerf~\cite{Zhi2021SemanticNeRF} & Instant-NGP~\cite{Mueller2022InstantNGP} (impl. by~\cite{torch-ngp}) \\ \midrule
                                  $\mathrm{PSNR}$                   & 19.9 {\scriptsize$\pm$0.1}        & 19.3 {\scriptsize$\pm$0.1}                                           \\
$\mathrm{mIoU}$                   & 50.0 {\scriptsize$\pm$0.5}        & 48.4 {\scriptsize$\pm$0.9}                                           \\
Model size ($\mathrm{MB}$)        & 4.9 & 50.0 \\
Training time / Step  ($\mathrm{s}$) &     0.19                              &  0.06                  
                                         \\
Total training time  ($\mathrm{min}$) &  633                                 &      5           \\
Inference time / Image  ($\mathrm{s}$)  &  2.8                                  &            0.3                                                        \\ \bottomrule
\end{tabular}}
    \caption{Pseudo-label performance on the training views of scene $1$, size of the associated model checkpoint, and the training and inference time using different NeRF frameworks. The implementation of~\cite{Zhi2021SemanticNeRF}
    has been adapted to include the newly-introduced semantic loss (cf. Sec.~\hbox{3.2} in the main paper).
    The results are averaged over $3$ runs.
    }
    \label{tab:comparison_nerf_frameworks}
    \vspace{-10pt}
\end{table}

As shown in Tab.~\ref{tab:comparison_nerf_frameworks}, the \pls produced by both implementations achieve a similar $\mathrm{mIoU}$, with 
\snerf slightly
outperforming
Instant-NGP.
Furthermore, the size of the models produced by \snerf is approximately $10$ times smaller than
the one
required by Instant-NGP, at the cost however of longer training ($\sim127\times$) and rendering ($\sim9\times$) time.

Since in a real-world deployment scenario achieving fast adaptation might be of high priority,
in the main paper we adopted the faster framework of Instant-NGP. However, the results above
indicate
that our method is agnostic to the specific NeRF framework chosen, and similar segmentation performance can be achieved by trading off
between
speed and model size
depending on the main requirements.
Further evaluations on the memory footprint in comparison also with the baselines of~\cite{Frey2022CLSemanticSegmentation} and~\cite{Wang2022CoTTA} are presented in Sec.~\ref{appendix:memory_footprint}.

\subsection{Ablation on the NeRF losses}
\label{appendix:comparison_semantic_losses}
To
investigate
the effect of
depth supervision~\cite{Deng2022DepthSupervisedNeRF} (through the
$\ell_1$ depth loss $\mathcal{L}_\mathrm{d}$) and of the
proposed modifications to the
semantic loss $\mathcal{L}_\mathrm{s}$
(cf. Sec.~\hbox{3.2} in the main paper),
we evaluate on each scene the \pls produced by our method when ablating on these factors.
For each scene, we train the NeRF model for $10$ epochs without joint training, as we find training without semantic loss modifications is unstable for longer epochs. We run each experiment 3 times and report average and standard deviation across the runs.
As shown in
Tab.~\ref{tab:ab_nerf_loss},
both components
induce
a significant improvement of the \pl quality. In particular, depth supervision and the use of our modified semantic loss instead of the one proposed in~\cite{Zhi2021SemanticNeRF} produce an increase respectively of $0.9\%\ \mathrm{mIoU}$ and $0.8\%\ \mathrm{mIoU}$ over the baseline with no modifications.
The combined use of both ablated factors further increases the \pl performance, resulting in a total improvement by $2.4\%\ \mathrm{mIoU}$.

\begin{figure*}[!ht]
\centering
\def\colwidth{0.14\textwidth}
\def\colwidthsem{0.12\textwidth}
\def\extraspaceaddedwidth{0.001\textwidth}
\newcolumntype{M}[1]{>{\centering\arraybackslash}m{#1}}
\addtolength{\tabcolsep}{-4pt}
\begin{subtable}{\linewidth}\centering
    {
\begin{tabular}{M{\colwidth} M{\colwidth} M{\colwidth}  M{\colwidth} M{\extraspaceaddedwidth}
M{\colwidth} M{\colwidth} }
 \multicolumn{4}{c}{Depth} &
 &
 \multicolumn{2}{c}{Color} \tabularnewline
 \cmidrule(r){1-4} \cmidrule(r){6-7}
 Ours w/o $\mathcal{L}_\mathrm{d}$ &  Ours with $\mathcal{L}_\mathrm{s}$ from~\cite{Zhi2021SemanticNeRF} & Ours  & Ground truth & 
 & Ours & Ground truth \tabularnewline
\includegraphics[width=\linewidth]{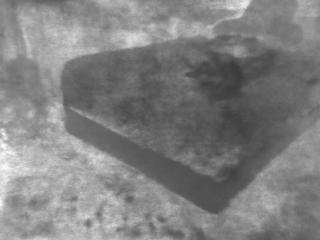} & 
\includegraphics[width=\linewidth]{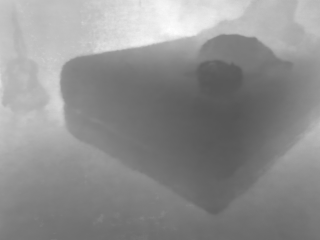} &
\includegraphics[width=\linewidth]{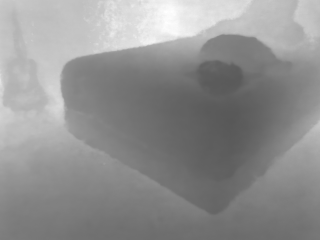} &
\includegraphics[width=\linewidth]{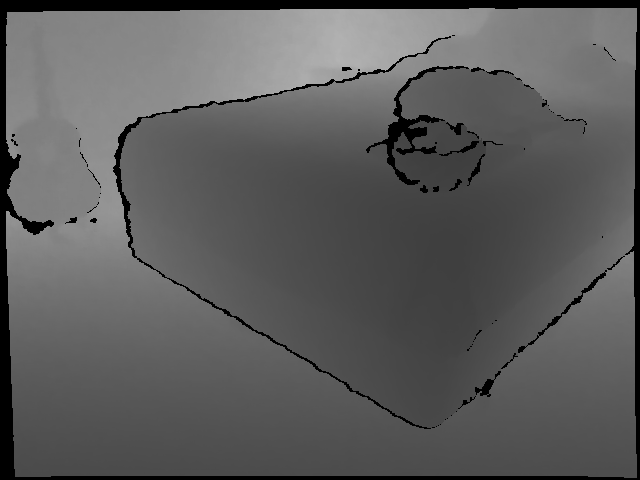} &
&
\includegraphics[width=\linewidth]{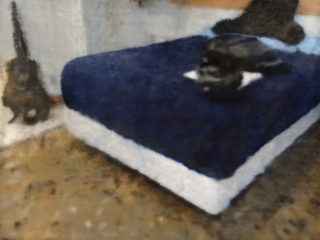} &
\includegraphics[width=\linewidth]{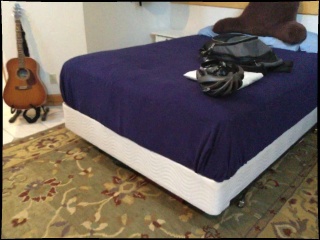}
  \tabularnewline
\includegraphics[width=\linewidth]{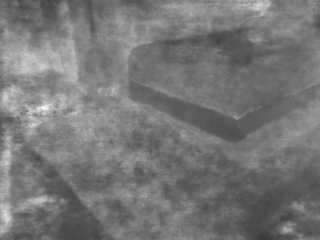} & 
\includegraphics[width=\linewidth]{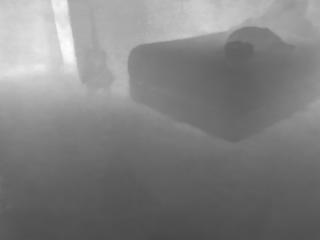} &
\includegraphics[width=\linewidth]{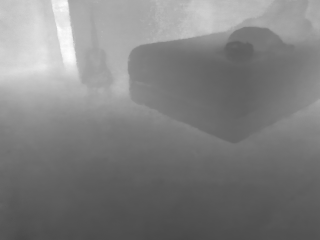} &
\includegraphics[width=\linewidth]{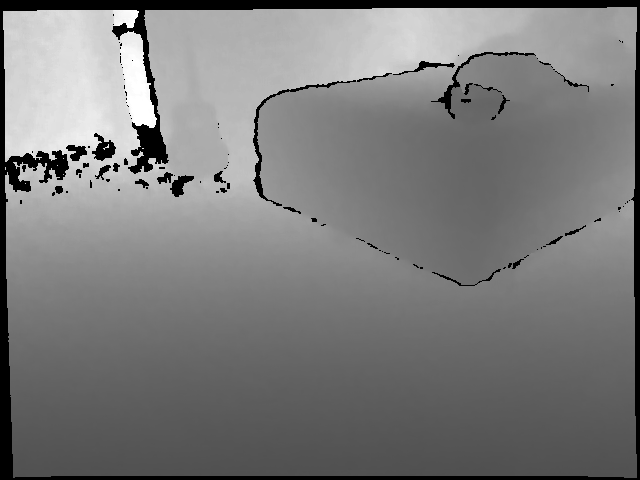} &
&
\includegraphics[width=\linewidth]{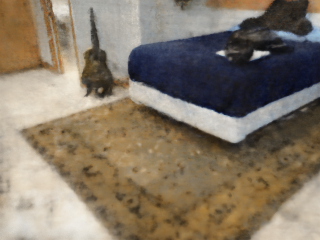} &
\includegraphics[width=\linewidth]{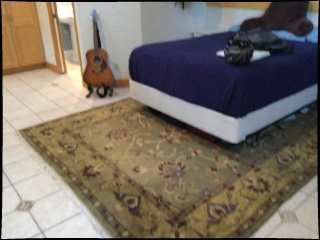}
\tabularnewline
\includegraphics[width=\linewidth]{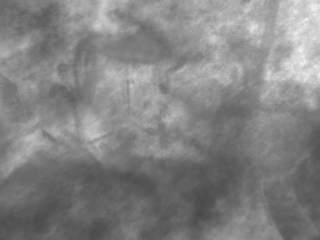} & 
\includegraphics[width=\linewidth]{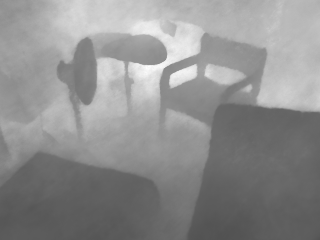} &
\includegraphics[width=\linewidth]{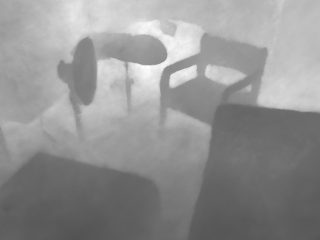} &
\includegraphics[width=\linewidth]{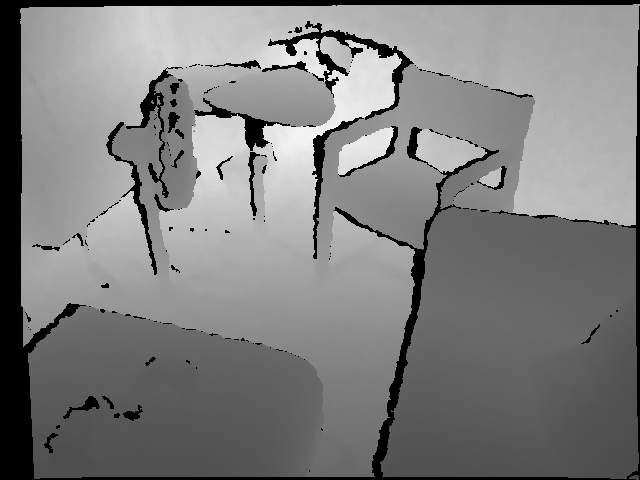} &
&
\includegraphics[width=\linewidth]{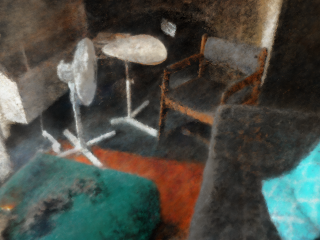} &
\includegraphics[width=\linewidth]{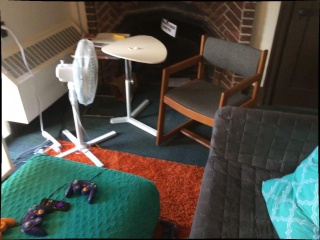}
\tabularnewline
\includegraphics[width=\linewidth]{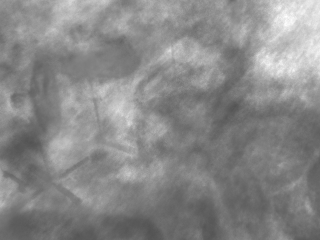} & 
\includegraphics[width=\linewidth]{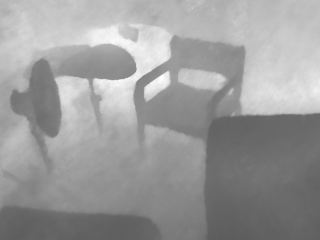} &
\includegraphics[width=\linewidth]{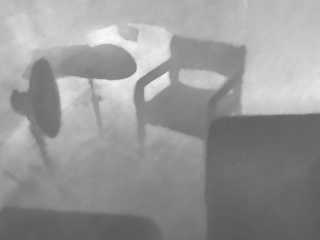} &
\includegraphics[width=\linewidth]{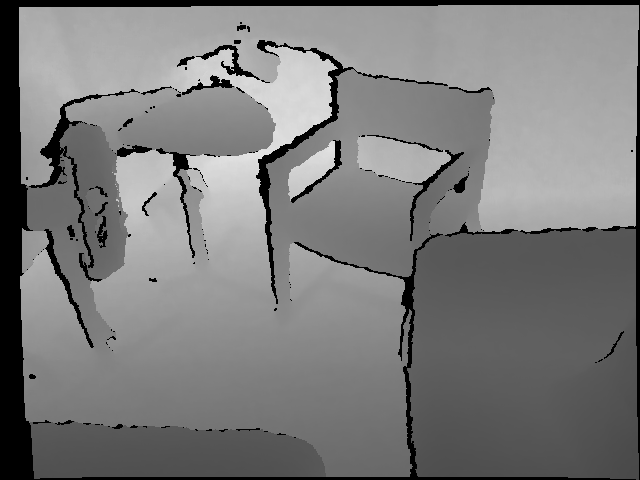} &
&
\includegraphics[width=\linewidth]{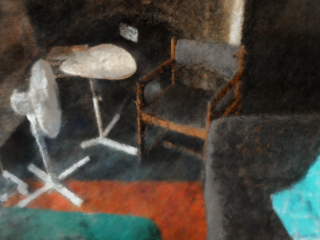} &
\includegraphics[width=\linewidth]{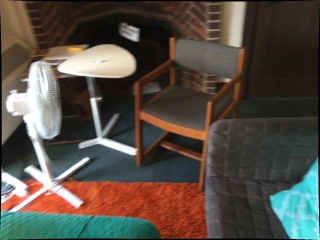}
\tabularnewline
\end{tabular}
}
\caption{Rendered depth}
\label{fig:depth_loss_ablation}
\vspace{15pt}
\end{subtable}
\begin{subtable}{\linewidth}\centering
    {\begin{tabular}{M{\colwidthsem} M{\colwidthsem} M{\colwidthsem} M{\colwidthsem}  M{\colwidthsem} M{\extraspaceaddedwidth}
M{\colwidthsem} M{\colwidthsem} }
 \multicolumn{5}{c}{Semantics} &
 &
 \multicolumn{2}{c}{Color} \tabularnewline
 \cmidrule(r){1-5} \cmidrule(r){7-8}
 Ours w/o $\mathcal{L}_\mathrm{d}$ & Ours with $\mathcal{L}_\mathrm{s}$ from~\cite{Zhi2021SemanticNeRF} & Ours  & Pre-train & Ground truth & 
 & Ours & Ground truth \tabularnewline
\includegraphics[width=\linewidth]{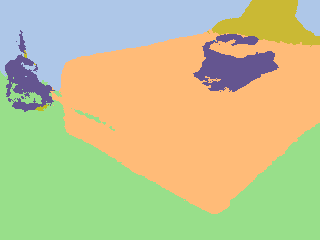} & 
\includegraphics[width=\linewidth]{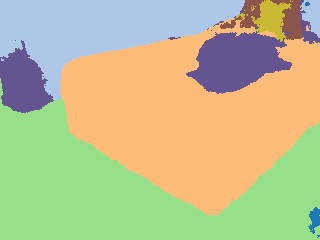} &
\includegraphics[width=\linewidth]{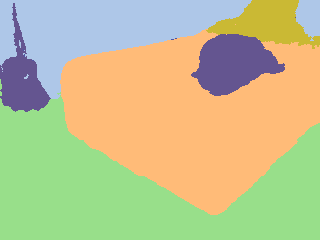} &
\includegraphics[width=\linewidth]{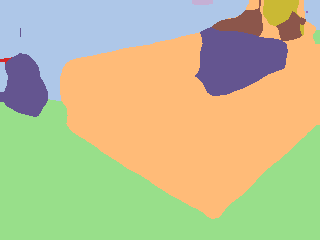} &
\includegraphics[width=\linewidth]{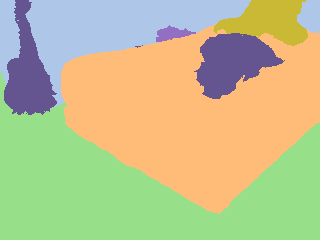} &
&
\includegraphics[width=\linewidth]{figures/comparison_nerf/0_ours_all_rgb.png} &
\includegraphics[width=\linewidth]{figures/comparison_nerf/0_gt_rgb.jpg}
  \tabularnewline
\includegraphics[width=\linewidth]{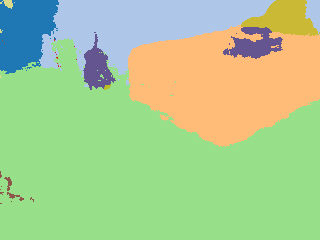} & 
\includegraphics[width=\linewidth]{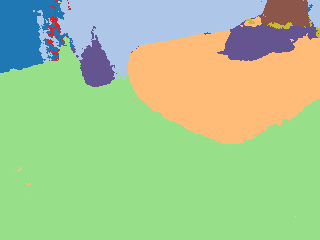} &
\includegraphics[width=\linewidth]{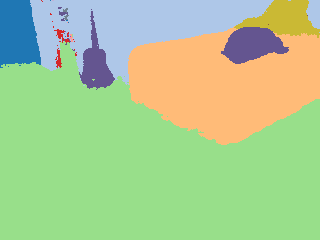} &
\includegraphics[width=\linewidth]{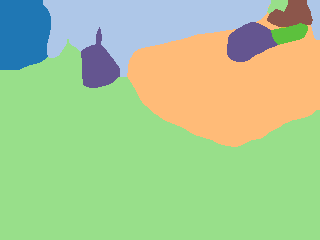} &
\includegraphics[width=\linewidth]{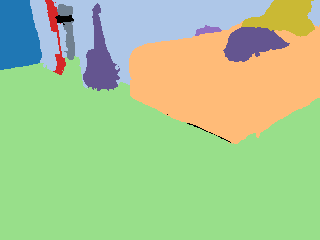} &
&
\includegraphics[width=\linewidth]{figures/comparison_nerf/1_ours_all_rgb.png} &
\includegraphics[width=\linewidth]{figures/comparison_nerf/1_gt_rgb.jpg}
\tabularnewline
\includegraphics[width=\linewidth]{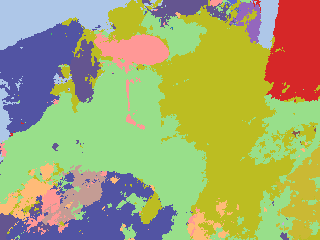} & 
\includegraphics[width=\linewidth]{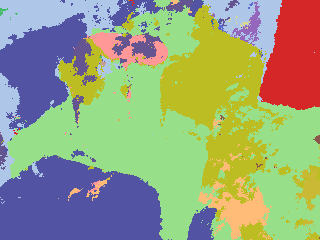} &
\includegraphics[width=\linewidth]{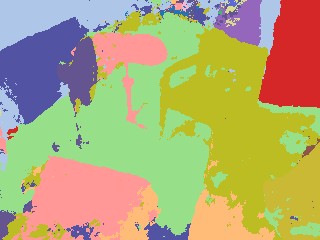} &
\includegraphics[width=\linewidth]{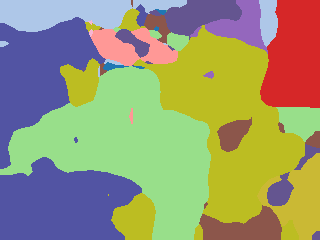} &
\includegraphics[width=\linewidth]{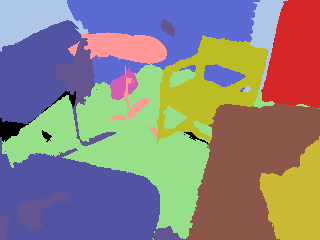} &
&
\includegraphics[width=\linewidth]{figures/comparison_nerf/2_ours_all_rgb.png} &
\includegraphics[width=\linewidth]{figures/comparison_nerf/2_gt_rgb.jpg}
\tabularnewline
\includegraphics[width=\linewidth]{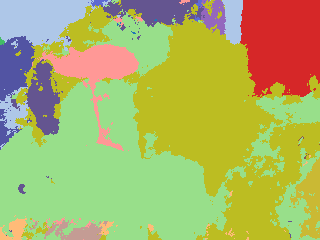} & 
\includegraphics[width=\linewidth]{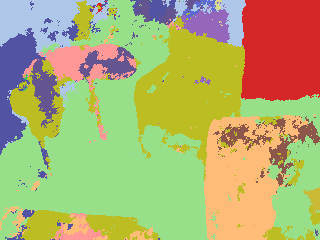} &
\includegraphics[width=\linewidth]{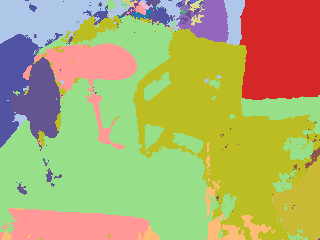} &
\includegraphics[width=\linewidth]{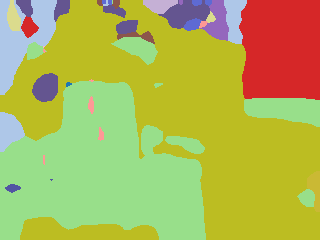} &
\includegraphics[width=\linewidth]{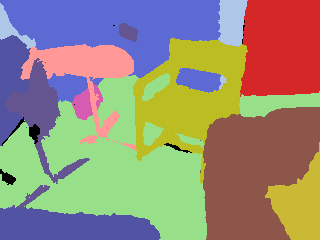} &
&
\includegraphics[width=\linewidth]{figures/comparison_nerf/3_ours_all_rgb.png} &
\includegraphics[width=\linewidth]{figures/comparison_nerf/3_gt_rgb.jpg}
\tabularnewline
\end{tabular}
}
\caption{Rendered semantics}
\label{fig:semantic_loss_ablation}
\end{subtable}
\caption{Effect on the rendered depth and semantics of
depth supervision and of the modification to the semantic loss. Black pixels in the ground-truth depth and ground-truth semantics denote respectively missing depth measurement and missing semantic annotation.
\label{fig:depth_and_semantic_losses_ablation}}
\end{figure*}

The effect of the proposed modifications
can also be observed in Fig.~\ref{fig:depth_and_semantic_losses_ablation}.
In particular, as shown in Fig.~\ref{fig:depth_loss_ablation}, the use of depth supervision is critical for properly reconstructing the scene geometry.
The large number of artifacts in the reconstruction when the depth loss is not used
are
also reflected in the semantic \pls, which
contain large levels of noise
and often fail to assign a uniform
class
to each entity in the scene (Fig.~\ref{fig:semantic_loss_ablation}).
Depth supervision
applied together
with the original semantic loss from~\cite{Zhi2021SemanticNeRF} resolves some of the artifacts in the \pls, but still results in suboptimal quality. The combined use of depth supervision and of our modified semantic loss produces cleaner and smoother \pls, which also attain higher segmentation accuracy, as shown in Tab.~\ref{tab:ab_nerf_loss}.

\section{One-step adaptation\label{sec:appendix_one_step_adaptation}}

In this Section, we report additional results on the one-step adaptation experiments.
\subsection{One-step adaptation performance on the training set of each scene}
Since in the scenario of a deployment of the semantic segmentation network on a real-world system a scene might be revisited from viewpoints similar to those used for training, in Tab.~\ref{tab:1_step_train} we report the one-step adaptation performance evaluated on the training views.
We compare our method to the baseline of CoTTA~\cite{Wang2022CoTTA} and to fine-tuning, both with the \pls of~\cite{Frey2022CLSemanticSegmentation} and with our NeRF-based \pls. For each method, we run the experiments 10 times and report average and standard deviation across the runs.
\begin{table*}
\centering
\resizebox{\linewidth}{!}{
\begin{tabular}{@{}ccccccc@{}}
\toprule
        & Pre-train & CoTTA~\cite{Wang2022CoTTA}                     & Fine-tuning ($\mathrm{GI}+\mathrm{ML}$) & Fine-tuning ($\mathrm{GI}+\mathrm{NL}$) & Fine-tuning ($\mathrm{NI}+\mathrm{NL}$) & Joint Training            \\ \midrule
Scene $1$ & 41.1      & 41.9{\scriptsize$\pm$0.0}   &   50.6{\scriptsize$\pm$0.1}             &   50.1{\scriptsize$\pm$0.6}              &   
50.7{\scriptsize$\pm$0.5}      & \textbf{55.5}{\scriptsize$\pm$1.3}  \\ 
Scene $2$ & 35.5     &35.6{\scriptsize$\pm$0.0}    &  33.5{\scriptsize$\pm$0.1}             &    35.7{\scriptsize$\pm$0.8}           &          
36.6{\scriptsize$\pm$0.3}      &   \textbf{39.5}{\scriptsize$\pm$0.8} \\ 
Scene $3$ & 23.5     & 23.7{\scriptsize$\pm$0.0}   &   24.4{\scriptsize$\pm$0.1}             &    26.9{\scriptsize$\pm$1.0}          &           27.1{\scriptsize$\pm$1.2}                                 &  \textbf{27.5}{\scriptsize$\pm$1.6}     \\ 
Scene $4$ &  62.8     &  63.0{\scriptsize$\pm$0.0}  &    66.1{\scriptsize$\pm$0.3}            &    63.2{\scriptsize$\pm$0.6}                                  &            66.1{\scriptsize$\pm$0.8}                                 &   \textbf{67.7}{\scriptsize$\pm$1.7}      \\ 
Scene $5$ & 49.8     & 49.8{\scriptsize$\pm$0.0}   &   51.2{\scriptsize$\pm$0.1}              &  57.1{\scriptsize$\pm$1.2}                                    &             \textbf{59.9}{\scriptsize$\pm$1.5}                                &    46.3{\scriptsize$\pm$0.3}      \\ 
Scene $6$ &  48.9    & 48.9{\scriptsize$\pm$0.0}   &  \textbf{53.1}{\scriptsize$\pm$0.1}              &   50.2{\scriptsize$\pm$0.4}                                    &                49.9{\scriptsize$\pm$0.4}                             &    50.7{\scriptsize$\pm$0.2}    \\ 
Scene $7$ &  39.7     &  39.8{\scriptsize$\pm$0.0}  &  41.4{\scriptsize$\pm$0.1}                 &   40.8{\scriptsize$\pm$0.6}                                   &                  42.1{\scriptsize$\pm$0.8}                           &    \textbf{43.8}{\scriptsize$\pm$1.6}   \\ 
Scene $8$ & 31.6      &31.7{\scriptsize$\pm$0.0}    &  36.2{\scriptsize$\pm$0.2}               &     34.4{\scriptsize$\pm$0.5}                                 &        33.9{\scriptsize$\pm$0.4}                                     &    \textbf{38.1}{\scriptsize$\pm$3.5}   \\ 
Scene $9$ & 31.7  & 31.7{\scriptsize$\pm$0.0}   &   32.7{\scriptsize$\pm$0.1}           &    \textbf{35.5}{\scriptsize$\pm$0.6}                                  &        34.9{\scriptsize$\pm$0.8}                                     &   32.5{\scriptsize$\pm$0.9}    \\ 
Scene $10$ & 52.5      & 52.7{\scriptsize$\pm$0.0}  &  57.8{\scriptsize$\pm$0.1}              &   57.1{\scriptsize$\pm$0.6}                                   &          \textbf{58.4}{\scriptsize$\pm$0.6}                                   &   57.4{\scriptsize$\pm$1.4}     \\ 
\midrule
Average &  41.7    & 41.9{\scriptsize$\pm$0.0}   &   44.7{\scriptsize$\pm$0.1}              &  45.1{\scriptsize$\pm$0.7}                                    &            \textbf{45.9}{\scriptsize$\pm$0.7}                                 &    \textbf{45.9}{\scriptsize$\pm$1.3}   \\ \bottomrule
\end{tabular}}
\caption{One-step adaptation performance on the training
views
of each scene. $\mathrm{GI}$ and $\mathrm{NI}$ denote respectively ground-truth color images and NeRF-rendered color images. $\mathrm{ML}$ and $\mathrm{NL}$ indicate adaptation using \pls formed respectively with the method of~\cite{Frey2022CLSemanticSegmentation} and with our
approach. In joint training, we use NeRF-based renderings and \pls. For each method, we run the experiments for 10 times and report average and standard deviation across the runs.}
\label{tab:1_step_train}
\end{table*}

Similarly to the results obtained on the validation views (cf. main paper), our method with joint training obtains the best average performance across all scenes. 
Unlike
what observed
on the validation views,
however,
on the training views
joint training does not result in an average performance improvement
over
fine-tuning with our NeRF-based \pls ($\mathrm{NI}+\mathrm{NL}$).
We note however that
these results are largely influenced by the outlier of Scene $5$, where joint training achieves significantly lower segmentation accuracy. In Sec.~\ref{sec:appendix_limitations} we analyze more in detail the failure cases of our method and focus specifically also on Scene $5$, which we find to contain several frames with 
extreme
illumination conditions, which makes it particularly challenging to properly reconstruct the geometry of certain parts of the scene.

\section{Multi-step adaptation}
\label{sec:appendix_multi_step_adaptation}

In the following Section, we include in-detail results for the multi-step adaptation experiments, reporting additionally a set of standard metrics used in the continual learning literature. 
We further
demonstrate
the use, enabled by our method, of
images and \pls rendered from \emph{novel} viewpoints in previous scenes for multi-step adaptation.
Remarkably, we find
that this modification induces a further improvement in the retention of knowledge from the previous scenes.
\subsection{Detailed per-step evaluation}
Table~\ref{tab:per_step_evaluation} reports the segmentation performance on the validation views of each scene after each step of adaptation, both for our method and for the baselines of~\cite{Wang2022CoTTA} and~\cite{Frey2022CLSemanticSegmentation}. For each method, we run the experiment $3$ times and report main and standard deviation across the runs. The results complement Tab.~3 in the main paper, confirming in particular that in all the adaptation steps our method is the most effective at preserving knowledge on the previous scenes.
\begin{table}[!h]
\centering
\resizebox{\linewidth}{!}{
\begin{tabular}{@{}lccll@{}}
\toprule
                                              & ACC Metric~\cite{Lopez2017GEM} & A Metric~\cite{Diaz2018DontForget} & FWT~\cite{Lopez2017GEM} & BWT~\cite{Lopez2017GEM} \\ \midrule
CoTTA~\cite{Wang2022CoTTA}                    &        44.6{\scriptsize$\pm$0.0}    &     40.9{\scriptsize$\pm$0.0}      &  \textbf{-0.2}{\scriptsize$\pm$0.0}   &  \textbf{-0.1}{\scriptsize$\pm$0.0}    \\
Mapping~\cite{Frey2022CLSemanticSegmentation} &       45.8{\scriptsize$\pm$0.6}     &  42.1{\scriptsize$\pm$0.5}          &   -1.1{\scriptsize$\pm$0.2}    &     -1.0{\scriptsize$\pm$0.6}  \\
Ours ($\mathbf{I}_\textrm{pre}$ replay only)  &      46.8{\scriptsize$\pm$0.8}      &  43.7{\scriptsize$\pm$0.6}         &  -1.4{\scriptsize$\pm$0.7}     &   
-1.4{\scriptsize$\pm$0.7}  \\
Ours                                          &      \textbf{47.2}{\scriptsize$\pm$0.5}    &     \textbf{44.3}{\scriptsize$\pm$0.2}        &   -1.1{\scriptsize$\pm$0.2}     &     -0.9{\scriptsize$\pm$0.4}   \\ \bottomrule
\end{tabular}}
\caption{Continual learning metrics extracted from Tab.~\ref{tab:per_step_evaluation}. 
}
\label{tab:cl_metrics}
\end{table}
\begin{table*}
    \centering
    \resizebox{\linewidth}{!}{
    \begin{tabular}{lcccccccccccrr}
      \toprule
     Method & Step  & Scene $1$ & Scene $2$ & Scene $3$ & Scene $4$ & Scene $5$ & Scene $6$ & Scene $7$ & Scene $8$ & Scene $9$ & Scene $10$ & Average Prev. & Average\\
      \midrule
      Pre-training & \multicolumn{1}{c}{$-$} & 43.9 & 41.3 & 23.0 & 50.2 & 40.1 & 37.6 & 55.8 & 27.9 &  54.9 & 73.5 & -- & 44.8\\ \midrule
      \multirow{10}{*}{CoTTA~\cite{Wang2022CoTTA}}   & $1$& 44.0{\scriptsize$\pm$0.0}&	\textcolor{gray!60}{{40.9}{\scriptsize$\pm$0.0}}&	\textcolor{gray!60}{22.9{\scriptsize$\pm$0.0}}&	\textcolor{gray!60}{50.3{\scriptsize$\pm$0.0}}&	\textcolor{gray!60}{40.1{\scriptsize$\pm$0.1}}&	\textcolor{gray!60}{37.5{\scriptsize$\pm$0.0}}&	\textcolor{gray!60}{55.9{\scriptsize$\pm$0.0}}&	\textcolor{gray!60}{27.6{\scriptsize$\pm$0.0}}&	\textcolor{gray!60}{54.7{\scriptsize$\pm$0.0}}&	\textcolor{gray!60}{73.6{\scriptsize$\pm$0.0}}&	-- & 44.7{\scriptsize$\pm$0.0}	  \\
     & $2$ & 44.0{\scriptsize$\pm$0.0}&	40.9{\scriptsize$\pm$0.0}&	\textcolor{gray!60}{22.9{\scriptsize$\pm$0.0}}&	\textcolor{gray!60}{50.3{\scriptsize$\pm$0.0}}&	\textcolor{gray!60}{40.1{\scriptsize$\pm$0.0}}&	\textcolor{gray!60}{37.5{\scriptsize$\pm$0.0}}&	\textcolor{gray!60}{55.9{\scriptsize$\pm$0.0}}&	\textcolor{gray!60}{27.6{\scriptsize$\pm$0.0}}&	\textcolor{gray!60}{54.8{\scriptsize$\pm$0.0}}&	\textcolor{gray!60}{73.6{\scriptsize$\pm$0.0}}&	44.0{\scriptsize$\pm$0.0} & 44.7{\scriptsize$\pm$0.0} \\
     & $3$ & 43.6{\scriptsize$\pm$0.1}&	40.7{\scriptsize$\pm$0.1}&	22.7{\scriptsize$\pm$0.0}&	\textcolor{gray!60}{50.1{\scriptsize$\pm$0.1}}&	\textcolor{gray!60}{39.9{\scriptsize$\pm$0.0}}&	\textcolor{gray!60}{37.5{\scriptsize$\pm$0.0}}&	\textcolor{gray!60}{56.1{\scriptsize$\pm$0.0}}&	\textcolor{gray!60}{27.3{\scriptsize$\pm$0.0}}&	\textcolor{gray!60}{54.6{\scriptsize$\pm$0.0}}&	\textcolor{gray!60}{73.7{\scriptsize$\pm$0.0}}& 42.2{\scriptsize$\pm$0.0} & 44.6{\scriptsize$\pm$0.0}\\
     & $4$ & 43.6{\scriptsize$\pm$0.0}&	40.5{\scriptsize$\pm$0.0}&	22.7{\scriptsize$\pm$0.0}&	50.2{\scriptsize$\pm$0.1}&	\textcolor{gray!60}{39.9{\scriptsize$\pm$0.0}}&	\textcolor{gray!60}{37.5{\scriptsize$\pm$0.0}}&	\textcolor{gray!60}{56.0{\scriptsize$\pm$0.1}}&	\textcolor{gray!60}{27.2{\scriptsize$\pm$0.0}}&	\textcolor{gray!60}{54.5{\scriptsize$\pm$0.0}}&	\textcolor{gray!60}{73.7{\scriptsize$\pm$0.0}}& 35.6{\scriptsize$\pm$0.0} & 44.6{\scriptsize$\pm$0.0}\\
     &   $5$ & 43.7{\scriptsize$\pm$0.1}&	40.5{\scriptsize$\pm$0.0}&	22.7{\scriptsize$\pm$0.0}&	\underline{50.2}{\scriptsize$\pm$0.1}&	40.0{\scriptsize$\pm$0.0}&	\textcolor{gray!60}{37.5{\scriptsize$\pm$0.0}}&	\textcolor{gray!60}{55.9{\scriptsize$\pm$0.1}}&	\textcolor{gray!60}{27.1{\scriptsize$\pm$0.0}}&	\textcolor{gray!60}{54.6{\scriptsize$\pm$0.0}}&	\textcolor{gray!60}{73.7{\scriptsize$\pm$0.0}}&	39.3{\scriptsize$\pm$0.0} & 44.6{\scriptsize$\pm$0.0}\\
     &   $6$ & 43.7{\scriptsize$\pm$0.0}&	40.4{\scriptsize$\pm$0.1}&	22.7{\scriptsize$\pm$0.0}&	\underline{50.3}{\scriptsize$\pm$0.1}&	40.0{\scriptsize$\pm$0.0}&	37.5{\scriptsize$\pm$0.0}&	\textcolor{gray!60}{55.9{\scriptsize$\pm$0.1}}&	\textcolor{gray!60}{27.0{\scriptsize$\pm$0.0}}&	\textcolor{gray!60}{54.5{\scriptsize$\pm$0.0}}&	\textcolor{gray!60}{73.7{\scriptsize$\pm$0.0}}&	39.4{\scriptsize$\pm$0.0} & 44.6{\scriptsize$\pm$0.0}\\
     &   $7$ &43.7{\scriptsize$\pm$0.1}&	40.4{\scriptsize$\pm$0.1}&	22.7{\scriptsize$\pm$0.1}&	50.3{\scriptsize$\pm$0.1}&	39.9{\scriptsize$\pm$0.1}&	37.6{\scriptsize$\pm$0.1}&	56.0{\scriptsize$\pm$0.1}&	\textcolor{gray!60}{26.9{\scriptsize$\pm$0.0}}&	\textcolor{gray!60}{54.5{\scriptsize$\pm$0.0}}&	\textcolor{gray!60}{73.7{\scriptsize$\pm$0.0}}&	39.1{\scriptsize$\pm$0.0} & 44.6{\scriptsize$\pm$0.0}														\\
     &   $8$ & 43.7{\scriptsize$\pm$0.0}&	40.4{\scriptsize$\pm$0.1}&	22.7{\scriptsize$\pm$0.1}&	\underline{50.3}{\scriptsize$\pm$0.1}&	39.9{\scriptsize$\pm$0.1}&	37.7{\scriptsize$\pm$0.1}&	56.0{\scriptsize$\pm$0.1}&	\textbf{26.9}{\scriptsize$\pm$0.0}&	\textcolor{gray!60}{54.5{\scriptsize$\pm$0.0}}&	\textcolor{gray!60}{73.7{\scriptsize$\pm$0.0}}&	41.5{\scriptsize$\pm$0.0} & 44.6{\scriptsize$\pm$0.0}															\\
     &   $9$ & 43.7{\scriptsize$\pm$0.0}&	40.3{\scriptsize$\pm$0.1}&	22.7{\scriptsize$\pm$0.1}&	\underline{50.2}{\scriptsize$\pm$0.1}&	39.9{\scriptsize$\pm$0.1}&	\underline{37.7}{\scriptsize$\pm$0.1}&	56.0{\scriptsize$\pm$0.1}&	\underline{26.8}{\scriptsize$\pm$0.0}&	54.5{\scriptsize$\pm$0.0}&	\textcolor{gray!60}{73.8{\scriptsize$\pm$0.0}}& 39.7{\scriptsize$\pm$0.0}	& 44.6{\scriptsize$\pm$0.0}															\\
     &   $10$ & 43.7{\scriptsize$\pm$0.1}&	40.2{\scriptsize$\pm$0.1}&	22.7{\scriptsize$\pm$0.1}&	\underline{50.3}{\scriptsize$\pm$0.1}&	39.9{\scriptsize$\pm$0.1}&	\underline{37.6}{\scriptsize$\pm$0.0}&	56.1{\scriptsize$\pm$0.1}&	26.8{\scriptsize$\pm$0.0}&	54.4{\scriptsize$\pm$0.1}&	\textbf{73.8}{\scriptsize$\pm$0.0}&	41.3{\scriptsize$\pm$0.0} & 44.6{\scriptsize$\pm$0.0}															\\
      \midrule
        \multirow{10}{*}{Mapping~\cite{Frey2022CLSemanticSegmentation}}   &  $1$ &46.8{\scriptsize$\pm$0.4}&	\textcolor{gray!60}{36.0{\scriptsize$\pm$1.6}}&	\textcolor{gray!60}{24.2{\scriptsize$\pm$0.9}}&	\textcolor{gray!60}{48.3{\scriptsize$\pm$0.9}}&	\textcolor{gray!60}{40.0{\scriptsize$\pm$0.9}}&	\textcolor{gray!60}{35.3{\scriptsize$\pm$0.8}}&	\textcolor{gray!60}{55.5{\scriptsize$\pm$0.4}}&	\textcolor{gray!60}{29.2{\scriptsize$\pm$2.3}}&	\textcolor{gray!60}{55.7{\scriptsize$\pm$1.0}}&	\textcolor{gray!60}{73.9{\scriptsize$\pm$0.2}}& -- & 44.5{\scriptsize$\pm$0.5}	 \\
     &   $2$ & 46.5{\scriptsize$\pm$0.1}&	42.1{\scriptsize$\pm$2.0}&	\textcolor{gray!60}{23.6{\scriptsize$\pm$0.9}}&	\textcolor{gray!60}{48.4{\scriptsize$\pm$1.3}}&	\textcolor{gray!60}{41.3{\scriptsize$\pm$1.0}}&	\textcolor{gray!60}{35.5{\scriptsize$\pm$0.7}}&	\textcolor{gray!60}{54.8{\scriptsize$\pm$1.1}}&	\textcolor{gray!60}{28.3{\scriptsize$\pm$0.8}}&	\textcolor{gray!60}{56.5{\scriptsize$\pm$0.9}}&	\textcolor{gray!60}{73.7{\scriptsize$\pm$0.2}}&	46.5{\scriptsize$\pm$0.1} & 45.1{\scriptsize$\pm$0.2} \\
     &   $3$ & 43.0{\scriptsize$\pm$1.2}&	42.6{\scriptsize$\pm$2.8}&	23.6{\scriptsize$\pm$0.7}&	\textcolor{gray!60}{48.5{\scriptsize$\pm$0.7}}&	\textcolor{gray!60}{37.0{\scriptsize$\pm$2.1}}&	\textcolor{gray!60}{33.7{\scriptsize$\pm$0.6}}&	\textcolor{gray!60}{55.5{\scriptsize$\pm$2.0}}&	\textcolor{gray!60}{26.0{\scriptsize$\pm$0.8}}&	\textcolor{gray!60}{54.2{\scriptsize$\pm$1.4}}&	\textcolor{gray!60}{74.1{\scriptsize$\pm$0.3}}& 42.8{\scriptsize$\pm$1.0} & 43.8{\scriptsize$\pm$0.0}\\
     &   $4$ & 45.5{\scriptsize$\pm$0.3}&	42.9{\scriptsize$\pm$2.2}&	23.5{\scriptsize$\pm$0.8}&	\textbf{50.6}{\scriptsize$\pm$2.6}&	\textcolor{gray!60}{38.5{\scriptsize$\pm$0.8}}&	\textcolor{gray!60}{34.1{\scriptsize$\pm$0.9}}&	\textcolor{gray!60}{57.7{\scriptsize$\pm$0.3}}&	\textcolor{gray!60}{26.7{\scriptsize$\pm$1.3}}&	\textcolor{gray!60}{55.8{\scriptsize$\pm$1.9}}&	\textcolor{gray!60}{73.9{\scriptsize$\pm$0.2}}&	37.3{\scriptsize$\pm$0.9} & 44.9{\scriptsize$\pm$0.5}\\
     &   $5$ & 44.9{\scriptsize$\pm$0.6}&	42.9{\scriptsize$\pm$1.2}&	23.5{\scriptsize$\pm$0.7}&	\underline{50.2}{\scriptsize$\pm$2.5}&	\textbf{44.0}{\scriptsize$\pm$0.1}&	\textcolor{gray!60}{34.2{\scriptsize$\pm$0.7}}&	\textcolor{gray!60}{57.3{\scriptsize$\pm$0.6}}&	\textcolor{gray!60}{26.7{\scriptsize$\pm$0.4}}&	\textcolor{gray!60}{54.6{\scriptsize$\pm$1.8}}&	\textcolor{gray!60}{73.6{\scriptsize$\pm$0.7}}&	40.4{\scriptsize$\pm$0.6} & 45.2{\scriptsize$\pm$0.1}														\\
     &   $6$&44.8{\scriptsize$\pm$1.1}&	43.5{\scriptsize$\pm$0.6}&	22.8{\scriptsize$\pm$0.9}&	49.6{\scriptsize$\pm$2.4}&	43.9{\scriptsize$\pm$0.4}&	35.8{\scriptsize$\pm$0.5}&	\textcolor{gray!60}{57.9{\scriptsize$\pm$1.3}}&	\textcolor{gray!60}{25.7{\scriptsize$\pm$0.0}}&	\textcolor{gray!60}{56.1{\scriptsize$\pm$1.7}}&	\textcolor{gray!60}{73.3{\scriptsize$\pm$0.6}}& 40.9{\scriptsize$\pm$0.7} & 45.3{\scriptsize$\pm$0.3}\\
        &$7$ & 43.5{\scriptsize$\pm$1.6}&	43.7{\scriptsize$\pm$0.8}&	22.9{\scriptsize$\pm$1.1}&	\underline{50.4}{\scriptsize$\pm$2.7}&	43.4{\scriptsize$\pm$0.6}&	35.6{\scriptsize$\pm$0.3}&	56.7{\scriptsize$\pm$1.3}&	\textcolor{gray!60}{25.7{\scriptsize$\pm$1.7}}&	\textcolor{gray!60}{55.5{\scriptsize$\pm$2.6}}&	\textcolor{gray!60}{73.7{\scriptsize$\pm$0.4}}&	39.9{\scriptsize$\pm$1.1} & 45.1{\scriptsize$\pm$0.6}											\\
        &$8$ & 42.0{\scriptsize$\pm$0.7}&	43.5{\scriptsize$\pm$1.2}&	23.0{\scriptsize$\pm$0.7}&	\underline{50.3}{\scriptsize$\pm$2.5}&	43.8{\scriptsize$\pm$0.1}&	35.9{\scriptsize$\pm$1.5}&	57.1{\scriptsize$\pm$0.1}&	26.5{\scriptsize$\pm$1.8}&	\textcolor{gray!60}{56.1{\scriptsize$\pm$2.9}}&	\textcolor{gray!60}{73.9{\scriptsize$\pm$0.5}}&	42.2{\scriptsize$\pm$0.5} & 45.2{\scriptsize$\pm$0.2}															\\
      &$9$ & 43.0{\scriptsize$\pm$0.9}&	43.9{\scriptsize$\pm$1.2}&	22.2{\scriptsize$\pm$0.3}&	49.8{\scriptsize$\pm$2.4}&	43.6{\scriptsize$\pm$0.2}&	35.2{\scriptsize$\pm$0.7}&	\underline{56.8}{\scriptsize$\pm$0.2}&	25.6{\scriptsize$\pm$1.2}&	68.3{\scriptsize$\pm$1.4}&	\textcolor{gray!60}{74.1{\scriptsize$\pm$1.2}}&	40.0{\scriptsize$\pm$0.4} & 46.2{\scriptsize$\pm$0.2}															\\
       &$10$ & 42.5{\scriptsize$\pm$0.7}&	43.5{\scriptsize$\pm$1.3}&	22.5{\scriptsize$\pm$0.3}&	49.7{\scriptsize$\pm$2.5}&	43.6{\scriptsize$\pm$0.2}&	35.6{\scriptsize$\pm$1.1}&	55.6{\scriptsize$\pm$1.0}&	26.2{\scriptsize$\pm$1.4}&	65.8{\scriptsize$\pm$4.0}&	72.7{\scriptsize$\pm$1.0}& 42.8{\scriptsize$\pm$0.7}	& 45.8{\scriptsize$\pm$0.6}\\						      \midrule
        \multirow{10}{*}{ Ours ($\mathbf{I}_\textrm{pre}$ replay only)}   &  $1$ &53.3{\scriptsize$\pm$0.7}&	\textcolor{gray!60}{35.4{\scriptsize$\pm$1.8}}&	\textcolor{gray!60}{24.7{\scriptsize$\pm$0.1}}&	\textcolor{gray!60}{49.7{\scriptsize$\pm$1.6}}&	\textcolor{gray!60}{37.4{\scriptsize$\pm$1.0}}&	\textcolor{gray!60}{32.9{\scriptsize$\pm$0.2}}&	\textcolor{gray!60}{55.6{\scriptsize$\pm$1.0}}&	\textcolor{gray!60}{31.9{\scriptsize$\pm$1.1}}&	\textcolor{gray!60}{55.1{\scriptsize$\pm$1.2}}&	\textcolor{gray!60}{74.1{\scriptsize$\pm$0.7}}&	-- & 45.0{\scriptsize$\pm$0.3}															\\
     &   $2$ &  52.3{\scriptsize$\pm$0.3}&	\textbf{48.0}{\scriptsize$\pm$2.4}&	\textcolor{gray!60}{22.2{\scriptsize$\pm$0.4}}&	\textcolor{gray!60}{50.0{\scriptsize$\pm$0.1}}&	\textcolor{gray!60}{43.4{\scriptsize$\pm$0.9}}&	\textcolor{gray!60}{34.4{\scriptsize$\pm$1.4}}&	\textcolor{gray!60}{50.3{\scriptsize$\pm$0.8}}&	\textcolor{gray!60}{29.2{\scriptsize$\pm$1.9}}&	\textcolor{gray!60}{63.4{\scriptsize$\pm$3.5}}&	\textcolor{gray!60}{73.2{\scriptsize$\pm$1.3}}& 52.3{\scriptsize$\pm$0.3} & 46.7{\scriptsize$\pm$0.5}\\
     &   $3$ &51.8{\scriptsize$\pm$1.9}&	43.2{\scriptsize$\pm$1.6}&	20.5{\scriptsize$\pm$0.1}&	\textcolor{gray!60}{48.6{\scriptsize$\pm$0.9}}&	\textcolor{gray!60}{40.0{\scriptsize$\pm$2.1}}&	\textcolor{gray!60}{33.1{\scriptsize$\pm$1.9}}&	\textcolor{gray!60}{55.3{\scriptsize$\pm$0.6}}&	\textcolor{gray!60}{27.7{\scriptsize$\pm$1.5}}&	\textcolor{gray!60}{57.8{\scriptsize$\pm$4.7}}&	\textcolor{gray!60}{73.7{\scriptsize$\pm$0.6}}&	47.5{\scriptsize$\pm$1.1} & 45.2{\scriptsize$\pm$0.6}\\
     &   $4$ & 52.9{\scriptsize$\pm$1.3}&	41.9{\scriptsize$\pm$2.3}&	21.1{\scriptsize$\pm$0.8}&	49.0{\scriptsize$\pm$1.5}&	\textcolor{gray!60}{37.9{\scriptsize$\pm$0.9}}&	\textcolor{gray!60}{34.3{\scriptsize$\pm$1.7}}&	\textcolor{gray!60}{54.5{\scriptsize$\pm$0.6}}&	\textcolor{gray!60}{32.3{\scriptsize$\pm$1.1}}&	\textcolor{gray!60}{55.4{\scriptsize$\pm$0.7}}&	\textcolor{gray!60}{72.9{\scriptsize$\pm$2.0}}&	38.6{\scriptsize$\pm$1.1} & 45.2{\scriptsize$\pm$0.5}\\
     &   $5$ & 51.5{\scriptsize$\pm$0.8}&	41.7{\scriptsize$\pm$1.0}&	21.2{\scriptsize$\pm$0.9}&	48.8{\scriptsize$\pm$1.2}&	43.4{\scriptsize$\pm$0.0}&	\textcolor{gray!60}{35.2{\scriptsize$\pm$0.5}}&	\textcolor{gray!60}{56.4{\scriptsize$\pm$1.0}}&	\textcolor{gray!60}{29.3{\scriptsize$\pm$0.1}}&	\textcolor{gray!60}{53.2{\scriptsize$\pm$2.4}}&	\textcolor{gray!60}{72.2{\scriptsize$\pm$1.0}}& 40.8{\scriptsize$\pm$0.7} & 45.3{\scriptsize$\pm$0.5} \\
       &   $6$ &
    \underline{53.4}{\scriptsize$\pm$1.1}&	44.6{\scriptsize$\pm$1.2}&	20.5{\scriptsize$\pm$0.5}&	49.2{\scriptsize$\pm$1.5}&	\underline{44.4}{\scriptsize$\pm$0.6}&	39.0{\scriptsize$\pm$1.4}&	\textcolor{gray!60}{51.3{\scriptsize$\pm$5.3}}&	\textcolor{gray!60}{30.7{\scriptsize$\pm$2.4}}&	\textcolor{gray!60}{57.3{\scriptsize$\pm$2.3}}&	\textcolor{gray!60}{71.9{\scriptsize$\pm$1.6}}& 42.4{\scriptsize$\pm$0.3} & 46.3{\scriptsize$\pm$0.7}\\
     &   $7$ & \underline{52.1}{\scriptsize$\pm$0.5}&	\underline{45.5}{\scriptsize$\pm$2.5}&	21.1{\scriptsize$\pm$0.3}&	49.7{\scriptsize$\pm$1.1}&	44.0{\scriptsize$\pm$0.4}&	36.6{\scriptsize$\pm$1.8}&	\textbf{62.1}{\scriptsize$\pm$6.2}&	\textcolor{gray!60}{31.1{\scriptsize$\pm$0.7}}&	\textcolor{gray!60}{60.2{\scriptsize$\pm$2.5}}&	\textcolor{gray!60}{74.8{\scriptsize$\pm$0.5}}&	41.5{\scriptsize$\pm$0.7} & 47.7{\scriptsize$\pm$0.1}\\
     &   $8$ & 50.7{\scriptsize$\pm$2.1}&	\underline{47.1}{\scriptsize$\pm$2.5}&	21.0{\scriptsize$\pm$0.7}&	49.3{\scriptsize$\pm$1.6}&	\underline{44.3}{\scriptsize$\pm$1.7}&	38.2{\scriptsize$\pm$1.5}&	\underline{59.6}{\scriptsize$\pm$7.2}&	26.7{\scriptsize$\pm$3.0}&	\textcolor{gray!60}{57.0{\scriptsize$\pm$0.9}}&	\textcolor{gray!60}{74.2{\scriptsize$\pm$0.4}}& \doubleunderline{44.3}{\scriptsize$\pm$1.4} & 46.8{\scriptsize$\pm$1.1}\\
     &   $9$ & 51.4{\scriptsize$\pm$1.4}&	45.6{\scriptsize$\pm$2.5}&	20.0{\scriptsize$\pm$0.8}&	49.3{\scriptsize$\pm$1.4}&	\underline{45.8}{\scriptsize$\pm$1.6}&	36.6{\scriptsize$\pm$1.7}&	56.0{\scriptsize$\pm$4.0}&	26.6{\scriptsize$\pm$3.1}&	65.7{\scriptsize$\pm$5.6}&	\textcolor{gray!60}{73.1{\scriptsize$\pm$0.5}}&	41.4{\scriptsize$\pm$0.6} & 47.0{\scriptsize$\pm$0.9}\\
     &   $10$& 48.7{\scriptsize$\pm$1.5}&	44.5{\scriptsize$\pm$3.9}&	21.1{\scriptsize$\pm$0.3}&	50.1{\scriptsize$\pm$1.5}&	\underline{44.2}{\scriptsize$\pm$1.0}&	35.5{\scriptsize$\pm$1.9}&	\underline{56.8}{\scriptsize$\pm$3.5}&	\underline{28.3}{\scriptsize$\pm$3.2}&	65.8{\scriptsize$\pm$5.4}&	73.0{\scriptsize$\pm$0.5}& 43.9{\scriptsize$\pm$0.9}	& 46.8{\scriptsize$\pm$0.8}							\\
      \midrule
        \multirow{10}{*}{Ours}   &  $1$ & \textbf{53.7}{\scriptsize$\pm$1.3}&	\textcolor{gray!60}{36.6{\scriptsize$\pm$0.5}}&	\textcolor{gray!60}{24.5{\scriptsize$\pm$0.9}}&	\textcolor{gray!60}{49.7{\scriptsize$\pm$0.8}}&	\textcolor{gray!60}{39.7{\scriptsize$\pm$0.9}}&	\textcolor{gray!60}{34.0{\scriptsize$\pm$2.4}}&	\textcolor{gray!60}{56.5{\scriptsize$\pm$1.5}}&	\textcolor{gray!60}{31.7{\scriptsize$\pm$1.3}}&	\textcolor{gray!60}{56.4{\scriptsize$\pm$0.5}}&	\textcolor{gray!60}{74.8{\scriptsize$\pm$0.5}}&	-- & 45.7{\scriptsize$\pm$0.2}															\\
     &   $2$ &  \underline{53.2}{\scriptsize$\pm$0.9}&	46.3{\scriptsize$\pm$0.7}&	\textcolor{gray!60}{23.2{\scriptsize$\pm$0.5}}&	\textcolor{gray!60}{48.5{\scriptsize$\pm$1.1}}&	\textcolor{gray!60}{41.9{\scriptsize$\pm$0.9}}&	\textcolor{gray!60}{33.7{\scriptsize$\pm$1.5}}&	\textcolor{gray!60}{56.4{\scriptsize$\pm$1.7}}&	\textcolor{gray!60}{30.4{\scriptsize$\pm$1.2}}&	\textcolor{gray!60}{59.1{\scriptsize$\pm$0.5}}&	\textcolor{gray!60}{74.1{\scriptsize$\pm$0.5}}&	\doubleunderline{53.2}{\scriptsize$\pm$0.9} & 46.7{\scriptsize$\pm$0.2}\\
     &   $3$ &\underline{52.3}{\scriptsize$\pm$1.1}&	\underline{44.0}{\scriptsize$\pm$0.6}&	\textbf{24.3}{\scriptsize$\pm$2.0}&	\textcolor{gray!60}{49.2{\scriptsize$\pm$0.5}}&	\textcolor{gray!60}{38.5{\scriptsize$\pm$2.8}}&	\textcolor{gray!60}{32.6{\scriptsize$\pm$0.4}}&	\textcolor{gray!60}{53.2{\scriptsize$\pm$0.9}}&	\textcolor{gray!60}{28.0{\scriptsize$\pm$0.3}}&	\textcolor{gray!60}{59.8{\scriptsize$\pm$5.7}}&	\textcolor{gray!60}{73.8{\scriptsize$\pm$0.8}}&	\doubleunderline{48.2}{\scriptsize$\pm$0.8} & 45.6{\scriptsize$\pm$0.3}\\
     &   $4$ & \underline{53.5}{\scriptsize$\pm$0.6}&	\underline{46.3}{\scriptsize$\pm$1.4}&	\underline{24.7}{\scriptsize$\pm$2.9}&	49.1{\scriptsize$\pm$0.9}&	\textcolor{gray!60}{37.3{\scriptsize$\pm$3.4}}&	\textcolor{gray!60}{34.8{\scriptsize$\pm$2.5}}&	\textcolor{gray!60}{54.8{\scriptsize$\pm$2.0}}&	\textcolor{gray!60}{29.8{\scriptsize$\pm$1.2}}&	\textcolor{gray!60}{59.3{\scriptsize$\pm$4.0}}&	\textcolor{gray!60}{72.9{\scriptsize$\pm$0.4}}&	\doubleunderline{41.5}{\scriptsize$\pm$0.8} & 46.3{\scriptsize$\pm$0.6}\\
     &   $5$ & \underline{53.0}{\scriptsize$\pm$1.1}&	\underline{44.4}{\scriptsize$\pm$0.8}&	\underline{24.8}{\scriptsize$\pm$2.9}&	49.1{\scriptsize$\pm$0.7}&	43.7{\scriptsize$\pm$0.3}&	\textcolor{gray!60}{32.7{\scriptsize$\pm$1.7}}&	\textcolor{gray!60}{56.0{\scriptsize$\pm$1.8}}&	\textcolor{gray!60}{29.3{\scriptsize$\pm$1.3}}&	\textcolor{gray!60}{59.0{\scriptsize$\pm$2.3}}&	\textcolor{gray!60}{73.2{\scriptsize$\pm$0.5}}&	\doubleunderline{42.8}{\scriptsize$\pm$0.8} & 46.5{\scriptsize$\pm$0.2} \\
       &   $6$ &
     53.0{\scriptsize$\pm$0.9}&	\underline{45.0}{\scriptsize$\pm$0.9}&	\underline{24.8}{\scriptsize$\pm$2.5}&	49.0{\scriptsize$\pm$0.2}&	44.1{\scriptsize$\pm$0.5}&	\textbf{40.4}{\scriptsize$\pm$1.5}&	\textcolor{gray!60}{54.1{\scriptsize$\pm$1.8}}&	\textcolor{gray!60}{29.5{\scriptsize$\pm$2.1}}&	\textcolor{gray!60}{60.0{\scriptsize$\pm$1.9}}&	\textcolor{gray!60}{72.8{\scriptsize$\pm$0.4}}& \doubleunderline{43.2}{\scriptsize$\pm$0.8} & 47.3{\scriptsize$\pm$0.7}\\
     &   $7$ & 51.6{\scriptsize$\pm$0.4}&	44.7{\scriptsize$\pm$0.5}&	\underline{23.8}{\scriptsize$\pm$2.6}&	49.6{\scriptsize$\pm$0.5}&	\underline{44.1}{\scriptsize$\pm$0.3}&	\underline{39.2}{\scriptsize$\pm$2.0}&	55.8{\scriptsize$\pm$0.8}&	\textcolor{gray!60}{28.6{\scriptsize$\pm$1.8}}&	\textcolor{gray!60}{62.1{\scriptsize$\pm$6.4}}&	\textcolor{gray!60}{73.7{\scriptsize$\pm$0.3}}&	\doubleunderline{42.2}{\scriptsize$\pm$0.8} & 47.3{\scriptsize$\pm$0.8}\\
     &   $8$ & \underline{50.9}{\scriptsize$\pm$0.3}&	46.0{\scriptsize$\pm$0.4}&	\underline{24.3}{\scriptsize$\pm$2.1}&	49.5{\scriptsize$\pm$0.2}&	44.1{\scriptsize$\pm$0.5}&	\underline{38.9}{\scriptsize$\pm$1.2}&	54.9{\scriptsize$\pm$2.1}&	26.2{\scriptsize$\pm$0.9}&	\textcolor{gray!60}{59.5{\scriptsize$\pm$2.4}}&	\textcolor{gray!60}{74.2{\scriptsize$\pm$0.2}}&	44.1{\scriptsize$\pm$0.2} & 46.9{\scriptsize$\pm$0.2}\\
     &   $9$ & \underline{51.6}{\scriptsize$\pm$0.3}&	\underline{46.4}{\scriptsize$\pm$1.5}&	\underline{23.6}{\scriptsize$\pm$2.1}&	49.0{\scriptsize$\pm$0.3}&	44.1{\scriptsize$\pm$0.3}&	37.4{\scriptsize$\pm$1.4}&	55.4{\scriptsize$\pm$2.8}&	25.9{\scriptsize$\pm$0.4}&	\textbf{68.9}{\scriptsize$\pm$3.2}&	\textcolor{gray!60}{73.2{\scriptsize$\pm$0.1}}&	\doubleunderline{41.7}{\scriptsize$\pm$0.2} & 47.6{\scriptsize$\pm$0.2}\\
     &   $10$& \underline{50.8}{\scriptsize$\pm$0.4}&	\underline{44.6}{\scriptsize$\pm$1.1}&	\underline{23.7}{\scriptsize$\pm$2.1}&	49.4{\scriptsize$\pm$0.1}&	43.8{\scriptsize$\pm$0.5}&	37.0{\scriptsize$\pm$1.9}&	54.8{\scriptsize$\pm$1.8}&	26.1{\scriptsize$\pm$0.7}&	\underline{69.6}{\scriptsize$\pm$1.0}&	72.5{\scriptsize$\pm$1.6} & \doubleunderline{44.3}{\scriptsize$\pm$0.3} &	47.2{\scriptsize$\pm$0.5}\\
      \bottomrule
    \end{tabular}}
    \caption{
    Detail of the multi-step performance evaluated on the validation set of each scene. At Step $i$,
    the performance of
    the adapted network $f_{\theta_i}$ on all the scenes
    is reported (for scenes $\mathcal{S}_j, j>i$ the values are greyed out). $\mathrm{Pre{\text-}training}$ denotes the performance of the pre-trained network $f_{\theta_0}$.
    For each Step $i$, we highlight: in \textbf{bold}, the performance of the method which achieves highest $\mathrm{mIoU}$ on the current scene $\mathcal{S}_i$, which is indicative of the adaptation performance; in \underline{underlined}, for each scene $\mathcal{S}_j,\ 1\le j\le i-1$ the performance of the method which  achieves highest $\mathrm{mIoU}$ on $\mathcal{S}_j$, which denotes the ability to preserve previous knowledge; in \doubleunderline{double-underlined}, the performance of the method which achieves highest \emph{average} $\mathrm{mIoU}$ on the previous scenes $\mathcal{S}_j, 1\le j<i$, which also provides an indication of the ability to counteract forgetting. For each method, the results are averaged over $3$ runs. All Ours are with joint training.
    }
    \label{tab:per_step_evaluation}
\end{table*}

To facilitate the analysis of the results, in Tab.~\ref{tab:cl_metrics} we further report a set of metrics commonly used in the continual learning literature.
Our method
achieves the best performance
both
according to
the ACC metric~\cite{Lopez2017GEM} and to the A metric~\cite{Diaz2018DontForget}, meaning that it obtains the best average $\mathrm{mIoU}$ across all
previously visited scenes
both at the final step and
at any arbitrary adaptation step.
The baseline of
CoTTA~\cite{Wang2022CoTTA}
attains
the best forward transfer (FWT)~\cite{Lopez2017GEM} and backward transfer (BWT)~\cite{Lopez2017GEM},
which indicate respectively the influence that previous scenes have on the performance on future scenes and the influence that adaptation on the current scenes has on the performance on the previous scenes (negative BWT corresponds to catastrophic forgetting). An important point to notice, however, is that the performance of CoTTA also does not vary significantly with respect to the pre-trained model, and in particular does not improve on average.
Among the other methods, our method achieves the best FWT and BWT, which demonstrates the effectiveness of our NeRF-based replay buffer in alleviating forgetting and improving the generalization performance.

\subsection{``Replaying" from novel viewpoints\label{sec:appendix_replay_novel_viewpoints}}
A key 
feature
enabled by
our method is the possibility of rendering both photorealistic color images and \pls from any arbitrary viewpoint inside a reconstructed scene. Crucially, this can include
also
\emph{novel} viewpoints not seen during deployment and training, which can then
be used for adaptation, at the fixed storage cost
given
by
the size of the NeRF model parameters.
In the following, we present an experiment demonstrating this idea in the multi-step adaptation scenario.
Using the notation introduced in the paper, in each step $i\in\{1,\dots,10\}$, the semantic segmentation network $f_{\theta_{i-1}}$ is adapted on scene $\mathcal{S}_i$, and for each previous scene $\mathcal{S}_j,\ 1\le j<i$ images and \pls rendered from viewpoints $\hat{\mathbf{P}}_j$ are inserted in a rendering buffer and mixed to the data from the current scene. However, unlike the experiments in the main paper, 
we do not enforce
that for each scene $\mathcal{S}_j$ the viewpoints $\hat{\mathbf{P}}_j$ used for the rendering buffer coincide with those used in training $\mathbf{P}_j:=\{\boldsymbol{P}_j^k\}_{k\in\{1,\cdots,|\mathbf{I}_j|\}}$, but instead allow novel viewpoints to be used, that is,  $|\hat{\mathbf{P}}_j\backslash(\hat{\mathbf{P}}_j\cap\mathbf{P}_j)|>0$.

\begin{figure*}
\vspace{-30pt}
\centering
\begin{subfigure}[b]{0.45\textwidth}
\centering
\includegraphics[width=\linewidth]{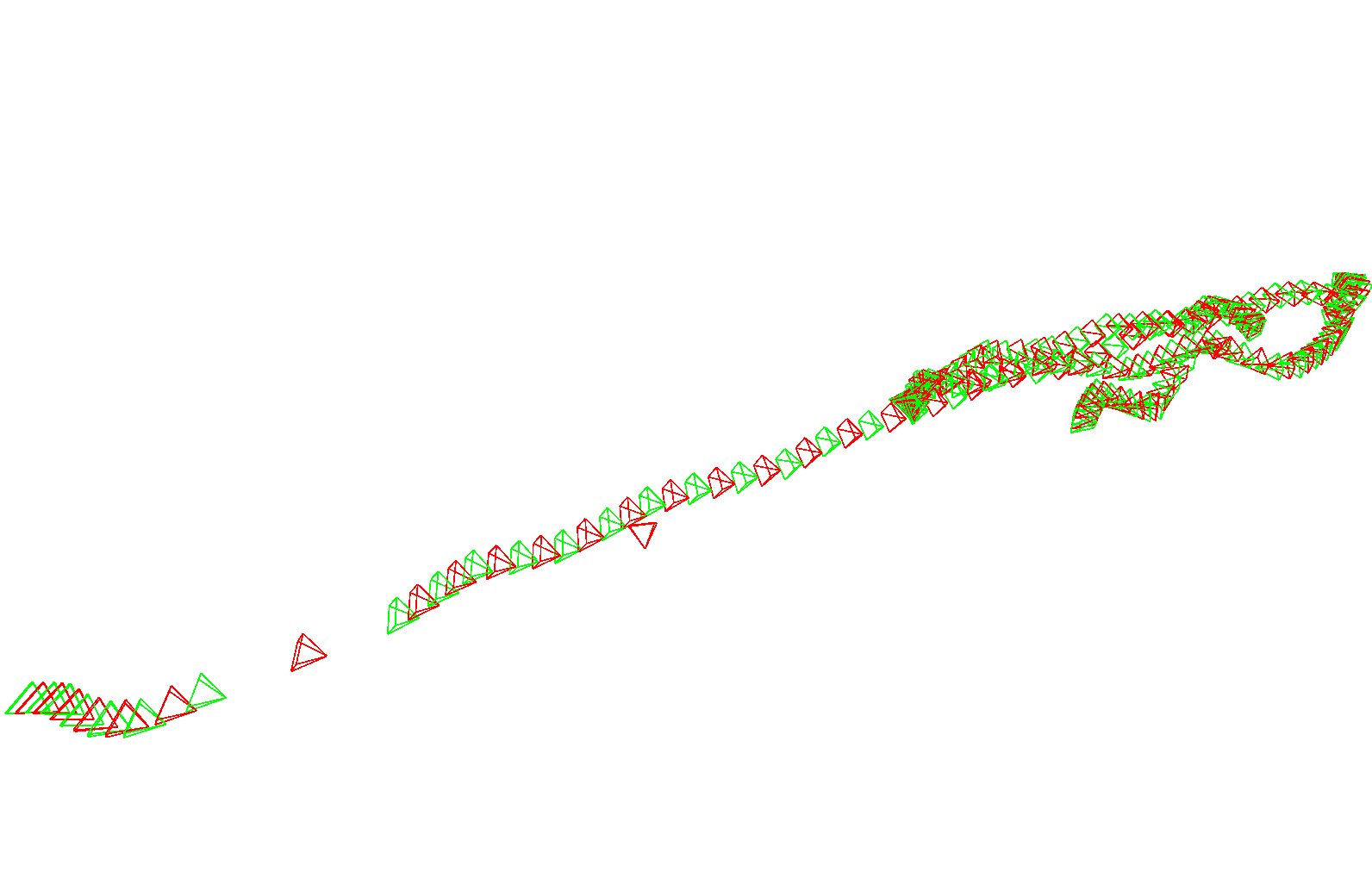}
\end{subfigure}
\hfill
\begin{subfigure}[b]{0.45\textwidth}
\centering
\includegraphics[width=\linewidth]{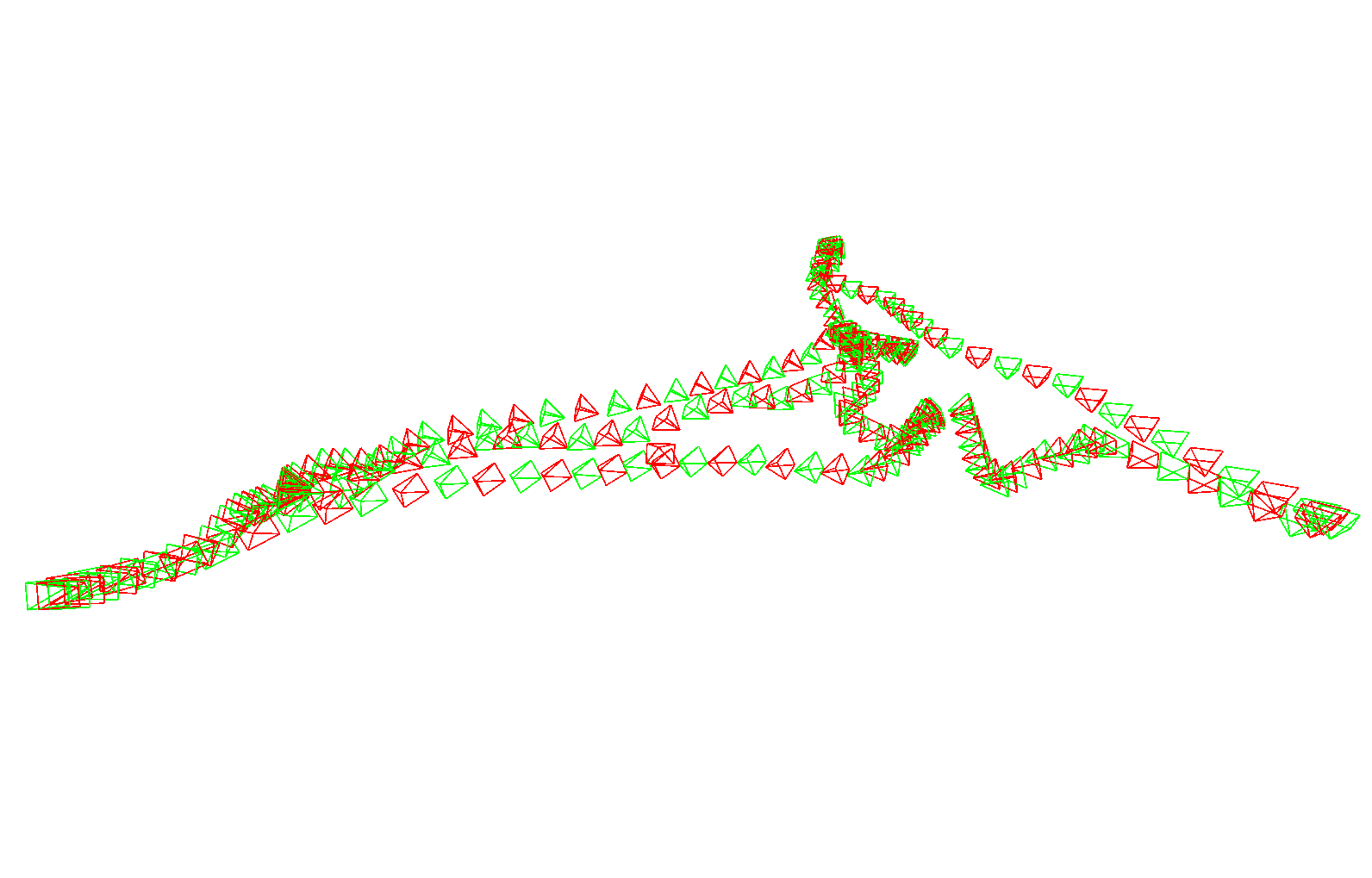}
\end{subfigure}
 \caption{Visualization of the novel viewpoints used for adaptation in Sec.~\ref{sec:appendix_replay_novel_viewpoints} for two example scenes (Scene $5$, left side, and Scene $6$, right side). The viewpoints $\mathbf{P}_j$ used for training and the novel viewpoints $\hat{\mathbf{P}}_j$ used for ``replay" are shown in green and red, respectively.}
 \label{fig:interpolated_poses}
 \vspace{10pt}
 \end{figure*}

Specifically, in the
presented
experiment we apply simple average interpolation of the training poses, and
for each viewpoint $\hat{\boldsymbol{P}}_j^k\in\hat{\mathbf{P}}_j$
we compute its
rotation component
through spherical linear interpolation~\cite{Shoemake1985Slerp} of the rotation components of $\boldsymbol{P}_j^k$ and $\boldsymbol{P}_j^{k+1}$, and
its
translation component
as the average of the translation components of $\boldsymbol{P}_j^k$ and $\boldsymbol{P}_j^{k+1}$.
An example visualization of the obtained poses can be found in Fig.~\ref{fig:interpolated_poses}. The results of the experiment are shown in Tab.~\ref{tab:multi_step_novel_viewpoints} in the main paper.

As can be observed from the $\mathrm{Adapt}$ results, replaying from novel viewpoints achieves similar adaptation performance on the current scene as the other baselines of Ours, but with a slightly larger variance. 

The crucial observation, however, is that this strategy outperforms all the other baselines in terms of retention of previous knowledge ($\mathrm{Previous}$) in almost all the steps, and improves on our method with replay of the training viewpoints on average by ${0.7}\%\ {\mathrm{mIoU}}$.
This improvement can be attributed to the novel viewpoints effectively acting as a positive augmentation mechanism
and inducing
an increase of knowledge on the previous scenes.
In other words, rather than simply counteracting forgetting,
the model de facto keeps
learning
on the previous scenes, through the use of newly generated data points.

We believe this opens up interesting avenues for replay-based adaptation.
In particular, more sophisticated strategies
to select the viewpoints from which to render could be designed, and further increase the knowledge retention on the previous scenes, without reducing the performance on the current scene.

\section{Memory footprint}
\label{appendix:memory_footprint}
In the following, we report the memory footprint of the different methods, denoting with $N$ the number of previous scenes at a given adaptation step.

For each previous scene, our method stores the corresponding NeRF model, which has a size of \SI{50.0}{MB} with Instant-NGP~\cite{Mueller2022InstantNGP, torch-ngp} and of \SI{4.9}{MB} with \snerf~\cite{Zhi2021SemanticNeRF}. This results in either $(N\times\num{50.0})\mathrm{MB}$ or $(N\times\num{4.9})\mathrm{MB}$ of total data being stored in the \emph{long-term} memory. Note however that during adaptation we only render data from a small subset of views to populate the replay buffer, hence the effective size of the data from the previous scenes that need to be stored in running memory during adaptation is \SI{14.0}{MB}. Additionally, we save 
one
randomly selected data point
every $10$ samples in the pre-training dataset,
taking up
additional \SI{64.6}{MB} of space.

Similarly to us, the method of~\cite{Frey2022CLSemanticSegmentation} requires \SI{14.0}{MB} for the replay buffer and \SI{64.6}{MB} for the replay from the pre-training dataset, but stores voxel-based maps instead of NeRF models, taking up
\SI{71.8}{MB} for each scene. Importantly, since the voxel-based maps only 
include
semantic information and cannot be used to render \textit{color} images, the method of~\cite{Frey2022CLSemanticSegmentation} additionally needs to save
color images for the training viewpoints. 
In the $10$ scenes that we used for our experiments, their size amounted on average to approximately \SI{30.0}{MB} per scene, resulting in a total storage space of around $(N\times\SI{101.8}{MB})$ required for the previous scenes.

In each
step, in addition to the model that gets adapted on the current scene, CoTTA~\cite{Wang2022CoTTA} requires storing the teacher model from which \pls for online adaptation are generated, and an additional version of the original, pre-trained model, to preserve source knowledge. The parameters of the DeepLab network used in our experiments have a size of \SI{224.3}{MB}, resulting in a total of $(2\times224.3)\mathrm{MB} = \SI{448.6}{MB}$ of data that need to be stored.
\begin{figure}
    \centering
    \includegraphics[width=\linewidth]{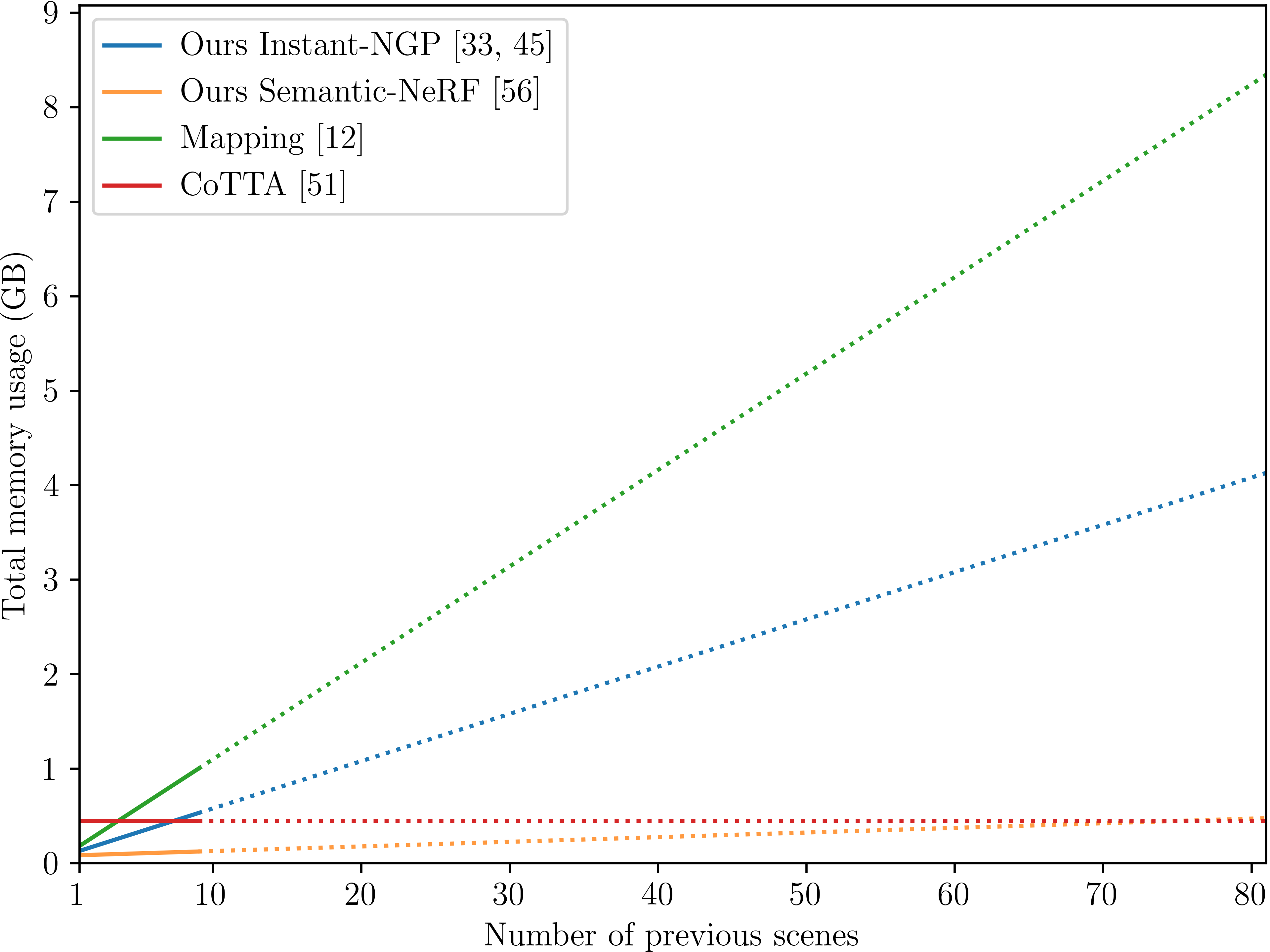}
    \caption{
   Memory footprint of the different methods as a function of the number of the previous scenes. Please refer to the text and to Tab.~\ref{tab:memory_footprint} for a detailed explanation. We use solid lines for the number of scenes used in our experiments.
    }
    \label{fig:memory_footprint}
    \vspace{-10pt}
\end{figure}

\begin{table*}[!ht]
\centering
\resizebox{0.77\linewidth}{!}{
\begin{tabular}{llcccr}
\toprule
& & \multicolumn{2}{c}{Previous scenes} & Source knowledge & Total \\
\cmidrule(r){3-4}
& & Offline & Online\\
\midrule
\multirow{2}{*}{Ours} & Instant-NGP~\cite{Mueller2022InstantNGP, torch-ngp} & $(N\times\num{49.9})\mathrm{MB}^\star$ &
\multirow{2}{*}{\SI{14.0}{MB}} & \multirow{2}{*}{\SI{64.6}{MB}} & $(78.6 + N\times49.9)\mathrm{MB}$\\
& Semantic-NeRF~\cite{Zhi2021SemanticNeRF} & $(N\times\num{4.9})\mathrm{MB}^\star$ & & & $(78.6 + N\times4.9)\mathrm{MB}$\\
\multicolumn{2}{l}{CoTTA~\cite{Wang2022CoTTA}} & -- & $\SI{224.3}{MB}^\dagger$ & $\SI{224.3}{MB}^\dagger$ &   \SI{448.6}{MB}\\
\multicolumn{2}{l}{Mapping~\cite{Frey2022CLSemanticSegmentation}} & $\sim(N\times\num{101.8})\mathrm{MB}^{\star\star}$ & \SI{14.0}{MB} & \SI{64.6}{MB} & $\sim(78.6 + N\times101.8)\mathrm{MB}$ \\
\bottomrule
\end{tabular}
}
\caption{Comparison of the memory footprint of different methods. $N$ denotes the number of previous scenes. ${}^\star$~The numbers refer to the storage cost required by the NeRF models. For actual adaptation (Online), only renderings from a subset of views are used, and inserted in a memory buffer of size $\SI{14.0}{MB}$. ${}^{\star\star}$~The numbers refer to the storage cost required by each voxel-based map (\SI{71.8}{MB}), plus the explicit training views that need to be stored for each scene, which amount to an average of $\sim\SI{30.0}{MB}$ per scene. Similarly to Ours, for actual adaptation, a memory buffer of size $\SI{14.0}{MB}$ is used. ${}^\dagger$~CoTTA requires storing a teacher model for online adaptation, and an additional version of the original, pre-trained model, to preserve source knowledge. 
}
\label{tab:memory_footprint}
\vspace{-10pt}
\end{table*}

A comparison of the memory footprint of the different methods as a function of the number of previous scenes can be found in 
tabular form in Tab.~\ref{tab:memory_footprint} and
in graphical form in
Fig.~\ref{fig:memory_footprint}. For our method and for~\cite{Frey2022CLSemanticSegmentation}, we include in the total size both the data stored offline and the one inserted in the replay buffer.

Note that using the lighter implementation of \snerf~\cite{Zhi2021SemanticNeRF}, the comparison is in our favour up to 
$75$ scenes, and up to 
$91$ scenes when only considering the size of the NeRF models.

\vspace{-1ex}
\section{Further visualizations}
\label{sec:appendix_further_visualizations}
\begin{figure*}[!ht]
\centering
\def\colwidth{0.11\textwidth}
\newcolumntype{M}[1]{>{\centering\arraybackslash}m{#1}}
\addtolength{\tabcolsep}{-4pt}
\begin{tabular}{m{0em} M{\colwidth} M{\colwidth} M{\colwidth}  M{\colwidth}   M{\colwidth}    }
 & {\small Ground-truth images} & {\small Ground-truth labels}  & {\small Mapping}~\cite{Frey2022CLSemanticSegmentation} &  {\small Ours Fine-tuning} & {\small Ours Joint Training}\tabularnewline
& 
\includegraphics[width=\linewidth]{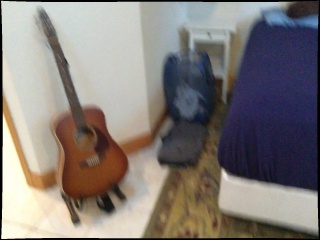} & 
\includegraphics[width=\linewidth]{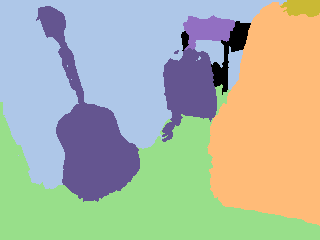} &
\includegraphics[width=\linewidth]{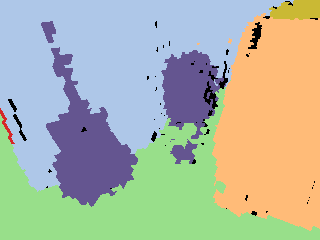} & 
\includegraphics[width=\linewidth]{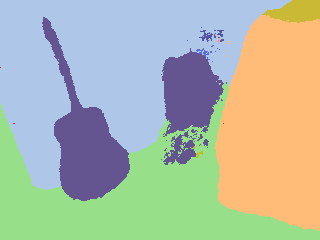} &
\includegraphics[width=\linewidth]{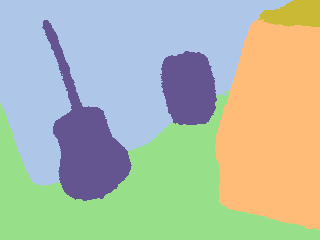} 
  \tabularnewline
&
\includegraphics[width=\linewidth]{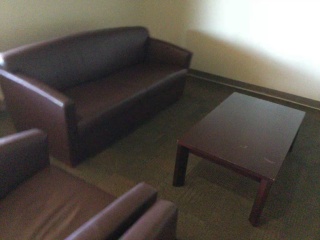} & 
\includegraphics[width=\linewidth]{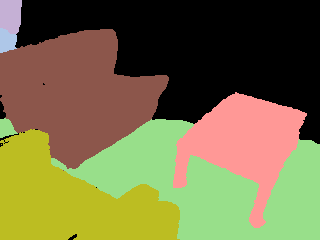} &
\includegraphics[width=\linewidth]{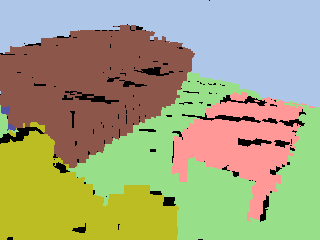} & 
\includegraphics[width=\linewidth]{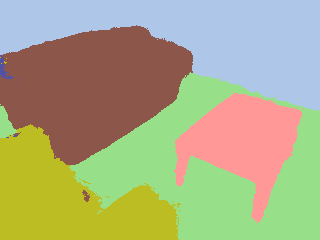} &
\includegraphics[width=\linewidth]{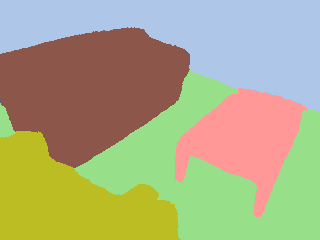} 
  \tabularnewline
& 
\includegraphics[width=\linewidth]{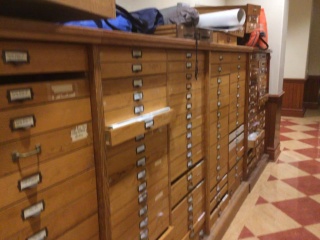} & 
\includegraphics[width=\linewidth]{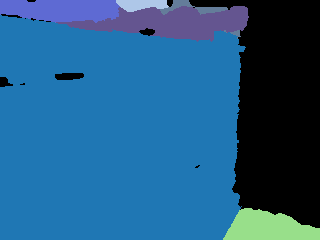} &
\includegraphics[width=\linewidth]{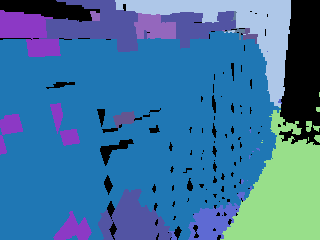} & 
\includegraphics[width=\linewidth]{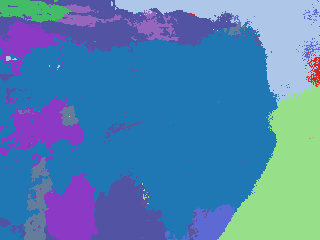} &
\includegraphics[width=\linewidth]{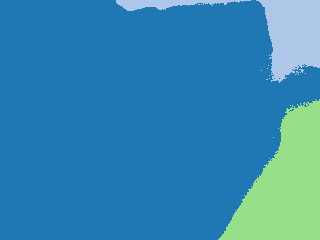} 
  \tabularnewline
\end{tabular}
\addtolength{\tabcolsep}{4pt}
\vspace{-2ex}
\caption{Comparison of example \pls obtained on the training views by the different methods.
}
\label{fig:further_visualizations_pseudolabels} 
\vspace{1ex}
\end{figure*}

In Fig.~\ref{fig:further_visualizations_pseudolabels} we provide examples of the \pls produced on the training views by our method and by the different baselines.
As previously observed by the authors of~\cite{Frey2022CLSemanticSegmentation}, the mapping-based \pls suffer from artifacts induced by the discrete voxel-based representation.
Thanks to the continuous representation enabled by the coordinate-based multi-layer perceptrons, our NeRF-based \pls produce instead smoother and sharper segmentations.
However, they occasionally fail to assign a uniform class label to each object in the scene (cf. last row in Fig.~\ref{fig:further_visualizations_pseudolabels}). This phenomenon,
which we also observe in the mapping-based \pls, can be attributed to the inconsistent per-frame predictions of DeepLab, that cannot be fully filtered-out by the 3D fusion mechanism. By jointly training the per-frame segmentation network and the 3D-aware \snerf, we are however able to effectively reduce the extent of this phenomenon, producing more uniform \pls.

\vspace{-2ex}
\begin{figure*}[!ht]
\centering
\def\colwidth{0.11\textwidth}
\newcolumntype{M}[1]{>{\centering\arraybackslash}m{#1}}
\addtolength{\tabcolsep}{-4pt}
\begin{tabular}{m{0em} M{\colwidth} M{\colwidth} M{\colwidth} M{\colwidth}  M{\colwidth}   M{\colwidth}  M{\colwidth}  }
 & {\small Ground-truth images} & {\small Ground-truth labels}  &
{\small Pre-train}
 & {\small CoTTA}~\cite{Wang2022CoTTA} & {\small Mapping}~\cite{Frey2022CLSemanticSegmentation}  & {\small Ours Fine-tuning} & {\small Ours Joint Training}\tabularnewline
& 
\includegraphics[width=\linewidth]{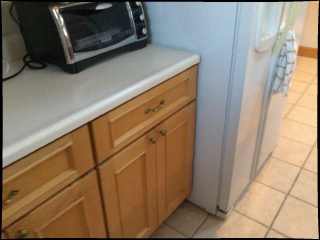} &
\includegraphics[width=\linewidth]{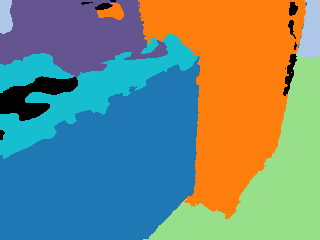} &\includegraphics[width=\linewidth]{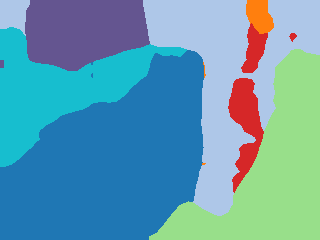} &
\includegraphics[width=\linewidth]{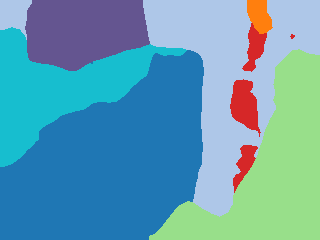} &\includegraphics[width=\linewidth]{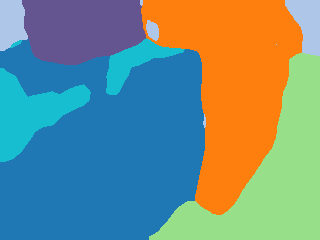} &
\includegraphics[width=\linewidth]{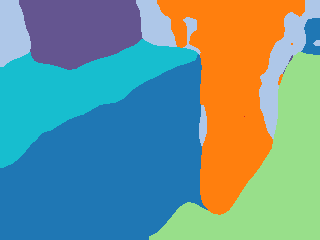}  &
\includegraphics[width=\linewidth]{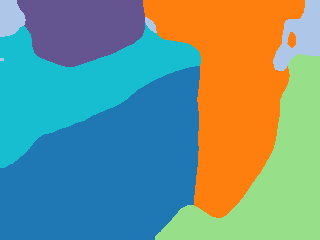} 
  \tabularnewline
&
\includegraphics[width=\linewidth]{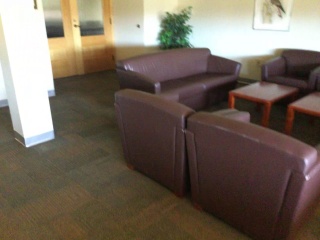} &
\includegraphics[width=\linewidth]{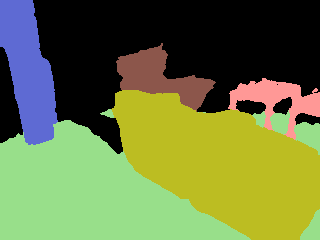} &\includegraphics[width=\linewidth]{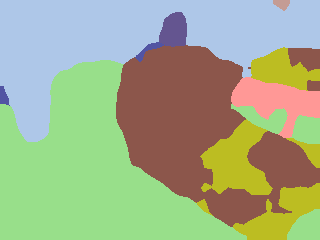} &
\includegraphics[width=\linewidth]{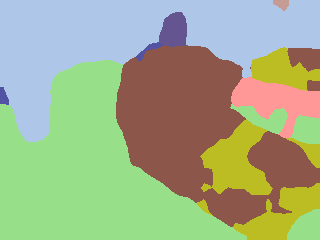} &\includegraphics[width=\linewidth]{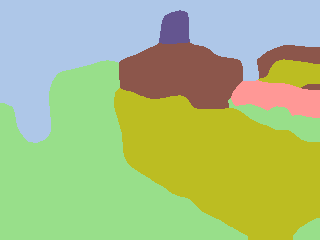} &
\includegraphics[width=\linewidth]{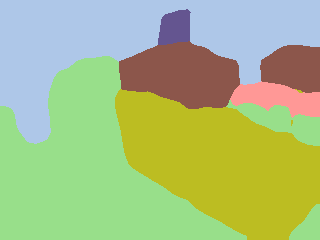}  &
\includegraphics[width=\linewidth]{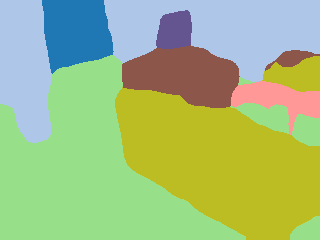} 
\tabularnewline
& 
\includegraphics[width=\linewidth]{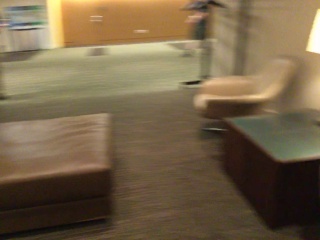} &
\includegraphics[width=\linewidth]{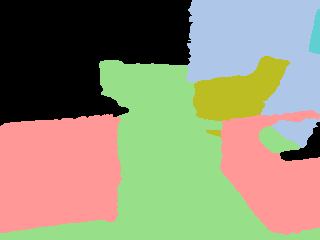} &\includegraphics[width=\linewidth]{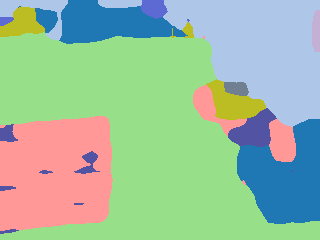} &
\includegraphics[width=\linewidth]{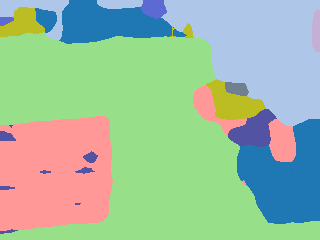} &\includegraphics[width=\linewidth]{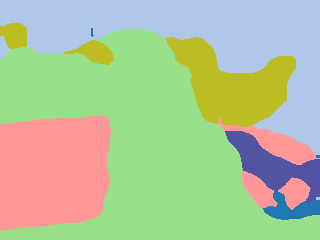} &
\includegraphics[width=\linewidth]{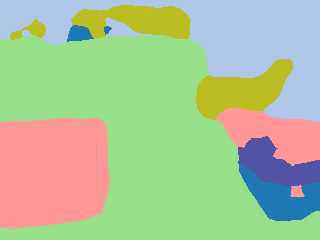}  &
\includegraphics[width=\linewidth]{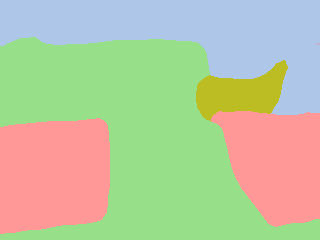} 
\tabularnewline
\end{tabular}
\addtolength{\tabcolsep}{4pt}
\vspace{-2ex}
\caption{Comparison of the predictions of the semantic segmentation network on the validation views, when adapted using the different methods. 
\label{fig:further_visualizations_network_predictions}}
\vspace{1ex}
\end{figure*}
\vspace{2ex}
Figure~\ref{fig:further_visualizations_network_predictions} further shows examples of the predictions returned by the segmentation network on the validation views after 
being
adapted using the different methods. In accordance with what observed in the quantitative evaluations, while being able to preserve knowledge, CoTTA achieves limited improvements with respect to the initial performance. As a consequence, the predicted labels match very closely those of the pre-trained network.
Fusing the predictions from multiple viewpoints into a 3D representation allows both the baseline of~\cite{Frey2022CLSemanticSegmentation} and our method to
reduce the amount of
artifacts due to misclassifications in the per-frame predictions.
The positive effect of this 3D fusion can be
successfully
transferred to the segmentation network through adaptation, as visible by comparing the predictions in the three rightmost columns of Fig.~\ref{fig:further_visualizations_network_predictions} to those of the pre-trained network (third column from the left in Fig.~\ref{fig:further_visualizations_network_predictions}).
We observe that fine-tuning with NeRF-based \pls instead of voxel-based \pls often results in a more consistent class assignment to different pixels of the same instance. This effect is amplified when using joint training, which often 
produces
more accurate
\pls compared to fine-tuning.

\section{Limitations}
\label{sec:appendix_limitations}
\begin{figure*}[!ht]
\centering
\def\colwidth{0.115\textwidth}
\newcolumntype{M}[1]{>{\centering\arraybackslash}m{#1}}
\addtolength{\tabcolsep}{-4pt}
\begin{tabular}{m{0em} M{\colwidth}  M{\colwidth} M{\colwidth} M{\colwidth} M{\colwidth}  M{\colwidth}  M{\colwidth}  M{\colwidth}   }
 & {\small Ground-truth images} & {\small NeRF-rendered images} & {\small Ground-truth depth} & {\small NeRF-rendered depth}  & {\small Ground-truth labels} &
{\small Predictions  pre-training
}
 & {\small NeRF-rendered semantics} & {\small Predictions after adaptation}
\tabularnewline
 & 
\includegraphics[width=\linewidth]{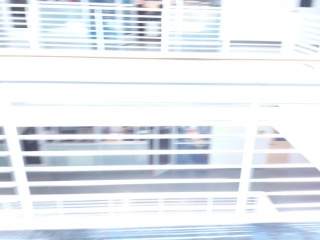} & \includegraphics[width=\linewidth]{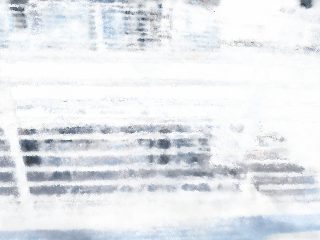}  & 
\includegraphics[width=\linewidth]{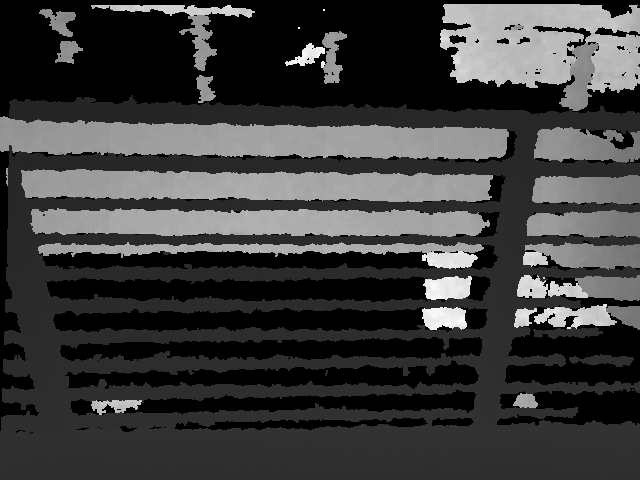} & 
\includegraphics[width=\linewidth]{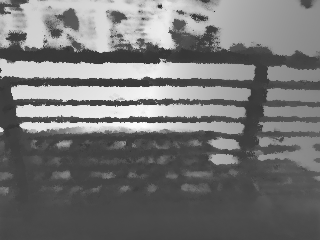} 
&
\includegraphics[width=\linewidth]{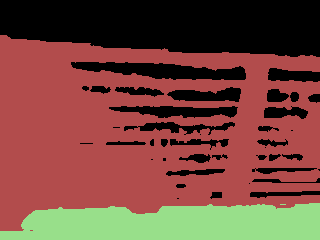} &\includegraphics[width=\linewidth]{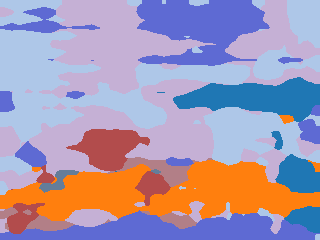}  &\includegraphics[width=\linewidth]{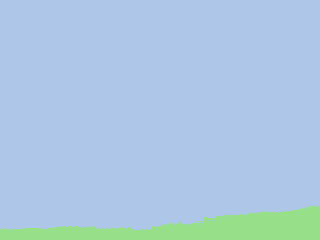} &
\includegraphics[width=\linewidth]{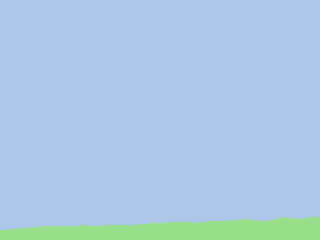} 
  \tabularnewline
  &
\includegraphics[width=\linewidth]{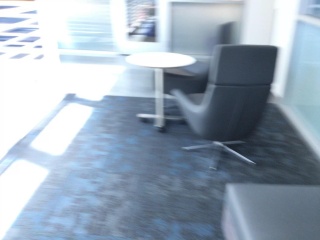} & \includegraphics[width=\linewidth]{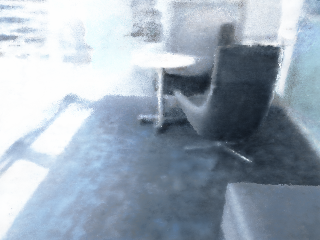}  & 
\includegraphics[width=\linewidth]{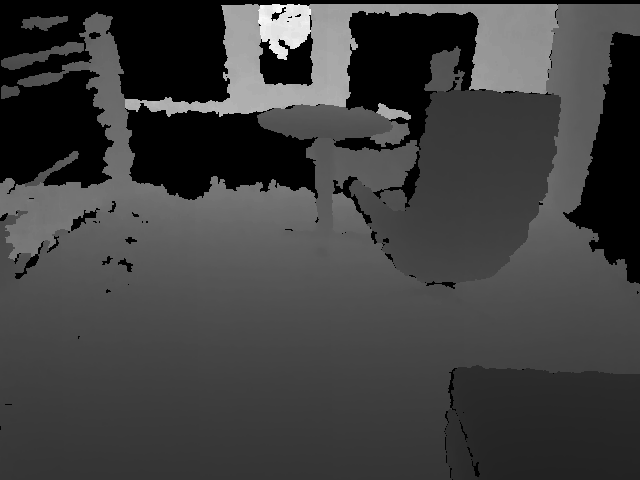} & 
\includegraphics[width=\linewidth]{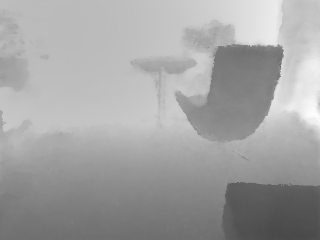} 
&
\includegraphics[width=\linewidth]{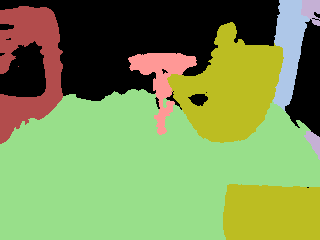} &\includegraphics[width=\linewidth]{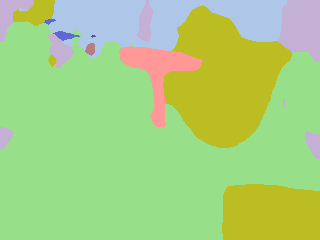}  &\includegraphics[width=\linewidth]{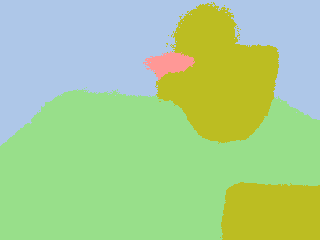} &
\includegraphics[width=\linewidth]{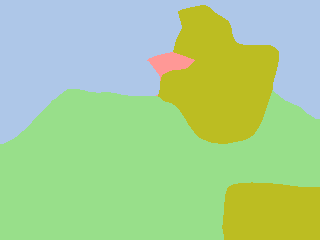} 
  \tabularnewline
&
\includegraphics[width=\linewidth]{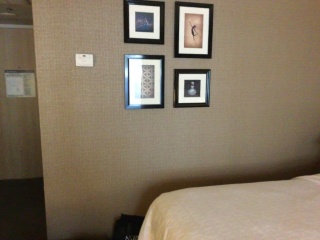} & \includegraphics[width=\linewidth]{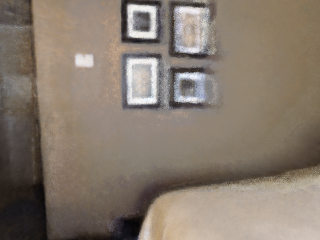}  & 
\includegraphics[width=\linewidth]{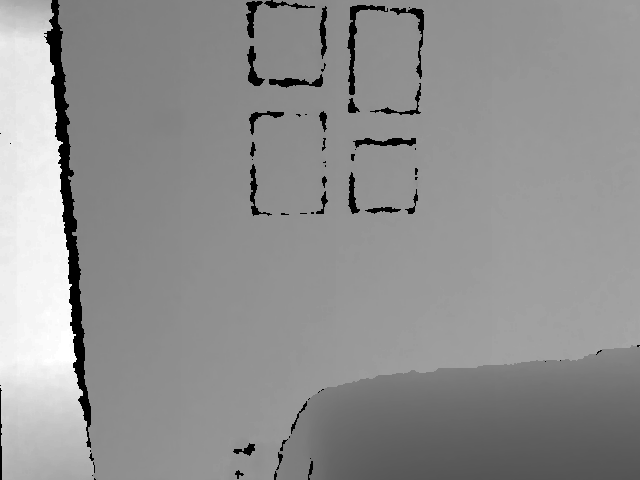} & 
\includegraphics[width=\linewidth]{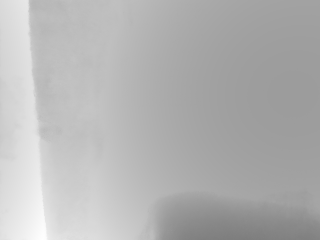} 
&
\includegraphics[width=\linewidth]{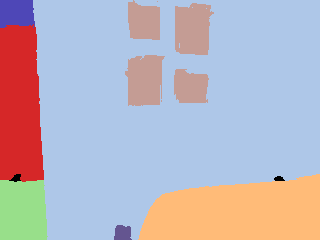} &\includegraphics[width=\linewidth]{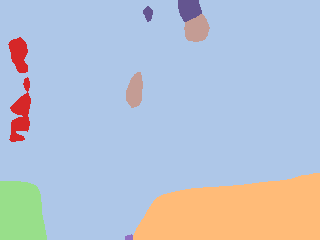}  &\includegraphics[width=\linewidth]{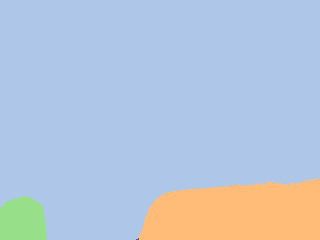} &
\includegraphics[width=\linewidth]{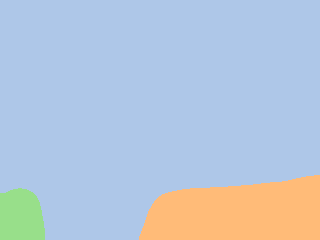} 
\tabularnewline
\end{tabular}
\addtolength{\tabcolsep}{4pt}
\vspace{-2ex}
\caption{\small Examples of failure cases. First row: Poor lighting, incomplete ground-truth depth measurements, and noisy initial predictions result in both the rendered \pls and the predictions from the adapted network assigning uniform labels to large parts of the scene and failing to correctly segment fine details in the scene. Second row: Motion blur, diffusion lighting, shadows, and insufficient
number of observations
can also degrade the reconstruction quality and make the label propagate to the wrong objects. Third row: Specular effects can break the assumptions of the volume rendering formulation of NeRF; large flat areas with small variations in depth can be hard to reconstruct, resulting in smoothed-out, uniform labels for the background.
\label{fig:failure_cases}}
\end{figure*}
Since our approach relies on the assumption that a good reconstruction of the scene
can be obtained,
we find that our method
achieves
suboptimal performance when this assumption is not fulfilled.
This is the case for instance for Scene $5$ (cf. first and second row in Fig.~\ref{fig:failure_cases}), in which a large number of frames are overexposed and the ground-truth depth measurements are missing for a large part of the 
frame.
Specular effects
(cf. third row in Fig.~\ref{fig:failure_cases}) can further break the assumptions required by the volume rendering formulation of NeRF.
Related to these problems is also the quality of the initial predictions of the segmentation network: Particularly when lighting conditions are poor, we observe that the predictions of the pre-trained segmentation network are very noisy (see, \eg, first row in Fig.~\ref{fig:failure_cases}). The combination of these factors results in the \pls produced by our method 
assigning a uniform label to a large part of the scene and failing to correctly segment smaller details.

We observe that these degenerate cases can have a particularly large influence on the quality of the \pls and of the network predictions when jointly training the segmentation network and NeRF.
We hypothesize that this might be due to the 2D-3D knowledge transfer enabled by our method
inducing
a negative feedback loop when poor segmentation predictions are combined with suboptimal reconstructed geometry.
A possible way to tackle this problem in future work
is by making
use of
regularization techniques, for instance by
limiting large deviations of the predictions across adaptation steps, to avoid collapse, or
by minimizing
the entropy of the semantic predictions of both NeRF and the segmentation network.

A general limitation of our method is that it assumes scenes to be static. Extending the pipeline to handle dynamic scenes through the use of temporally-aware NeRFs~\cite{Pumarola2021D-NeRF, Park2021Nerfies} is an interesting direction for future work.

\end{document}